\documentclass{article}

 \usepackage[main, final]{neurips_2026}

\usepackage[utf8]{inputenc} %
\usepackage[T1]{fontenc}    %
\usepackage{hyperref}       %
\usepackage{url}            %
\usepackage{booktabs}       %
\usepackage{amsfonts}       %
\usepackage{nicefrac}       %
\usepackage{microtype}      %
\usepackage{xcolor}         %

\usepackage{graphicx}
\usepackage{xspace}
\usepackage{subcaption}
\usepackage{amsmath}
\usepackage{amsthm}
\usepackage[capitalize,noabbrev]{cleveref}
\usepackage{algorithm}
\usepackage{algorithmic}
\usepackage{listings}
\usepackage{multirow}
\usepackage{booktabs}
\usepackage{adjustbox}
\usepackage{wrapfig}
\usepackage{caption}

\newcommand{\etal}{et~al.}
\newcommand{\mypar}[1]{\vspace{-2mm}\paragraph{#1}}

\usepackage{amsmath,amsfonts,bm}

\def\eqref#1{equation~\ref{#1}}

\def\1{\bm{1}}

\def\rd{{\textnormal{d}}}

\def\vmu{{\bm{\mu}}}
\def\valpha{{\bm{\alpha}}}

\def\va{{\bm{a}}}

\def\vp{{\bm{p}}}

\def\vw{{\bm{w}}}
\def\vx{{\bm{x}}}

\def\vz{{\bm{z}}}

\DeclareMathAlphabet{\mathsfit}{\encodingdefault}{\sfdefault}{m}{sl}
\SetMathAlphabet{\mathsfit}{bold}{\encodingdefault}{\sfdefault}{bx}{n}

\def\gA{{\mathcal{A}}}

\def\gD{{\mathcal{D}}}

\def\gX{{\mathcal{X}}}
\def\gY{{\mathcal{Y}}}
\def\gZ{{\mathcal{Z}}}

\def\sH{{\mathbb{H}}}

\def\sP{{\mathbb{P}}}

\def\sR{{\mathbb{R}}}

\DeclareMathOperator{\sign}{sign}

\title{Guiding Data Allocation for Robust Subpopulation Generalization}

\author{%
  Zhaoying Pan, \quad Yipei Wang, \quad Shenyu Lu, \quad Xiaoqian Wang\textsuperscript{\dag} \\[0.3em]
  Elmore Family School of Electrical and Computer Engineering, Purdue University \\[0.2em]
  \texttt{\{pan433, wang4865, lu876, joywang\}@purdue.edu} \\
}

\begin{document}
\maketitle

\begingroup
\renewcommand{\thefootnote}{\fnsymbol{footnote}}
\setcounter{footnote}{2}
\footnotetext{Corresponding author.}
\endgroup

\begin{abstract}

    Machine learning models often achieve high average accuracy in the training data while performing poorly on test data due to subpopulation shift. A common data-centric solution is to enforce balanced subgroup proportions, but full balance can be costly, infeasible, and not always necessary. In this work, we study whether balanced training data is the unique optimal configuration for robust subpopulation generalization. We extend a Gaussian feature-model framework to theoretically characterize the performance landscape across general training subgroup distributions. We show that multiple configurations, including substantially imbalanced ones, can perform comparably to the balanced one. We further derive gradient-based directions for adjusting subgroup proportions to improve adjusted accuracy and worst-group accuracy more efficiently. These findings suggest that strategically allocating data offers a more flexible and principled path to robust performance than simple balancing. The code is available at \href{https://github.com/zhaoyingpan/subpop-data-allocation}{https://github.com/zhaoyingpan/subpop-data-allocation}.

\end{abstract}

\section{Introduction}
    Machine learning models trained on data with imbalanced subgroup distributions often achieve high overall test performance but suffer from performance disparities in subgroups~\cite{geirhos2020shortcut, koh2021wilds, quinonero2022dataset} or favor the majority groups~\cite{he2009learning, johnson2019survey, zhang2023deep}. 
    This challenge has been widely studied under topics such as spurious correlation~\cite{arjovsky2019invariant, sagawa2020investigation, geirhos2020shortcut}, fairness~\cite{hardt2016equality, kearns2018preventing, mehrabi2021survey}, and class imbalance~\cite{krawczyk2016learning, lin2017focal, cui2019class} in machine learning, where the goal is to ensure robust generalization across subgroups or to improve the performance on the worst-performing group. One common framework to encompass the above studies is the subpopulation shift~\cite{koh2021wilds, yang2023change}, which refers to discrepancies between the subgroup proportions in the training and test data. In many cases, the evaluation focuses on the worst-group performance~\cite{sagawa2019distributionally} or the overall performance on balanced test data~\cite{yang2023change}, where models trained and evaluated on data with subpopulation shifts are likely to fail~\cite{koh2021wilds}.

\begin{figure*}[t!]
    \centering
    \begin{subfigure}{0.23\linewidth}
        \includegraphics[width=\linewidth]{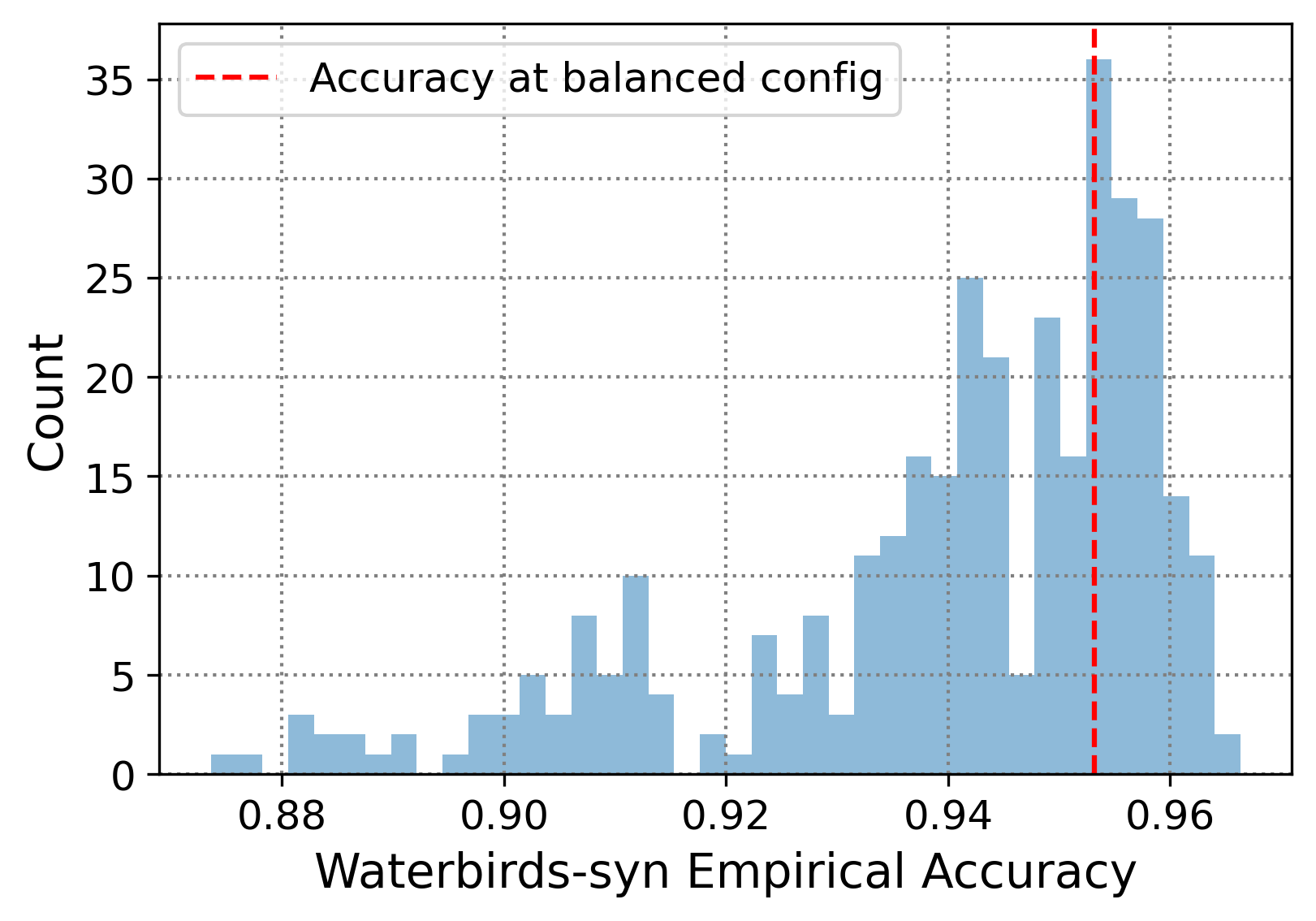}
        \caption{Optimal Accuracy.}
    \end{subfigure}
    \hfill
    \begin{subfigure}{0.53\linewidth}
        \includegraphics[width=\linewidth]{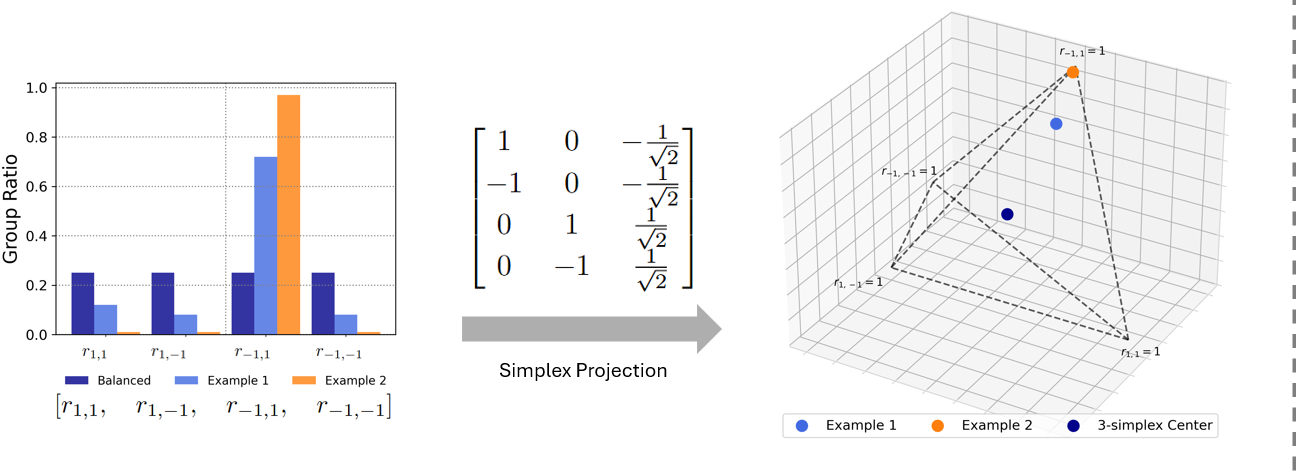}
        \caption{Projection of Group Sizes to Coordinates in 3-simplex.}
        \vspace{0.1mm}
    \end{subfigure}
    \hfill
    \begin{subfigure}{0.22\linewidth}
        \vspace{0.1mm}
        \includegraphics[width=\linewidth]{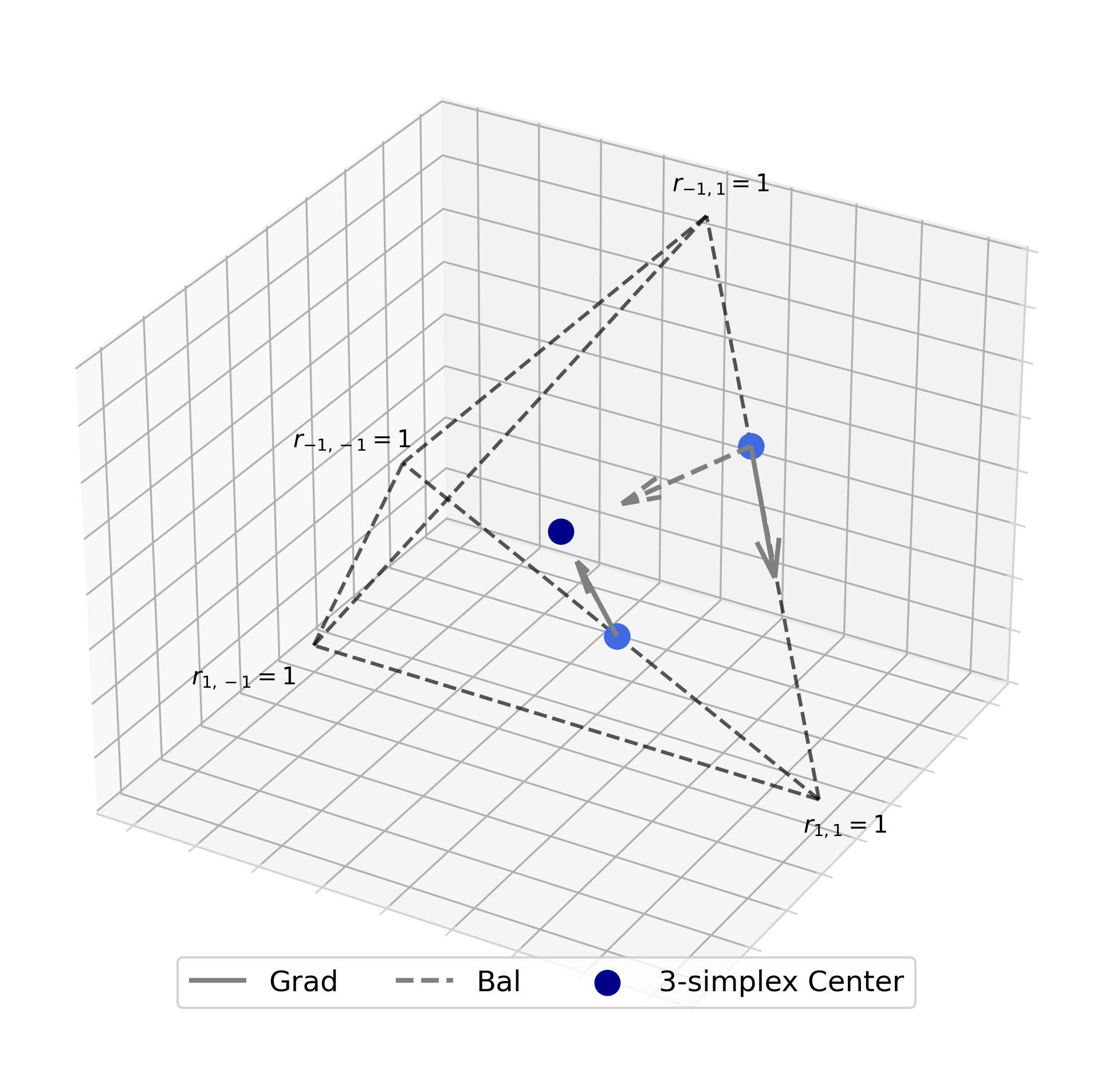}
        \caption{Estimated Directions.}
    \end{subfigure}
    \caption{{\bf} (a) Test accuracy varies over diverse training data configurations and many match or exceed the performance of the balanced configuration. (b) Data configurations visualized by projecting group sizes onto a 3-simplex, including the configurations for the balanced (center), the highly skewed (vertex), and a random configuration (unique coordinate). (c) Visualization of directions estimated according to the gradients (Grad) and balancing (Bal) towards the center.}
    \label{fig:teaser}
    \vspace{-3mm}
\end{figure*}

    A widely acknowledged strategy to ensure generalization against subpopulation shifts is to use fully balanced data to allow the models to be trained sufficiently on each subgroup to avoid potential bias towards specific groups~\cite{sagawa2019distributionally, idrissi2022simple, deng2023robust}. However, collecting fully balanced data is often prohibitively expensive or infeasible in real-world settings~\cite{koh2021wilds}. Many real-world datasets are inherently imbalanced~\cite{koh2021wilds}. The Waterbirds dataset~\cite{sagawa2019distributionally}, designed to classify bird species, exhibits a strong correlation between the bird species and backgrounds ({\emph e.g.} waterbirds appear more often in water backgrounds rather than land). Or in skin cancer diagnosis, benign cases greatly predominate over the malignant ones among all data~\cite{codella2019skin, wu2023discover}. Therefore, attempting to balance the datasets might require significant efforts and costs. 

    Motivated by the challenges of balancing subgroups, we ask a fundamental question ({\bf Optimality}): Is fully balanced data the only optimal configuration of training data for robust generalization under subpopulation shifts? We show the accuracy obtained from diverse configurations with variants of Waterbirds in Figure~\ref {fig:teaser} (a), where we find that many configurations for the training subgroup distribution can achieve test accuracy comparable to or even exceeding that of perfectly balanced training data.
    Building on this fact, we ask whether it is always most efficient to improve the data towards full balance, or whether there exists a shortcut that does not necessarily lead to balance, as the question of {\bf Divergence}. Prior work~\cite{wang2024effect} contributed theoretical solutions to the test accuracy when models are trained on data with spurious correlation. Following the framework proposed by~\cite{wang2024effect}, we examine the test accuracy and use the derivative of the theoretical test accuracy, which characterizes the direction that yields the greatest performance improvement.
    Considering the direction towards balanced subgroups as an oracle direction, we compare their similarity and find that many training configurations yield a different derived direction. This observation reveals that the derived directions might offer new paths other than balancing direction towards robust performance and suggests the potential of strategically adjusting the data distribution for better robustness ({\bf Efficiency}). In practice, this can guide data collection with principled allocation strategies to ensure the robustness of downstream model training.
    
    To validate our findings, we conducted a series of controlled experiments on synthetic and standard datasets, covering images and text. We varied the training data configurations systematically to compare the directions by measuring performance gains obtained along different directions, which shows the advantage of the derived directions on specific configurations and validates the Divergence question. Secondly, for those training configurations where directions differ, we conducted experiments that allocate data gradually along various directions, to demonstrate the potential application to dataset collection, which corresponds to the answer to Efficiency. Lastly, we confirm the Optimality by showing the existence of multiple configurations that achieve comparable optimal performance across datasets. Collectively, our empirical analysis demonstrates that our theoretical insights consistently apply to practical settings. Moreover, we demonstrate the potential applications of our method by integrating the allocation with resampling and last-layer retraining methods.
    Our contributions can be summarized as follows:

    1. We characterize the performance landscape by extending a prior Gaussian modeling framework and identify imbalanced configurations that perform comparably to balance.

    2. We derive principled directions for adjusting subgroup configurations to improve robustness more efficiently than enforcing full balance.
    
    3. We empirically validate our theoretical insights and show the potential in data allocation with extensive experiments.

\section{Related Work}\label{sec:related}

    \paragraph{Subpopulation Shifts}
    Subpopulation shift refers to changes in subgroup proportions between training and test distributions. The WILDS benchmark~\cite{koh2021wilds} introduced real-world datasets for evaluating robustness under distribution shifts, categorizing them into domain shift and subpopulation shift. Yang~\etal~\cite{yang2023change} further formalized this setting by unifying related benchmarks and methods under a common framework. A large body of work in subpopulation shifts has sought to improve robustness through model-side interventions, including group-robust training~\cite{sagawa2019distributionally, duchi2021learning}, feature regularization~\cite{zhang2022correct, wang2024effect, lu2024neural, wen2025elastic}, etc. These methods primarily address subpopulation shift by modifying the learning algorithm or training objective. In contrast, our work focuses on the data side of subpopulation robustness: we study how the subgroup proportions in the training data affect downstream performance and utilize it to improve the robustness.

    \mypar{Subgroup Allocation.}
    Closer to our work, prior studies have examined subgroup allocation itself and shown that balancing is not universally optimal. \cite{rolf2021representation} studies subgroup allocation through empirical scaling laws to show that the allocation optimizing a given objective does not need to be balanced and proposes that subgroup composition is a key determinant of performance. \cite{schrouff2024mind} provides an analysis of causal conditions on when balancing succeeds or fails as a fairness intervention. \cite{alloula2025representation} examines when subgroup balance matters in subgroup allocation, and proposes a sensitivity explanation based on latent separation. \cite{claucich2025fairness} argues that the effect of subgroup allocation depends on subgroup separation in pretrained latent representations, offering a representation-level explanation for balancing failure. Our work is most closely related to \cite{rolf2021representation, alloula2025representation} in the setup of analyzing performance for data allocation. Unlike the pilot-training sweeps of \cite{rolf2021representation}, which rely on extensive empirical exploration, our approach determines the allocation direction analytically from estimated feature separability at negligible cost.

    \mypar{Gaussian Data Modeling}
    Recent studies~\cite{montanari2019generalization, mei2021generalization} have employed Gaussian data modeling or linear approximations for theoretical derivations to understand phenomena such as benign overfitting and double descent. Although Gaussian data modeling is idealized, these works showed that the derived key insights generalize beyond Gaussian assumptions. Similar Gaussian modeling has also been utilized in the analysis of distribution shifts, such as spurious correlation~\cite{nagarajan2021understanding, wang2024effect}, class imbalance~\cite{liu2022self}, or out-of-distribution detection~\cite{ming2022impact, yao2022improving}. Our analysis builds directly on \cite{wang2024effect}, whose symmetric spurious-correlation setting is a special case of the more general subgroup configuration. We extend this framework to characterize the allocation landscape and analyze how performance changes as subgroup proportions vary independently.

    \mypar{Balance-based Robustness Methods.}
    Subgroup balance is commonly used in robustness methods such as resampling and last-layer retraining. Resampling methods over-sample minority groups or under-sample majority groups to construct a balanced training distribution, providing a simple and effective baseline for group robustness~\cite{idrissi2022simple}. Last-layer retraining methods instead fit or reweight a classifier on a held-out subset, which is often balanced or constructed using limited group annotations~\cite{kirichenko2023last,izmailov2022feature,labonte2023towards,qiu2023simple}. Group-robust sample reweighting further learns per-example weights with influence functions and observes that exact group balance can be suboptimal~\cite{qiao2025group}. Our allocation guidance complements both families, that Grad replaces balance as the target subgroup distribution for resampling or stage-two retraining.

\section{Subpopulation Robustness across Configurations}

\subsection{Data Modeling}

    Following prior theoretical studies of spurious correlation and subpopulation shifts~\citep{sagawa2020investigation,yao2022improving,idrissi2022simple,ming2022impact,liu2022self,wang2024effect},
    we consider binary classification over a dataset $\gD=\gX\times\gY$, where $\gX$ is the input space and $\gY=\{-1,1\}$ is the label space. Each example $\vx\in\gX$ has a label $y\in\{-1,1\}$ and is associated with $N$ binary
    attributes
        $\va=(a_1,\ldots,a_N)\in\gA=\{\pm1\}^N $.
    Without loss of generality, we let $a_1$ be the invariant attribute, which is always aligned with the label, i.e., $a_1\equiv y$. The remaining attributes $a_2,\ldots,a_N$ may be correlated with the label.
    Our Gaussian feature model and subsequent Bayesian-performance analysis build on~\cite{wang2024effect}. Although the notation of an invariant attribute is motivated by spurious correlation~\cite{wang2024effect}, we extend this scope to model different forms of subpopulation shift through the training subgroup configuration. We provide a detailed comparison with related work in Appendix~\ref{appd:related}.
    
    \textbf{Feature Modeling.} We represent each input $\vx$ by a feature vector $\vz=(\vz_1,\ldots,\vz_N)\in\gZ$, where each component $\vz_n$ corresponds to attribute $a_n$. Similar to \cite{wang2024effect}, we model each feature component as a Gaussian variable whose mean is determined by the corresponding
    attribute:
    \[
        \vz_n \mid a_n \sim \mathcal{N}(a_n\vmu_n,\Sigma_n).
    \]
    Equivalently, since $a_n\in\{\pm1\}$, the feature $\vz_n$ is centered at $\vmu_n$ when $a_n=1$ and at $-\vmu_n$ when $a_n=-1$. Here, $\vmu_n$ and $\Sigma_n$ denote the mean direction and covariance matrix of the $n$-th feature component.
    
    For a subgroup specified by attributes $\va$, the full feature vector follows a Gaussian distribution with mean and block-diagonal covariance 
    \[
        \vmu_\va = [a_1\vmu_1^\top,\ldots,a_N\vmu_N^\top]^\top \in \sR^d, \quad \Sigma=\operatorname{diag}(\Sigma_1,\ldots,\Sigma_N)\in\sH^{d\times d}_+,
    \]
    where $d=\sum_{n=1}^N d_n$ and $d_n$ is the dimension of $\vz_n$. Thus, each subgroup corresponds to one component in a Gaussian mixture over the feature space. We measure the separability of the $n$-th attribute by the Mahalanobis
    distance $m_n=\vmu_n^\top\Sigma_n^{-1}\vmu_n$.
    A larger $m_n$ indicates that the corresponding attribute is easier to classify from the feature representation.
    
    \textbf{Subpopulation modeling.} Subpopulation shift is modeled through changes in the subgroup proportions between training and test data. The $N$ binary attributes define $2^N$
    subgroups. We denote the training proportion of subgroup $\va$ by $r_\va$, with $r_\va\ge 0$ and $\sum_{\va\in\gA} r_\va=1$. For simplicity, we study models trained under arbitrary subgroup configurations $\{r_\va\}_{\va\in\gA}$
    and evaluate them on balanced test data throughout the analysis.
    For theoretical derivation, it is useful to reparameterize the subgroup proportions $r_\va$ using independent variables. Let
    \[
        \beta = \sP(Y=1), 
        \quad 
        \alpha_{\sign(y)}^{(n)} = \sP(a_n = y \mid Y = y)
    \]
    denote the positive-label proportion and the probability that attribute $a_n$ is aligned with the label within examples of label $y$. $\valpha$ further denotes all the attribute-label alignment probabilities. Since $a_1$ is the invariant attribute, we have $\alpha_+^{(1)}=\alpha_-^{(1)}=1$. This parameterization represents the $2^N$ subgroup proportions using $2N-1$ free variables, which allows a tractable subsequent theoretical analysis.

    \subsection{Analysis}\label{sec:grad}
    Given a linear classifier 
        $\hat{y}=\mathrm{sign}(\vw^\top \vz+b)$,
    we first characterize its accuracy on each subgroup. The training distribution is specified by the feature parameters $(\vmu,\Sigma)$ and the subgroup-configuration parameters $(\valpha,\beta)$. These parameters determine the Bayesian optimal
    linear classifier, which we then evaluate on balanced test data. The following lemma extends the symmetric group-accuracy expression of~\cite{wang2024effect} to arbitrary subgroup configurations.
    
    \vspace{3pt}
    \noindent\textbf{Lemma 1. (Bayesian Group Accuracy)}
    \textit{
    For a subgroup specified by attributes $\va\in\{\pm1\}^N$, the group accuracy of
    a linear classifier $(\vw,b)$ is
    \begin{align}
        Acc_{\va}(\vw,b)
        =
        \frac{1}{2}
        \left(
        1+
        \operatorname{erf}
        \left(
        \frac{\vmu_\va^\top \vw+a_1b}
        {\sqrt{2\vw^\top\Sigma\vw}}
        \right)
        \right)
    \end{align}
    where $\vmu_\va=[a_1\vmu_1^\top,\ldots,a_N\vmu_N^\top]^\top$ is the subgroup mean.
    }
    
    Following~\cite{wang2024effect}, we obtain the Bayesian optimal linear classifier by maximizing the overall accuracy on the training distribution. Under the Gaussian feature model, the optimal classifier is aligned with the feature directions:
        $\vw_n^* = \eta_n \Sigma_n^{-1}\vmu_n$,
    where each $\eta_n\in\sR$ is a scalar coefficient for the $n$-th feature component. The coefficients $\eta_1,\ldots,\eta_N$ and the bias $b$ are determined by the training statistics $(\vmu,\Sigma,\valpha,\beta)$ through the optimality conditions of the training accuracy. We provide the full derivation and the resulting equation system in the appendix. Once $(\vw^*,b^*)$ is solved, Lemma~1 gives the theoretical test accuracy of the classifier on each subgroup.
    
    The key implication is that the test performance can be solved and expressed as a function of the training subgroup configuration. However, as the number of classifier parameters and subgroup-configuration variables grows with the number of attributes, we focus the main analysis on the two-attribute setting for clarity. We also provide a demonstration of three attributes in Appendix~\ref{appd:n3}. Since $a_1\equiv y$, its alignment
    parameters are fixed as $\alpha_+^{(1)}=\alpha_-^{(1)}=1$ and can be omitted.
    Thus, the training configuration is fully described by
    $(\alpha_+,\alpha_-,\beta)
        :=
        (\alpha_+^{(2)},\alpha_-^{(2)},\beta).$
    The same procedure extends to more attributes, but with additional intermediate variables.
    
    Finally, we define the performance metric used in our theoretical analysis. Following prior work on subpopulation shifts~\cite{yang2023change}, two common
    metrics are worst-group accuracy,
        $\mathrm{WGA} = \min_{\va\in\gA} Acc_\va$,
    and adjusted accuracy,
        $\mathrm{AA} =
        \frac{1}{2^N}\sum_{\va\in\gA} Acc_\va$,
    which equals the overall accuracy on a balanced test distribution. Although WGA is commonly used, its minimum operator is non-smooth and makes gradient-based analysis less direct. We therefore use adjusted accuracy for the theoretical derivation, since it is a smooth average of subgroup accuracies and enables the subsequent analysis. We empirically report both
    adjusted accuracy and WGA to verify that the theoretical conclusions also hold under worst-group robustness.

    The same group-wise derivatives also provide a local WGA direction. While WGA $\nabla_{\vp}\mathrm{WGA}=\nabla_{\vp}Acc_{\va^*}$ employs the minimum operator, it is non-smooth and becomes non-differentiable when the worst-performing group shifts, making a global gradient-based analysis inapplicable. In our experiments, AA- and WGA-derived directions yield similar allocation behavior, with details provided in Appendix~\ref{appd:wga-direction}. We retain AA as the main objective and verify its empirical effect on WGA.

    \mypar{Optimality: Configurations that achieve robust performance.}
    We first ask whether the fully balanced configuration is the only training configuration that achieves optimal robust performance. To study this question, we specialize to the two-attribute case (\(N=2\)), where the Bayesian optimal classifier can be parameterized more simply.    
    Let \(\tau = \eta_2 / \eta_1\), where \(\eta_1\) and \(\eta_2\) are the coefficients of the two feature components in the Bayesian optimal classifier.
    Then the classifier is fully determined by the two scalars \((\tau,b)\), and the group accuracy for subgroup \(\va=(a_1,a_2)\) thus becomes
    \begin{align}
        Acc_{\va}
        =
        \frac{1}{2}
        \left(
        1+\operatorname{erf}
        \left(
        \frac{a_1m_1+\tau a_2m_2+a_1b}
        {\sqrt{2(m_1+\tau^2m_2)}}
        \right)
        \right)
        \label{eq:acc}
    \end{align}
    where $m_1, m_2$ represent the separability for the two attributes.
    Equation~\ref{eq:acc} shows that group accuracy is a continuous function of the training configuration through \((\tau,b)\). As a result, adjusted accuracy and worst-group accuracy are also continuous in the configuration. This suggests that the balanced configuration is not an isolated optimum: configurations near a high-performing point may also achieve comparably strong performance. To examine this possibility systematically, we study the performance landscape over a wide range of training configurations.
    
    \textit{Visualization notation.} 
    In the two-attribute setting, there are four subgroups. A training configuration is therefore specified by four subgroup proportions \(\{r_{\va}\}_{\va\in\{\pm1\}^2}\), which sum to one. To visualize all possible configurations, we map these four proportions to a 3-simplex, as shown in
    Figure~\ref{fig:teaser}(b). Each point in the simplex represents one training configuration. The center corresponds to the balanced configuration, where each subgroup has proportion \(25\%\). Each vertex corresponds to an extreme
    configuration, where all data belong to a single subgroup. Points closer to a vertex assign more data to that subgroup.
    
    Figure~\ref{fig:teaser}(b) illustrates this mapping with three examples: the balanced configuration at the center, an extreme configuration with \(r_{-1,1}=1\) at one vertex, and an intermediate configuration where \(r_{-1,1}\) is the largest subgroup. {\it This 3-simplex visualization provides a direct and complete view: each point inside the simplex represents one valid set of subgroup proportions, and every possible configuration maps to a unique point within the simplex.}
    
    To study the performance landscape, we uniformly sample the configuration parameters \(\alpha_+,\alpha_-,\beta\). Using \(n\) equally spaced values for each parameter gives \(n^3\) training configurations. For example, with \(n=49\), we obtain \(49^3\) configurations. For each configuration, we compute the theoretical adjusted accuracy using \(m_1=m_2=2\), which represents a moderate level of feature separability.
    
\begin{figure}[t!]
    \centering

    \begin{minipage}{0.48\linewidth}
        \centering
        \begin{subfigure}{0.46\linewidth}
            \includegraphics[width=\linewidth]{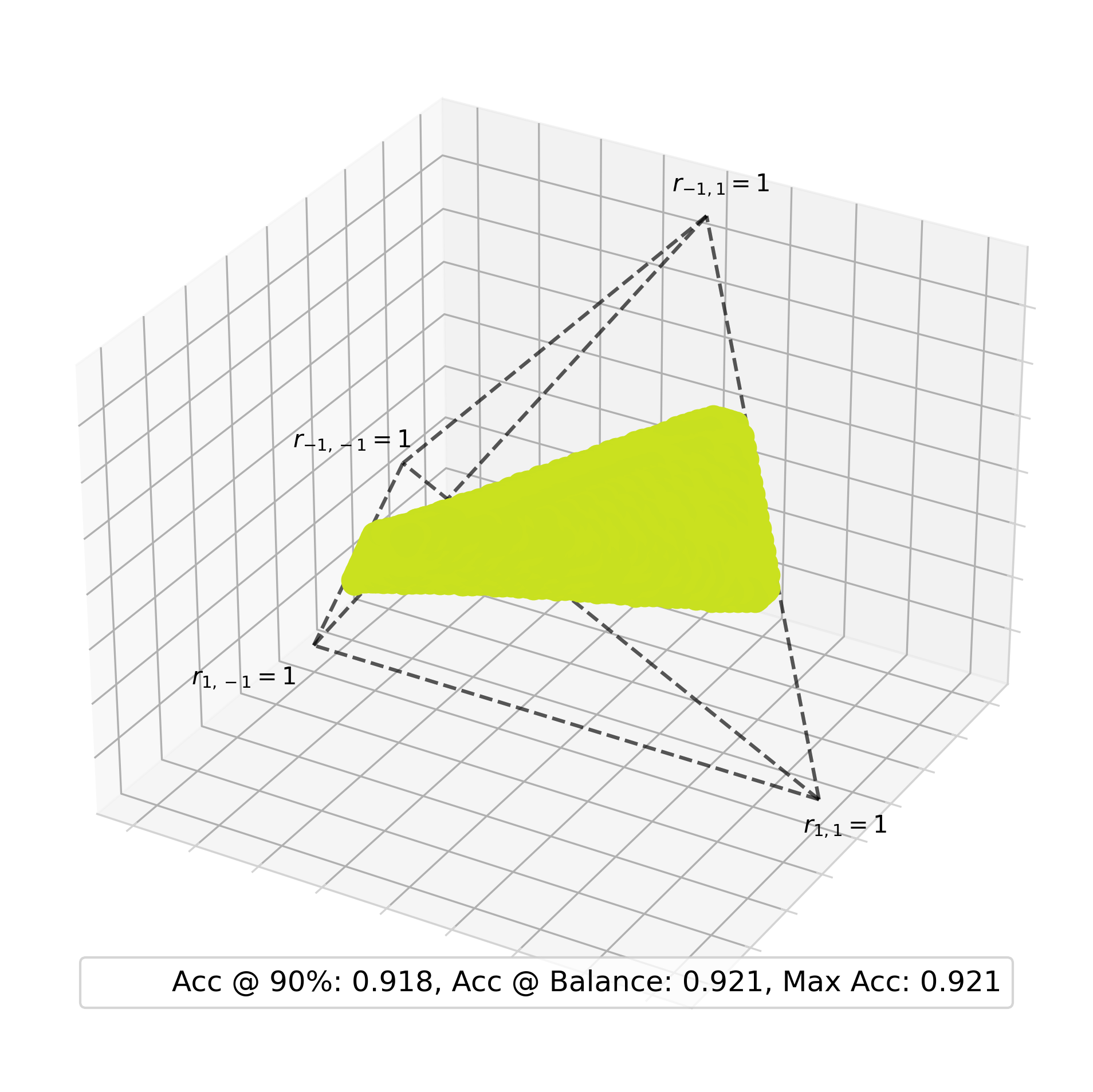}
        \end{subfigure}
        \hfill
        \begin{subfigure}{0.46\linewidth}
            \includegraphics[width=\linewidth]{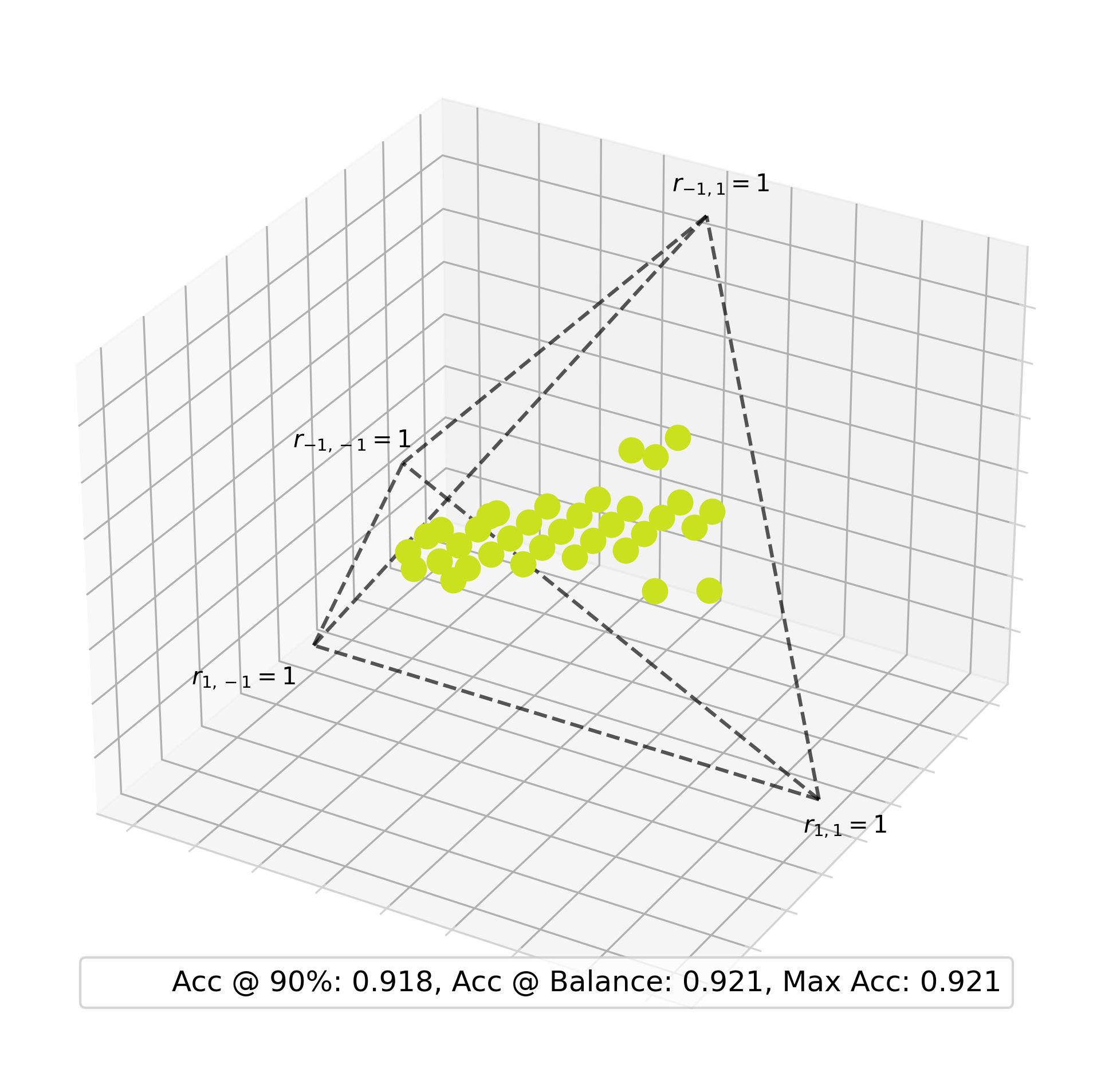}
        \end{subfigure}

        \caption{Visualization of group configurations achieving 90th-percentile accuracy, shown on a densely sampled (left, $n=49$) and a sparsely sampled (right, $n=7$) theoretical 3-simplex.}
        \label{fig:max_acc_thr}
    \end{minipage}
    \hfill
    \begin{minipage}{0.48\linewidth}
        \centering
        \begin{subfigure}{0.46\linewidth}
            \includegraphics[width=\linewidth]{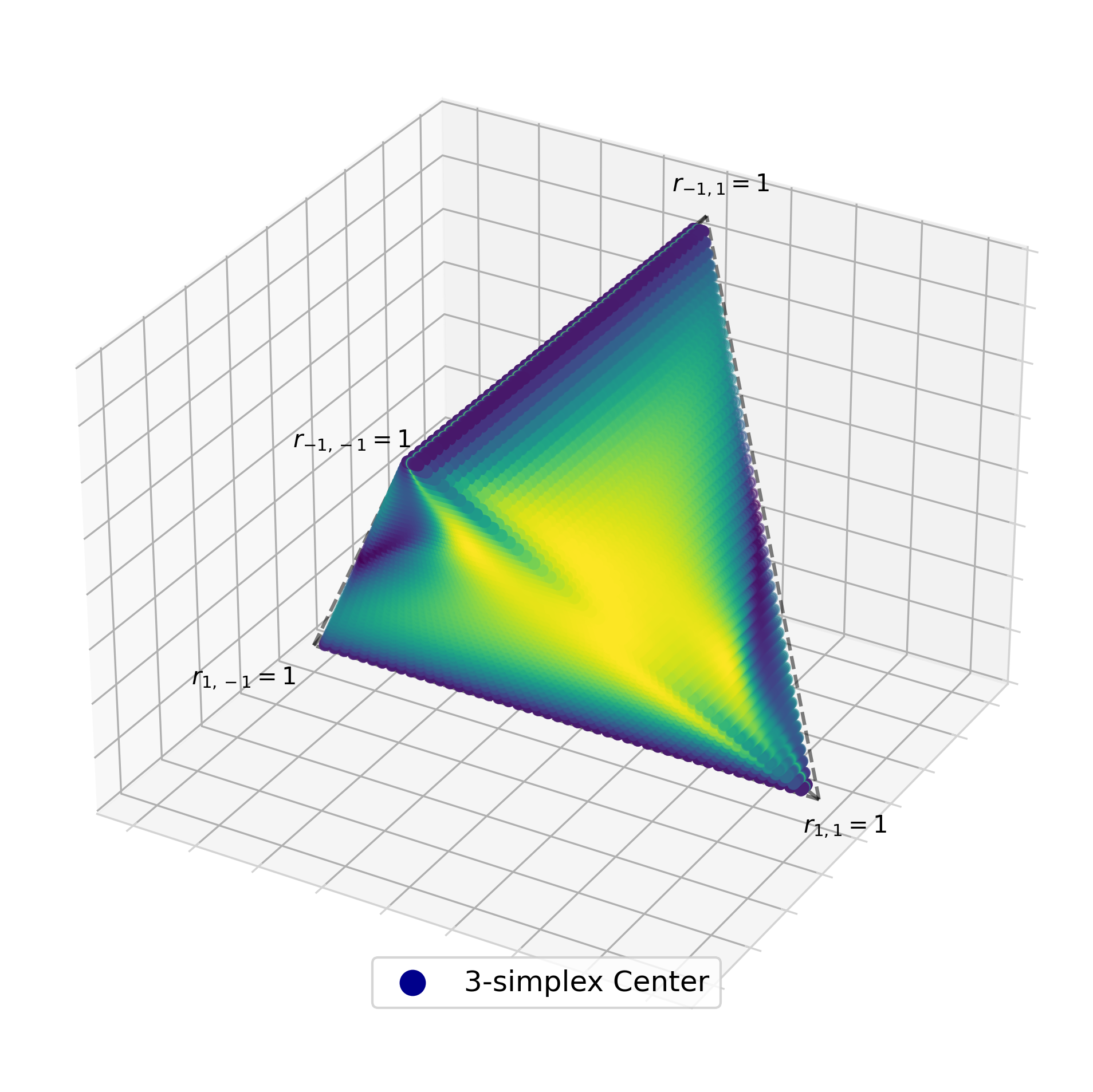}
        \end{subfigure}
        \hfill
        \begin{subfigure}{0.52\linewidth}
            \includegraphics[width=\linewidth]{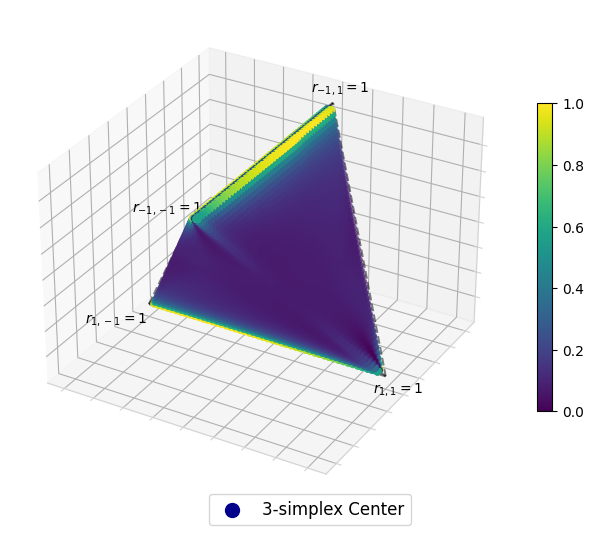}
        \end{subfigure}

        \caption{Visualization of the Grad and Bal direction comparison. Darker regions in theoretical cosine similarity (left) and lighter regions in accuracy gap (right) indicate greater disparity.}
        \label{fig:sim}
    \end{minipage}

    \vspace{-2mm}
\end{figure}

    Figure~\ref{fig:max_acc_thr} visualizes the configurations whose adjusted accuracy lies above the 90th percentile. For both \(n=49\) and \(n=7\), the balanced configuration achieves adjusted accuracy 0.921, while the 90th percentile threshold is 0.918. Thus, many configurations attain performance very close to the balanced one, indicating the existence of alternative high-performing, near-optimal configurations. Using a larger \(n\) produces a denser sampling of the simplex and therefore a smoother visualization of this high-performance region.

    \mypar{Efficiency: Efficient routes toward robust performance.}
    The optimality analysis shows that high-performing configurations are not limited to the fully balanced configuration. We next ask how to move from a given training configuration toward better robust performance as efficiently as
    possible.
    Let
        $A(\alpha_+,\alpha_-,\beta)
        =
        \frac{1}{4}\sum_{\va\in\mathcal{A}} Acc_{\va}$
    denote the adjusted accuracy in the two-attribute setting. Since \(A\) is a differentiable function of the training-configuration parameters \((\alpha_+,\alpha_-,\beta)\), its gradient indicates the local direction in which the configuration should be changed to increase adjusted accuracy most rapidly.
    
    \vspace{3pt}
    \noindent\textbf{Proposition 1 (Gradient of Adjusted Accuracy).}
    \textit{
    Consider the two-attribute setting with configuration parameters $\vp=(\alpha_+,\alpha_-,\beta)$.
    For each subgroup \(\va=(a_1,a_2)\in\{\pm1\}^2\), define
    \[
        f_{\va}(\tau,b)
        =
        \frac{a_1m_1+\tau a_2m_2+a_1b}
        {\sqrt{2D}},
        \qquad
        D=m_1+\tau^2m_2,
    \]
    where \(\tau\) and \(b\) are determined by the Bayesian optimality equations and
    therefore depend on \(\vp\). Then, for any
    \(p\in\{\alpha_+,\alpha_-,\beta\}\), the derivative of adjusted accuracy is
    \begin{equation}
    \begin{aligned}
        \frac{\partial A}{\partial p}
        &=
        \frac{1}{4}\sum_{\va\in\mathcal{A}}
        \frac{\partial Acc_{\va}}{\partial p} =
        \frac{1}{4}\sum_{\va\in\mathcal{A}}
        \frac{1}{\sqrt{\pi}}
        e^{-f_{\va}^{2}}
        \left(
            \frac{
            m_2\big(a_2m_1-\tau a_1(m_1+b)\big)
            }
            {D\sqrt{2D}}
            \frac{\partial \tau}{\partial p}
            +
            \frac{a_1}{\sqrt{2D}}
            \frac{\partial b}{\partial p}
        \right).
    \end{aligned}
    \label{eq:grad_adjusted_acc}
    \end{equation}
    The quantities \(\partial \tau/\partial p\) and \(\partial b/\partial p\) are
    obtained by differentiating the Bayesian optimality equations that define
    \(\tau\) and \(b\).
    }
    \vspace{3pt}
    
    The full derivation is provided in Appendix~\ref{appd:proposition1}. The takeaway is that a principled direction is solvable for changing the training subgroup configuration.
    Specifically, we define the gradient direction as
    \[
        d_{\mathrm{Grad}}
        =
        \frac{\nabla_{\vp} A}{\|\nabla_{\vp} A\|_2}, \quad \nabla_{\vp}A
        =
        \left(
        \frac{\partial A}{\partial \alpha_+},
        \frac{\partial A}{\partial \alpha_-},
        \frac{\partial A}{\partial \beta}
        \right)
    \]
    Starting from an initial configuration \(\vp^{(0)}\), we update the configuration by
        $\vp^{(t+1)}
        =
        \vp^{(t)}
        +
        \delta d_{\mathrm{Grad}}^{(t)}$,
    where \(\delta>0\) is the step size. This update performs gradient ascent on the theoretically estimated adjusted accuracy.

    \mypar{Baseline and robust directions.}
    We compare with two additional directions. The first is the balancing direction (Bal), which serves as an oracle baseline and always updates the data from the current configuration toward the fully balanced configuration.
    The second is the sign-gradient direction (Sign). Grad depends on the feature separability values $m_n$, which are known in the Gaussian model but must be estimated for real data. Since such estimates can be noisy or inaccurate, we introduce Sign, which uses only the signs of the gradient entries.
    This provides a more robust alternative when the separability estimates are imprecise. We include more details of implementing these directions in Appendix~\ref{appd:implementation}.
    
    \mypar{Divergence between Grad and Bal.}
    We next ask when Grad and Bal point in different directions. For each training configuration, we compute the cosine similarity between the two directions and visualize it in Figure~\ref{fig:sim} (left), where low similarity means that balancing may not be the most efficient update. We observe that while many configurations have reasonably aligned directions, some edge regions of the simplex show strong disagreement between Grad and Bal.
    
    To quantify whether this directional disagreement leads to different performance gains, we also compare the accuracy after updating the configuration along each direction. Starting from the same configuration $\vp$, we run 10 update steps along Grad and Bal, obtaining $\vp_{\mathrm{Grad}}$ and $\vp_{\mathrm{Bal}}$.
    We then measure
        $A(\vp_{\mathrm{Grad}}) - A(\vp_{\mathrm{Bal}})$.
    Figure~\ref{fig:sim} (right) shows that regions with low cosine similarity also tend to have larger positive accuracy gaps, indicating that Grad can provide a more efficient route to robust performance than moving directly toward balance.

\section{Experiments}
    \subsection{Experimental Setup}
    
    \paragraph{Goals.}
    Although our theory uses Gaussian feature modeling, the resulting directions only require empirically accessible quantities, such as configurations and estimated feature separability, rather than Gaussian means or covariances. We therefore evaluate whether the theoretical insights transfer to realistic non-Gaussian features learned by deep models. Our experiments address three questions:
    (1) \textbf{Divergence}: when do Grad and Bal point in different directions?
    (2) \textbf{Efficiency}: when they diverge, does Grad provide more efficient guidance for data allocation? and
    (3) \textbf{Optimality}: can non-balanced configurations achieve performance comparable to balanced configurations?
    
    \mypar{Datasets.}
    Following prior work on subpopulation shifts~\cite{yang2023change}, we evaluate on three image datasets, Waterbirds~\cite{sagawa2019distributionally},
    UrbanCars~\cite{li2023whac}, and CelebA~\cite{liu2015deep}, and one text dataset, CivilComments~\cite{borkan2019nuanced}. Although some of the datasets are primarily used for spurious correlation studies, we systematically vary subgroup configurations to validate our insights under a broader scope of subpopulation shifts.
    We construct subsets to create CelebA-subset and CivilComments-subset, and generate flexible synthetic variants of Waterbirds and UrbanCars, denoted as Waterbirds-syn and UrbanCars-syn, using their foreground-background construction
    schemes~\cite{wah2011caltech,krause20133d,zhou2017places}. Dataset details are provided in Appendix~\ref{appd:dataset}.
    
    \mypar{Models.}
    Due to a wide range of configurations to validate our insights, we use one standard pretrained encoder per modality: ResNet50~\cite{he2016deep} for image datasets and BERT~\cite{devlin2019bert} for the text dataset. For each dataset, we attach a one-layer classification head and fine-tune the full model. These choices follow common setups in subpopulation-shift studies ~\cite{yang2023change, li2025let, zheng2025neurontune}. Additional training details are provided in Appendix~\ref{appd:exp_detail}.

    Since our conclusions rely on relative gains between directions rather than absolute performance, they are less tied to a specific encoder. Moreover, as the feature separability is estimated from one set of empirical results, it implicitly reflects the model’s capacity to distinguish attributes, thus supporting the model-agnostic nature of our method. We verify the model-agnostic characteristics with additional encoders in Appendix~\ref{appd:encoder}.
    
    \mypar{Metrics and directions.}
    We report adjusted accuracy (AA), the average accuracy across groups, and worst-group accuracy (WGA), the minimum group accuracy~\cite{yang2023change}. All directions denoted Grad in the main experiments are derived from AA, and we report WGA separately to test whether this guidance also benefits the weakest group in Appendix~\ref{appd:wga-results}.
    
    We compare four update directions for changing subgroup configurations: \textit{Grad}, our derived gradient direction; \textit{Bal}, an oracle baseline that points toward full balance; \textit{Sign}, which uses only the signs of the gradient entries and is more robust to separability-estimation error; and \textit{Random}, a random unit direction used as a neutral control.

\begin{figure*}[t]
    \centering
    \begin{subfigure}{0.24\linewidth}
        \includegraphics[width=\linewidth]{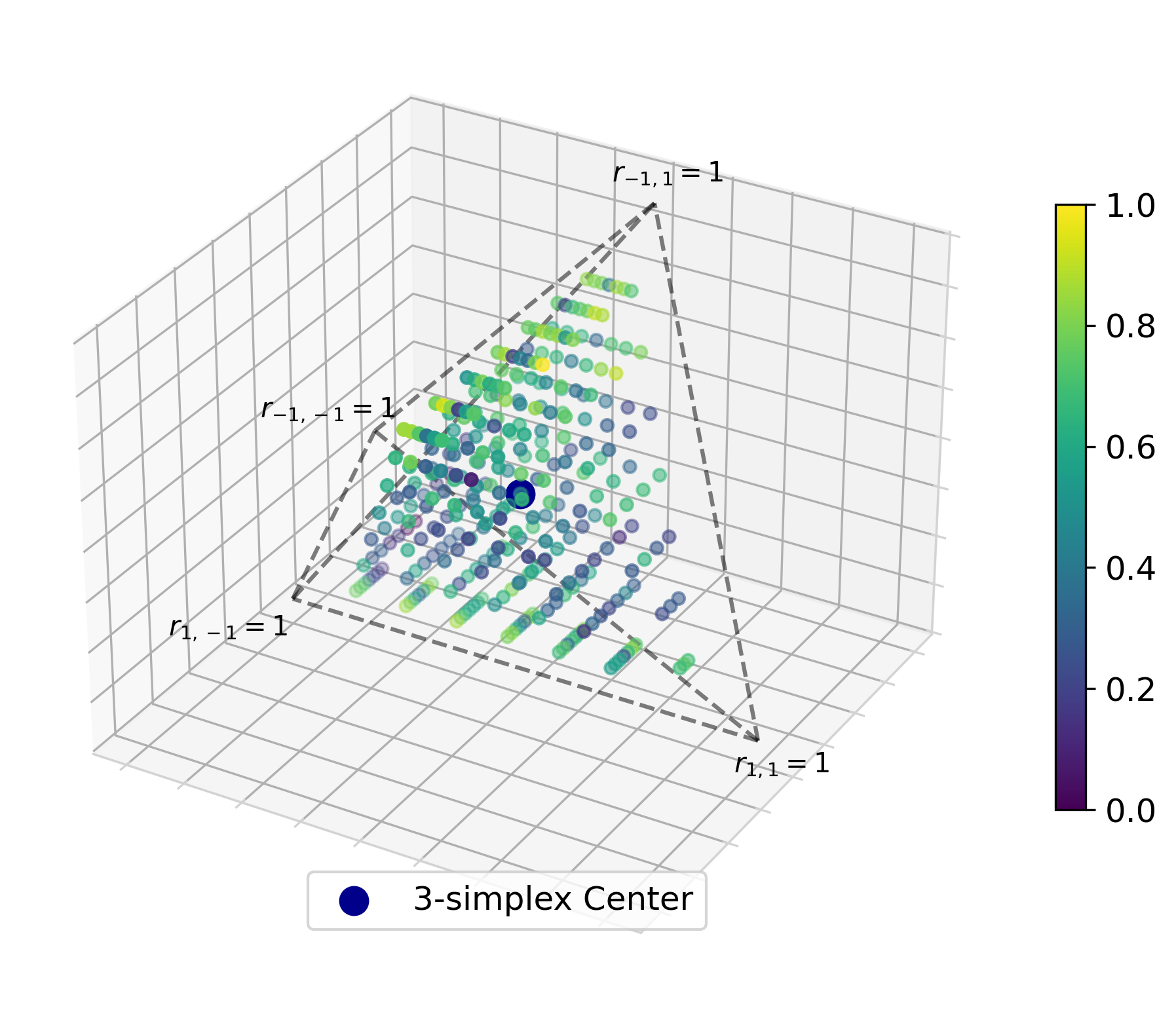}
        \caption{Waterbirds-syn}
    \end{subfigure}
    \hfill
    \begin{subfigure}{0.24\linewidth}
        \includegraphics[width=\linewidth]{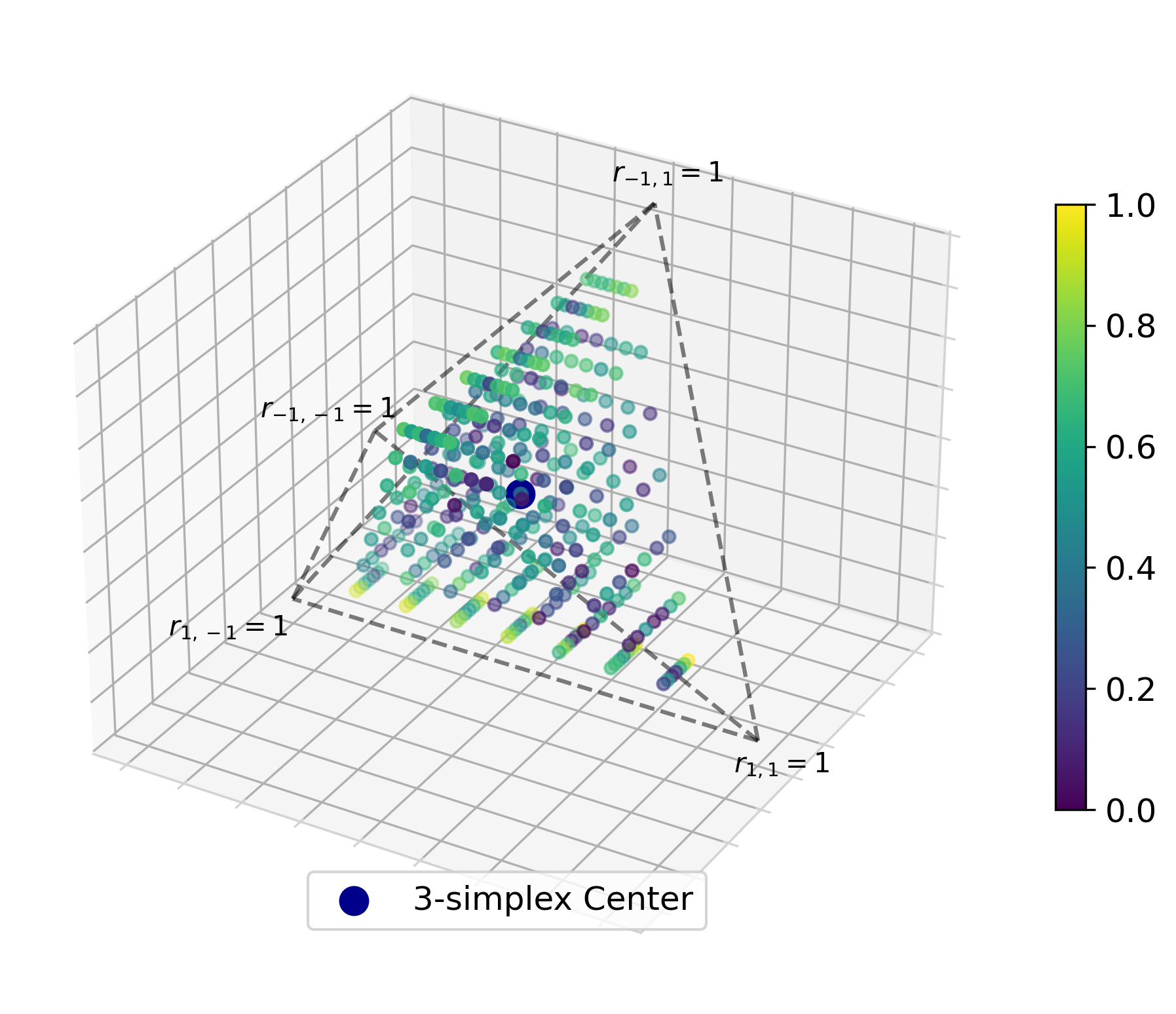}
        \caption{UrbanCars-syn}
    \end{subfigure}
    \hfill
    \begin{subfigure}{0.24\linewidth}
        \includegraphics[width=\linewidth]{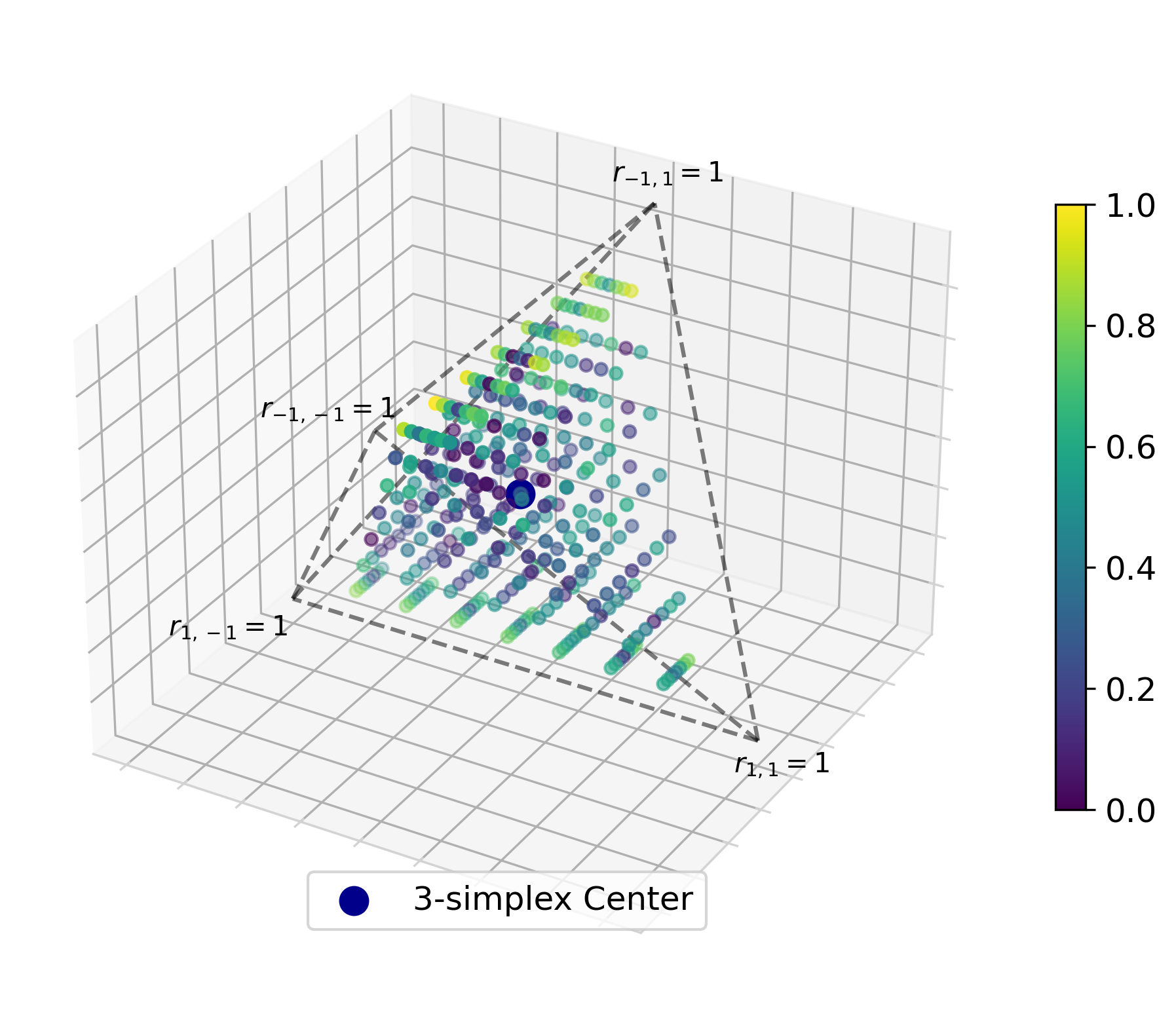}
        \caption{CelebA-subset}
    \end{subfigure}
    \hfill
    \begin{subfigure}{0.24\linewidth}
        \includegraphics[width=\linewidth]{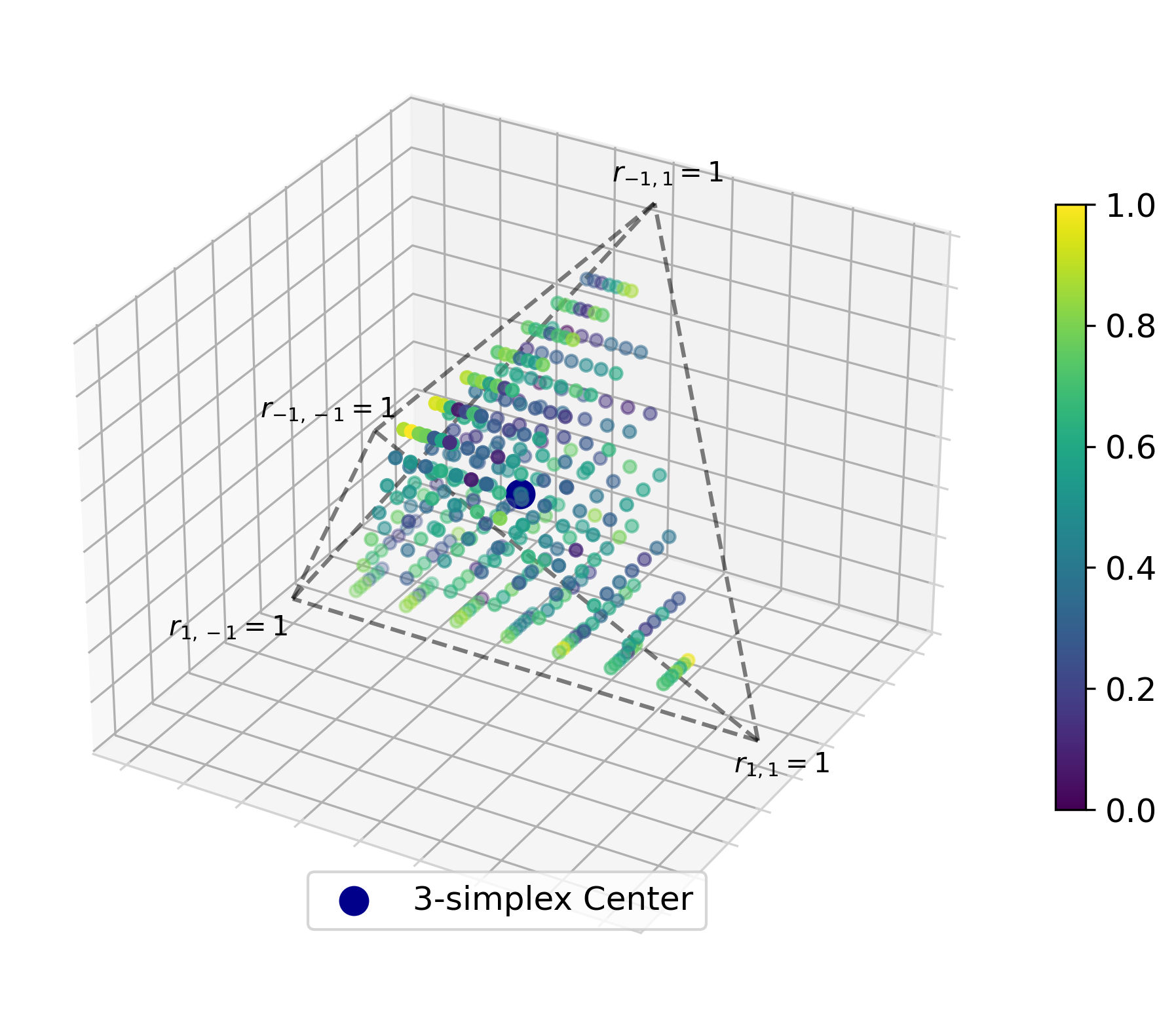}
        \caption{CivilComments-subset}
    \end{subfigure}
    \caption{{\bf Existence of divergent directions across datasets.} Empirical comparison of Grad/Bal directions with accuracy gaps obtained from $7^3=343$ configurations of training data. Lighter colors are observed near edges connecting $r_{-1,-1}$ to $r_{-1, 1}$, and $r_{1,-1}$ to $r_{1, 1}$, indicating a greater disparity between Grad/Bal directions and aligning with the theoretical accuracy gap trends in \cref{fig:sim}.}
    \vspace{-2mm}
\label{fig:emp-sim}
\end{figure*}

\begin{figure*}[h!]
    \centering

    \begin{subfigure}{0.24\linewidth}
        \includegraphics[width=\linewidth]{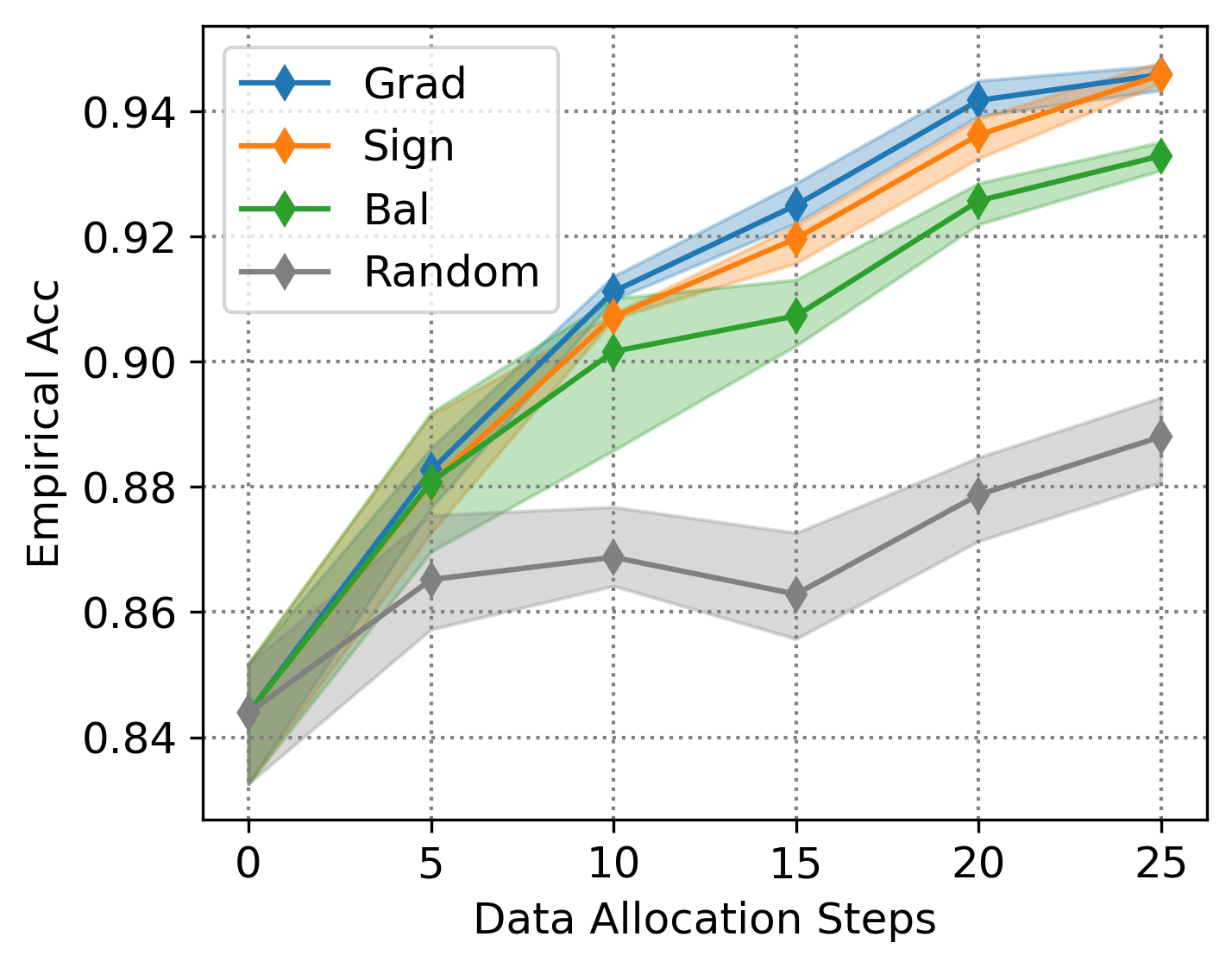}
    \end{subfigure}
    \hfill
    \begin{subfigure}{0.24\linewidth}
        \includegraphics[width=\linewidth]{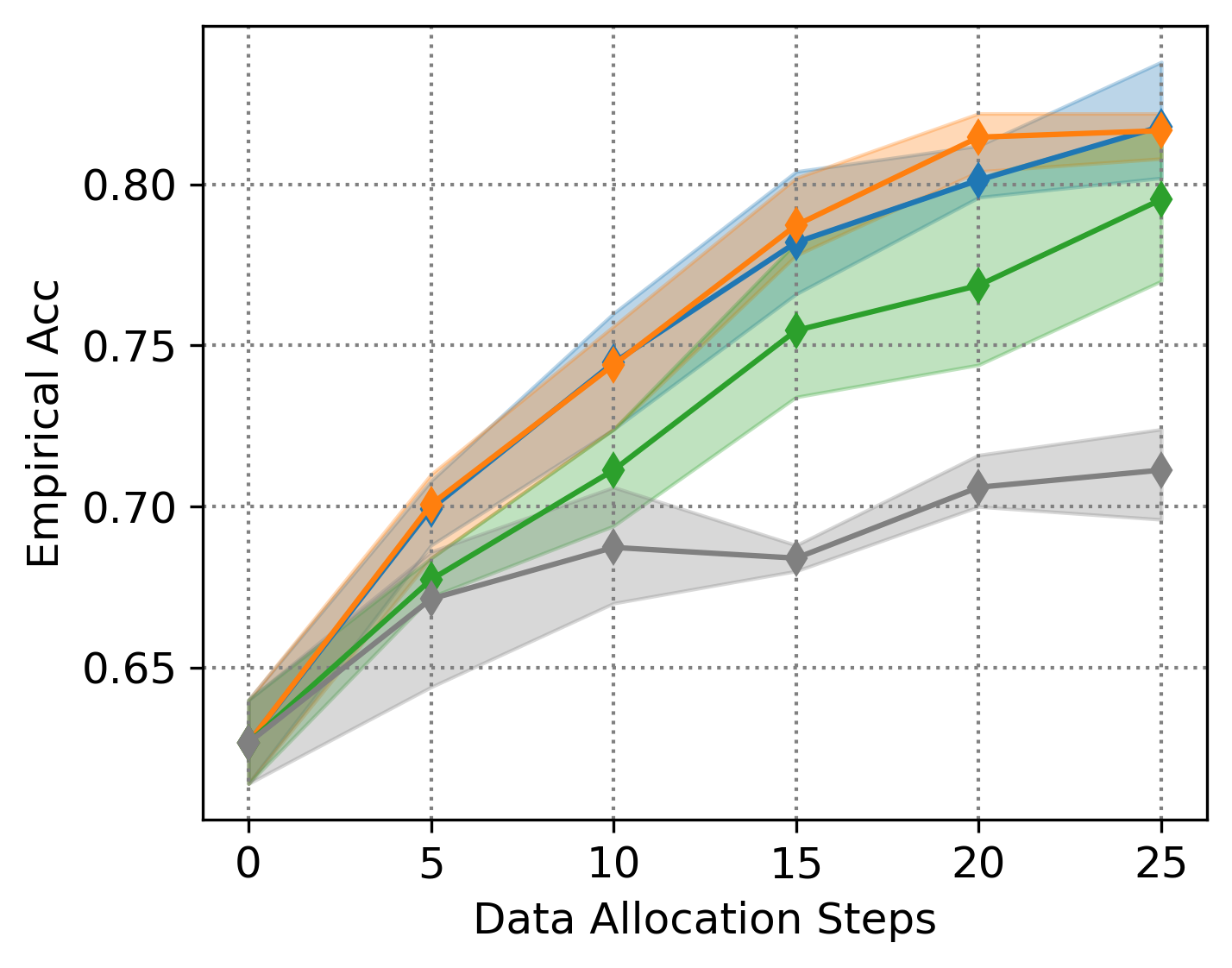}
    \end{subfigure}
    \hfill
    \begin{subfigure}{0.24\linewidth}
        \includegraphics[width=\linewidth]{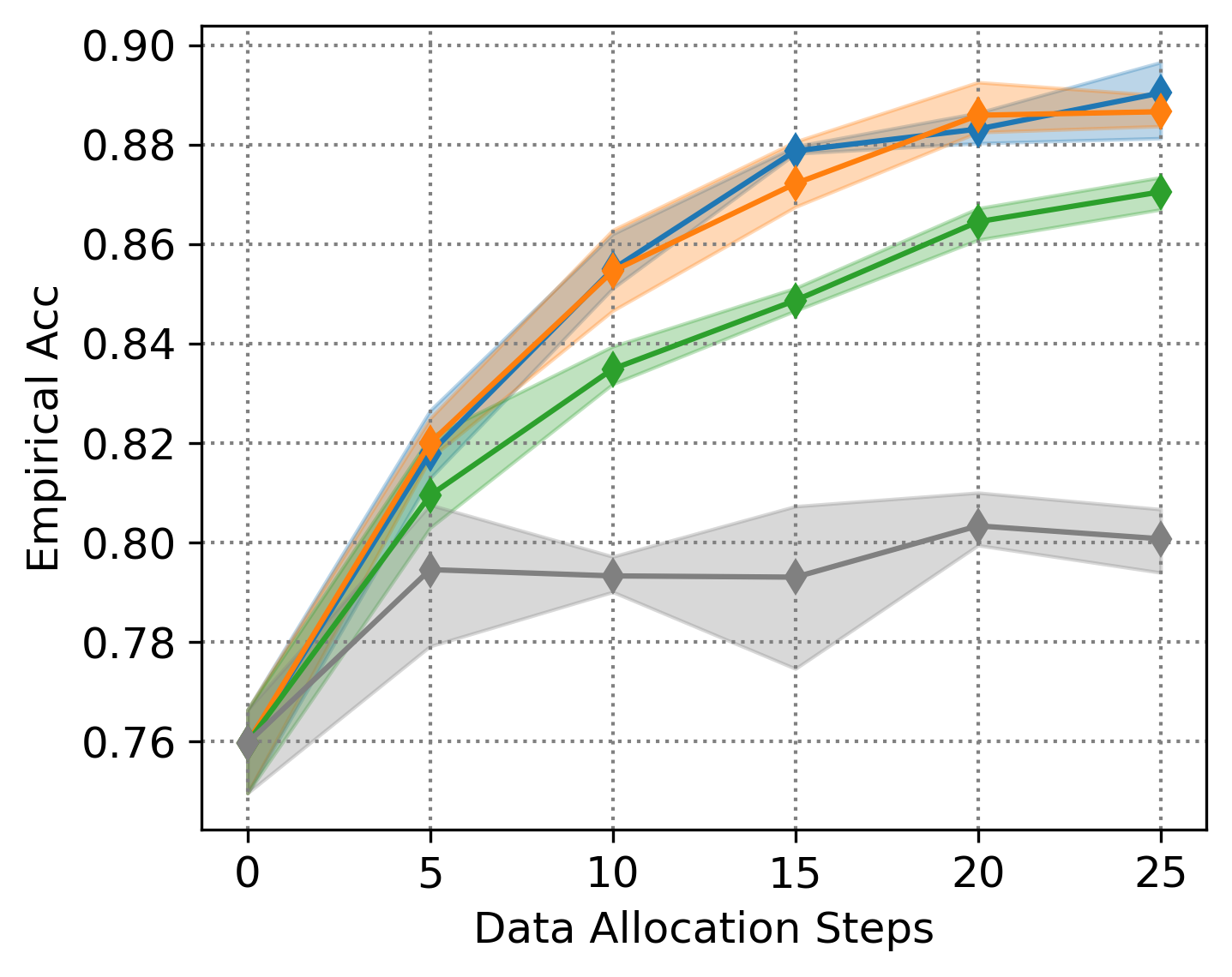}
    \end{subfigure}
    \hfill
    \begin{subfigure}{0.24\linewidth}
        \includegraphics[width=\linewidth]{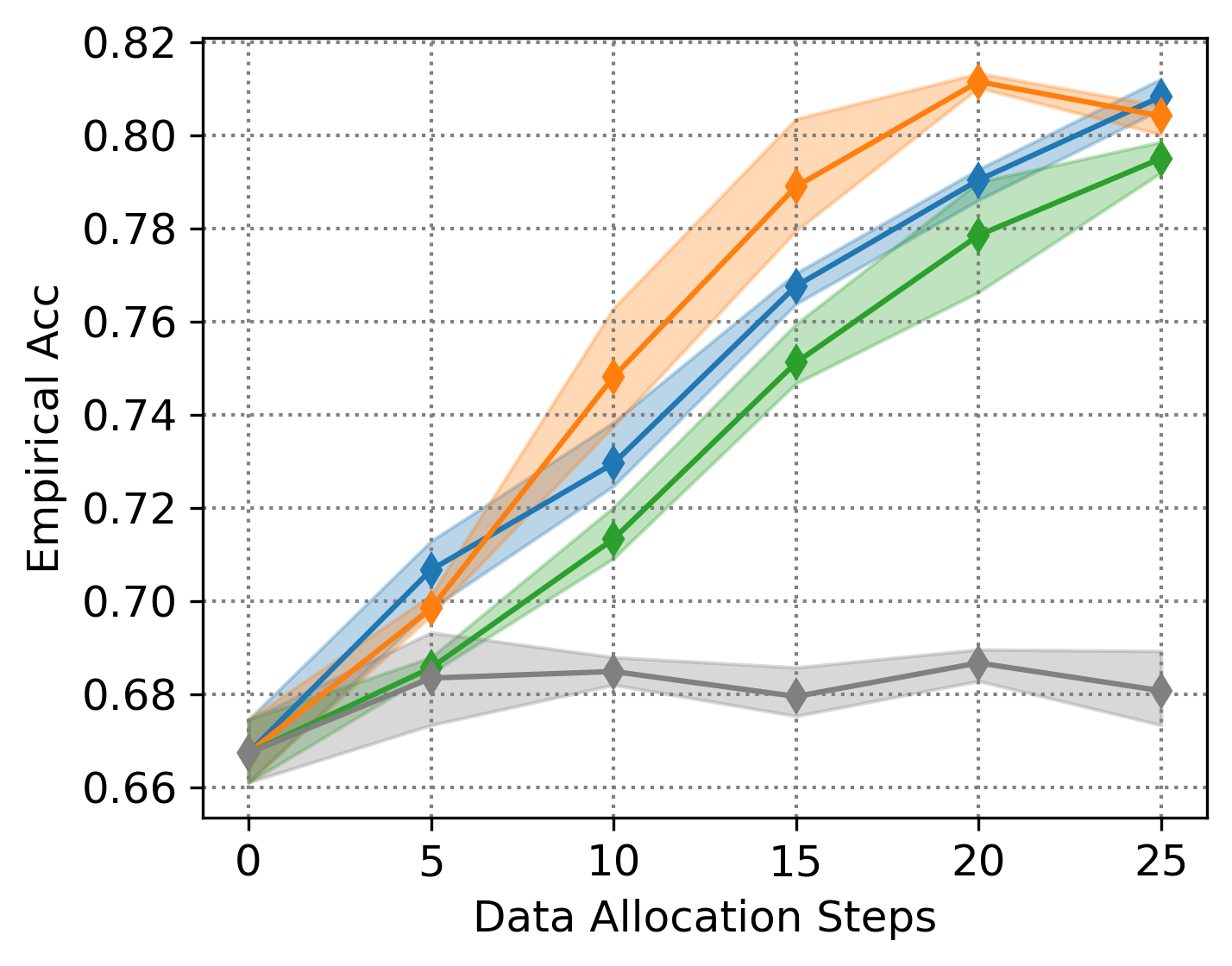}
    \end{subfigure}

    \vspace{1em}

    \begin{subfigure}{0.24\linewidth}
        \includegraphics[width=\linewidth]{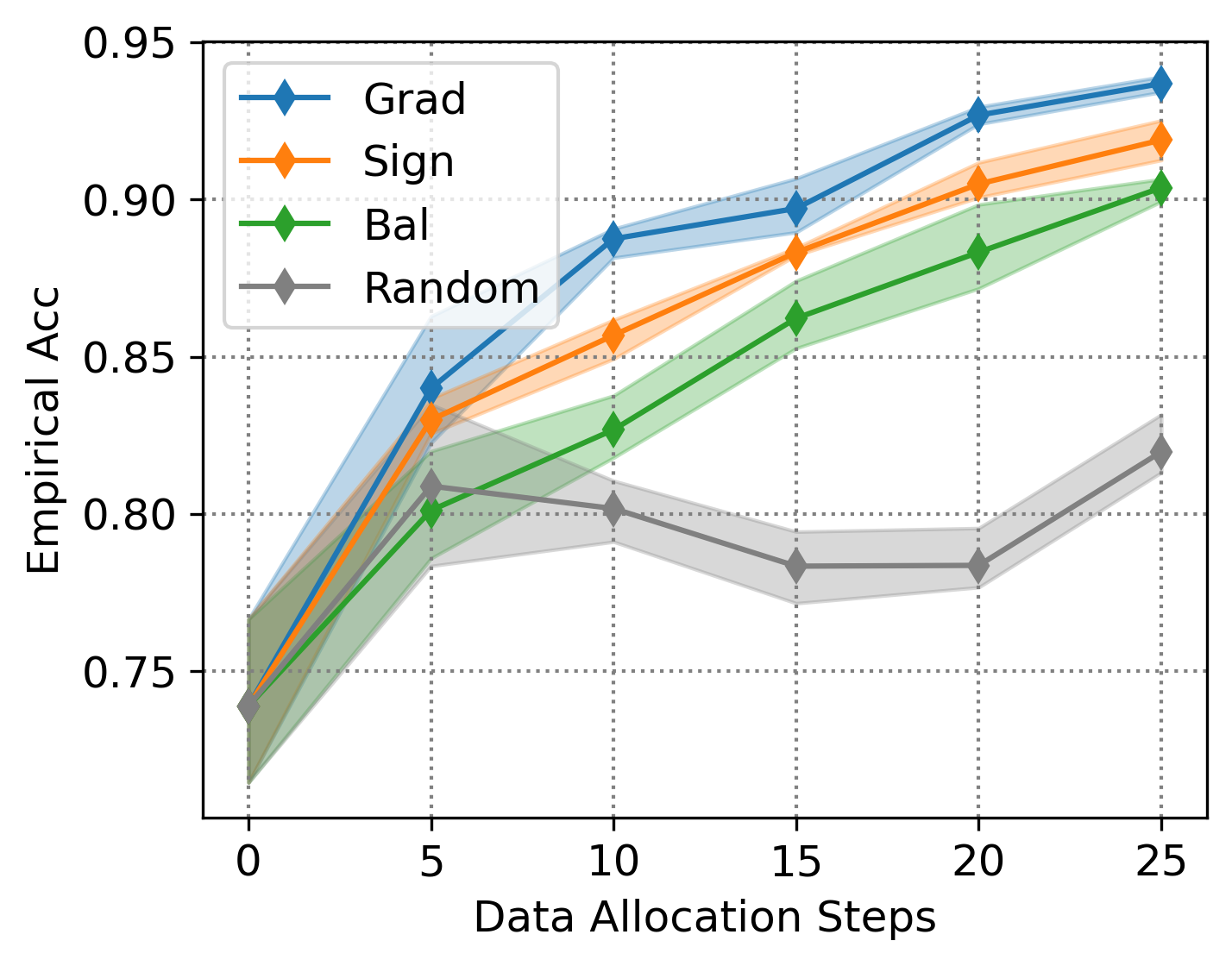}
    \end{subfigure}
    \hfill
    \begin{subfigure}{0.24\linewidth}
        \includegraphics[width=\linewidth]{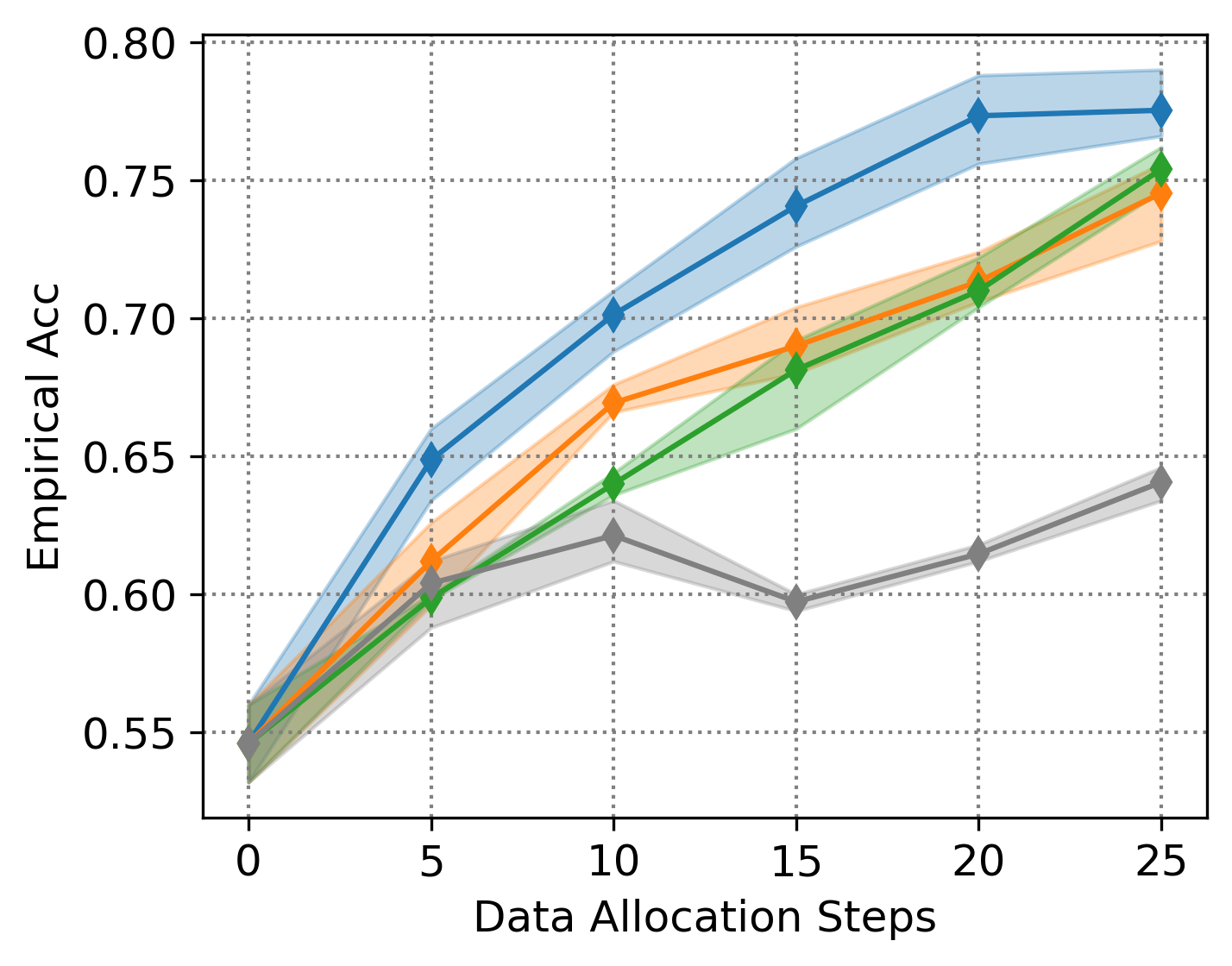}
    \end{subfigure}
    \hfill
    \begin{subfigure}{0.24\linewidth}
        \includegraphics[width=\linewidth]{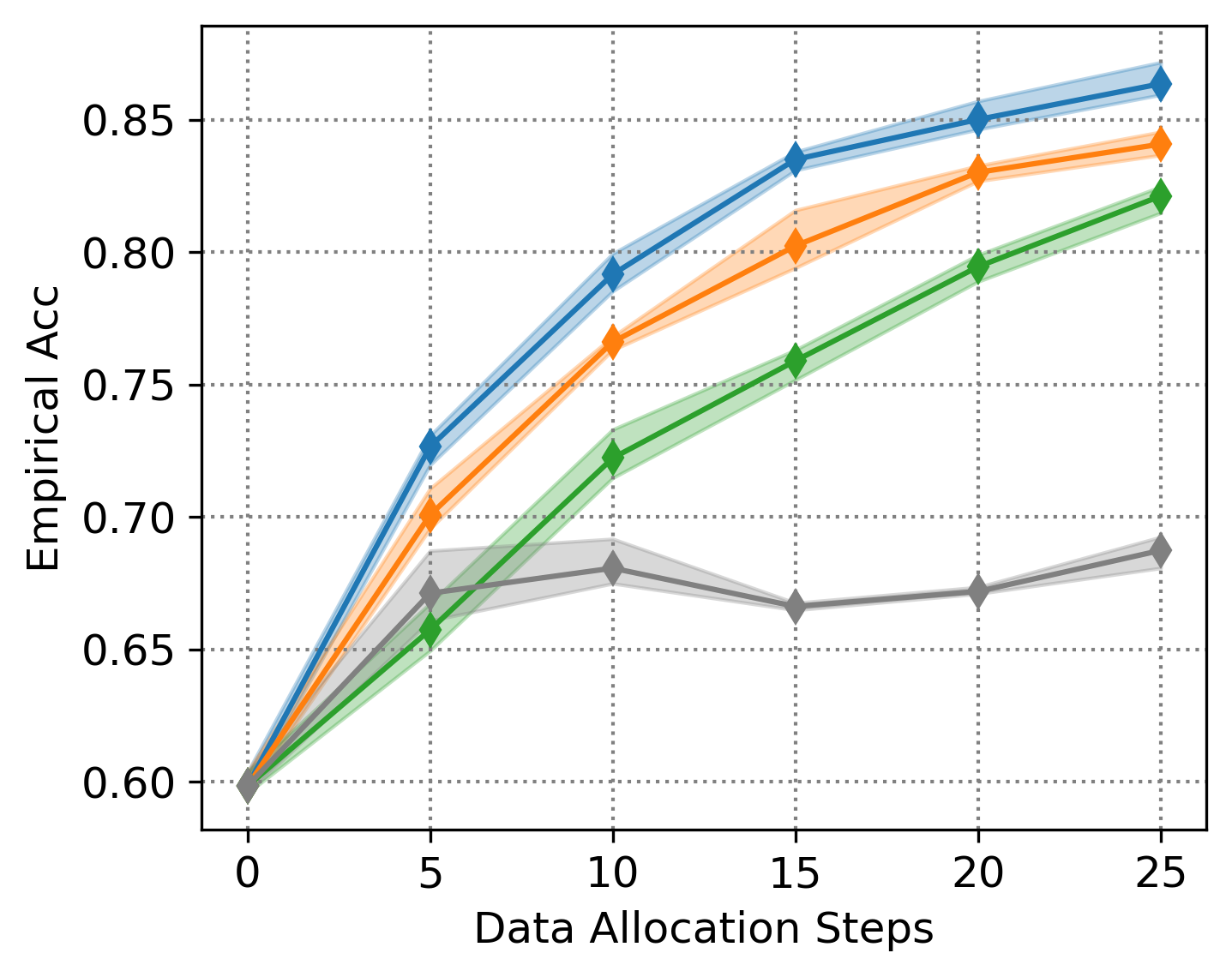}
    \end{subfigure}
    \hfill
    \begin{subfigure}{0.24\linewidth}
        \includegraphics[width=\linewidth]{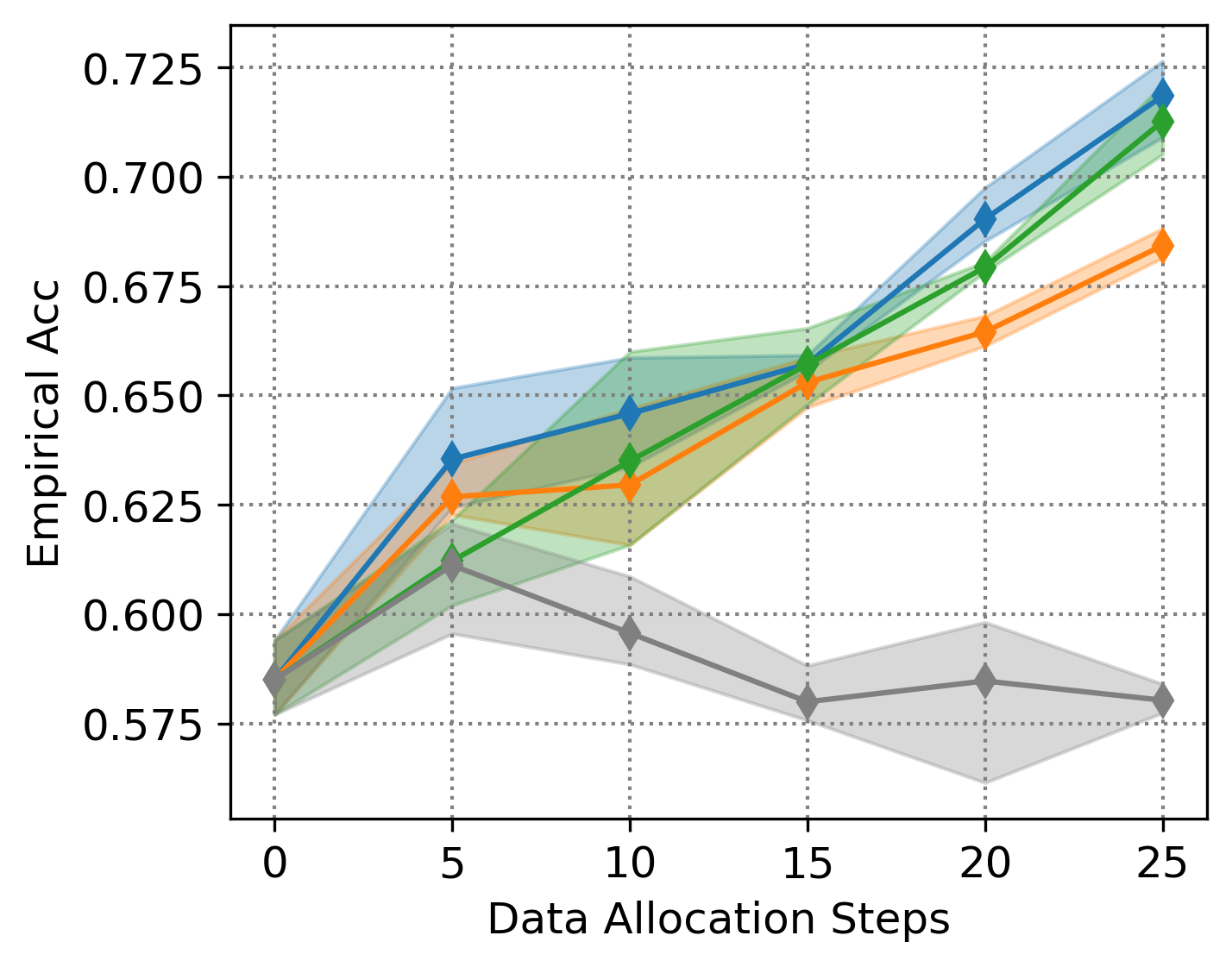}
    \end{subfigure}

    \vspace{1em}

    \begin{subfigure}{0.24\linewidth}
        \includegraphics[width=\linewidth]{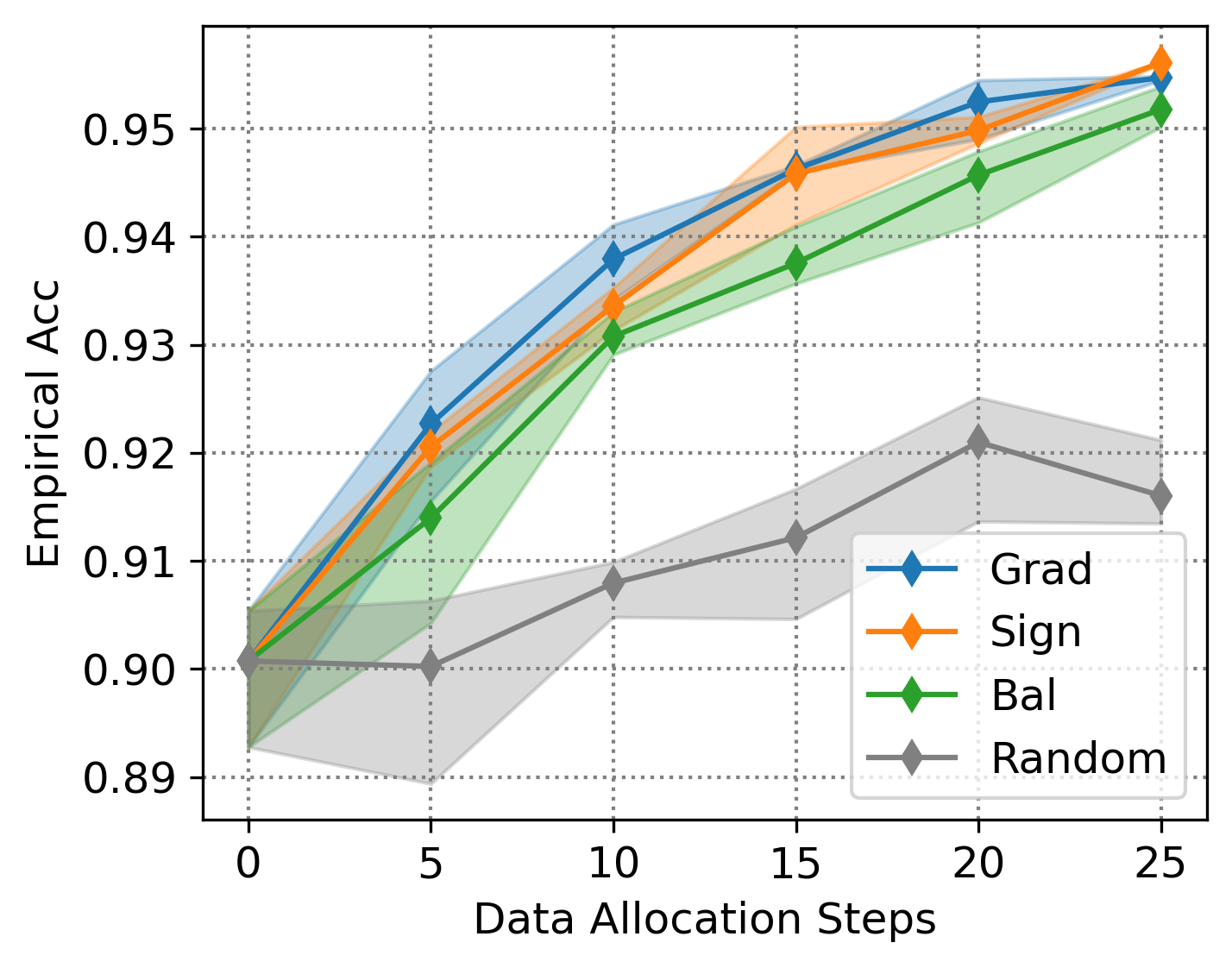}
        \caption{Waterbirds-syn}
    \end{subfigure}
    \hfill
    \begin{subfigure}{0.24\linewidth}
        \includegraphics[width=\linewidth]{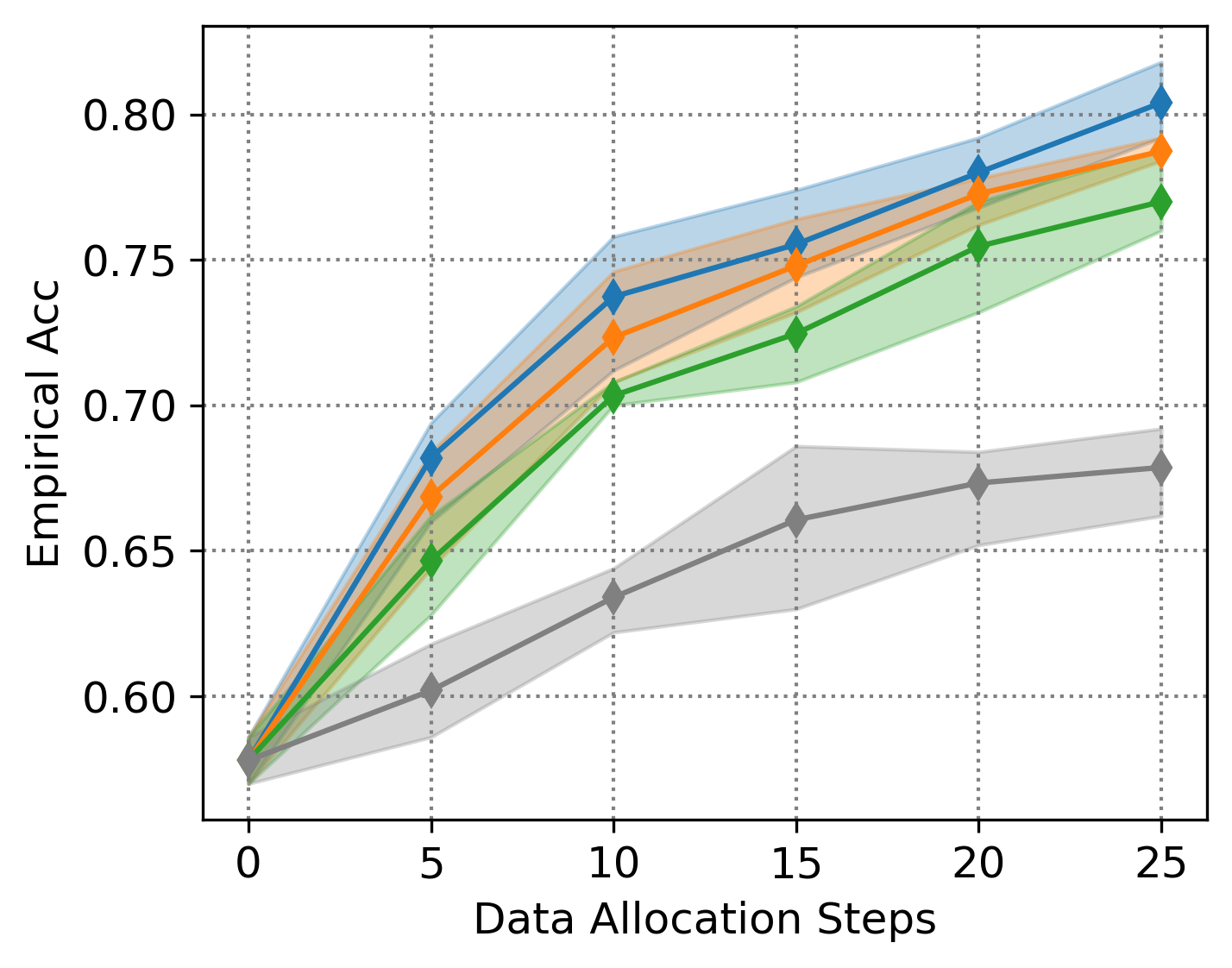}
        \caption{UrbanCars-syn}
    \end{subfigure}
    \hfill
    \begin{subfigure}{0.24\linewidth}
        \includegraphics[width=\linewidth]{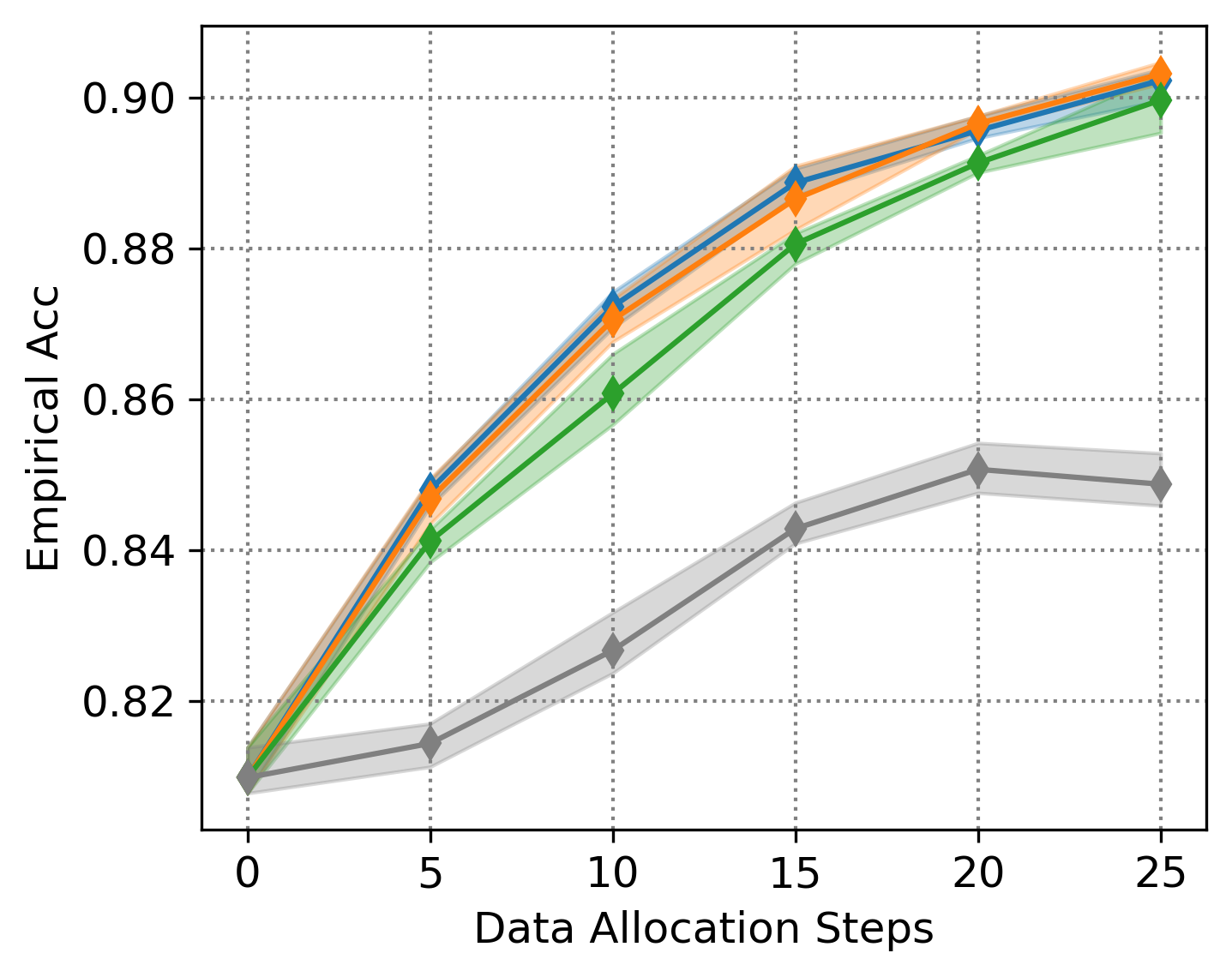}
        \caption{CelebA-subset}
    \end{subfigure}
    \hfill
    \begin{subfigure}{0.24\linewidth}
        \includegraphics[width=\linewidth]{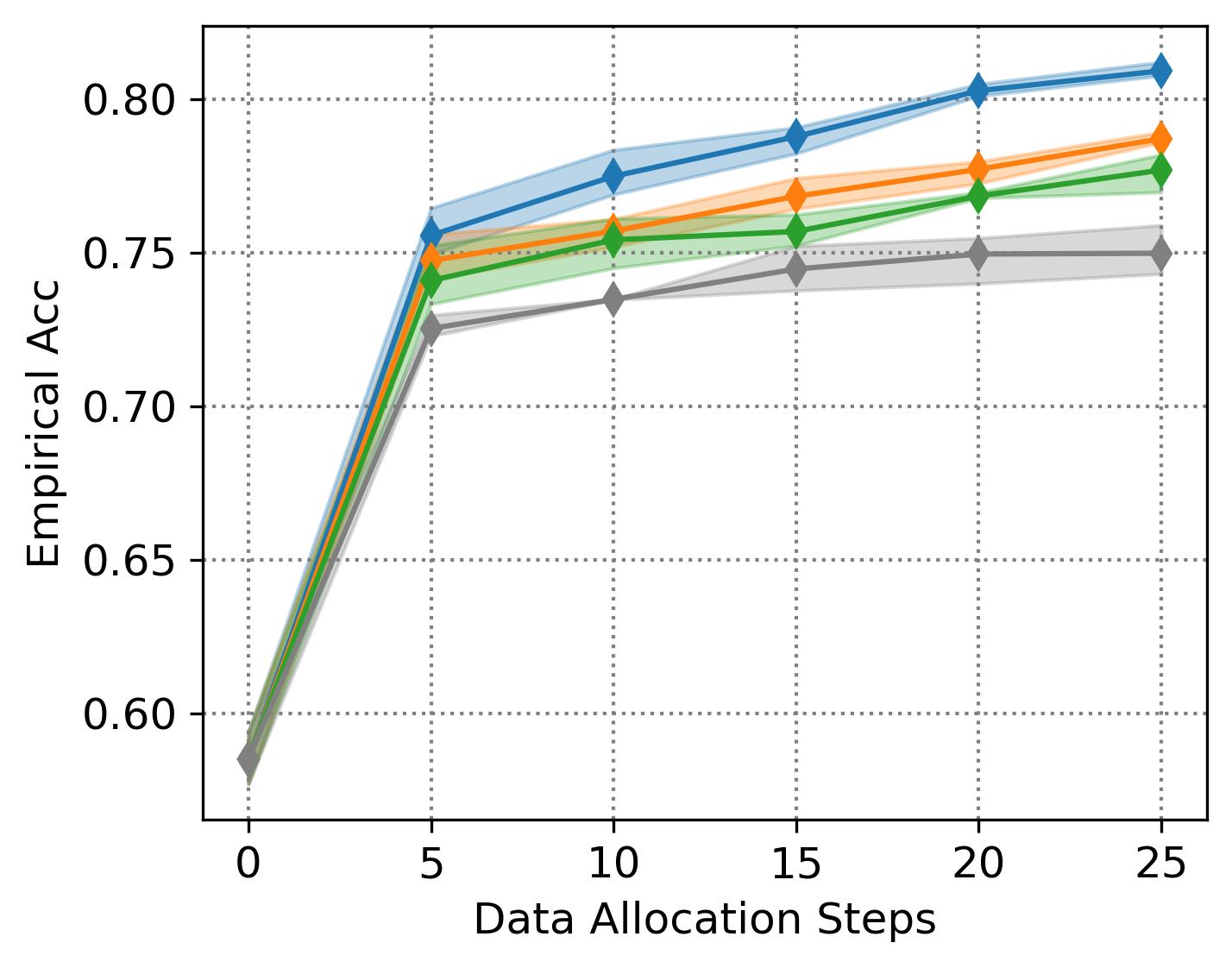}
        \caption{CivilComments-subset}
    \end{subfigure}

    \caption{{\bf Efficiency of data allocation along directions across datasets and initial configurations.} The efficiency of directions is validated by the comparison of the performance change as the data is allocated according to various directions.
    Each row corresponds to one allocation experiment with an initial configuration, and each column corresponds to a dataset. The legends are shared across plots.}
    \label{fig:allocation}
    \vspace{-3mm}
\end{figure*}

\begin{figure*}[h]
    \centering
    \begin{subfigure}{0.24\linewidth}
        \includegraphics[width=\linewidth]{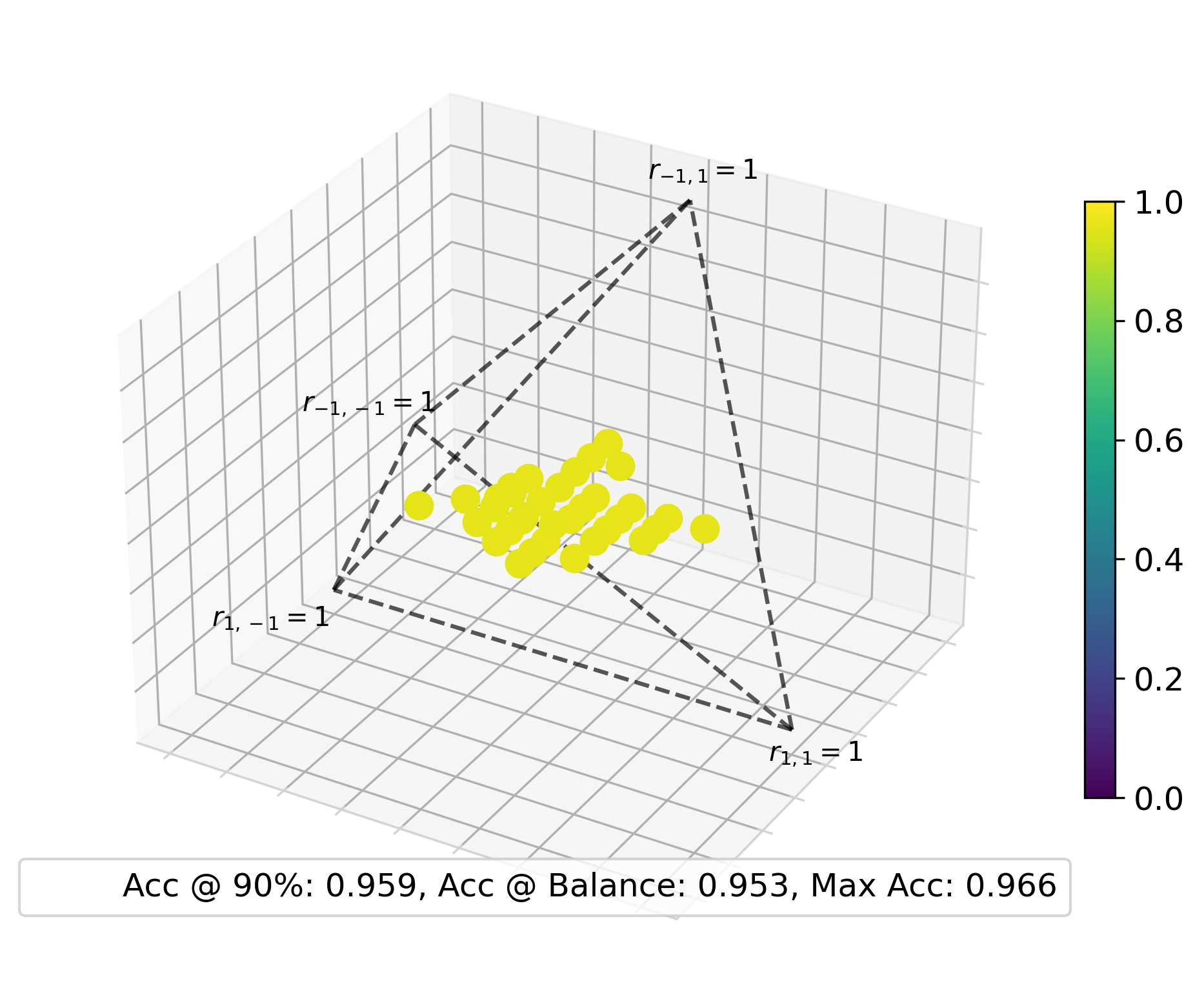}
        \caption{Waterbirds-syn}
    \end{subfigure}
    \hfill
    \begin{subfigure}{0.24\linewidth}
        \includegraphics[width=\linewidth]{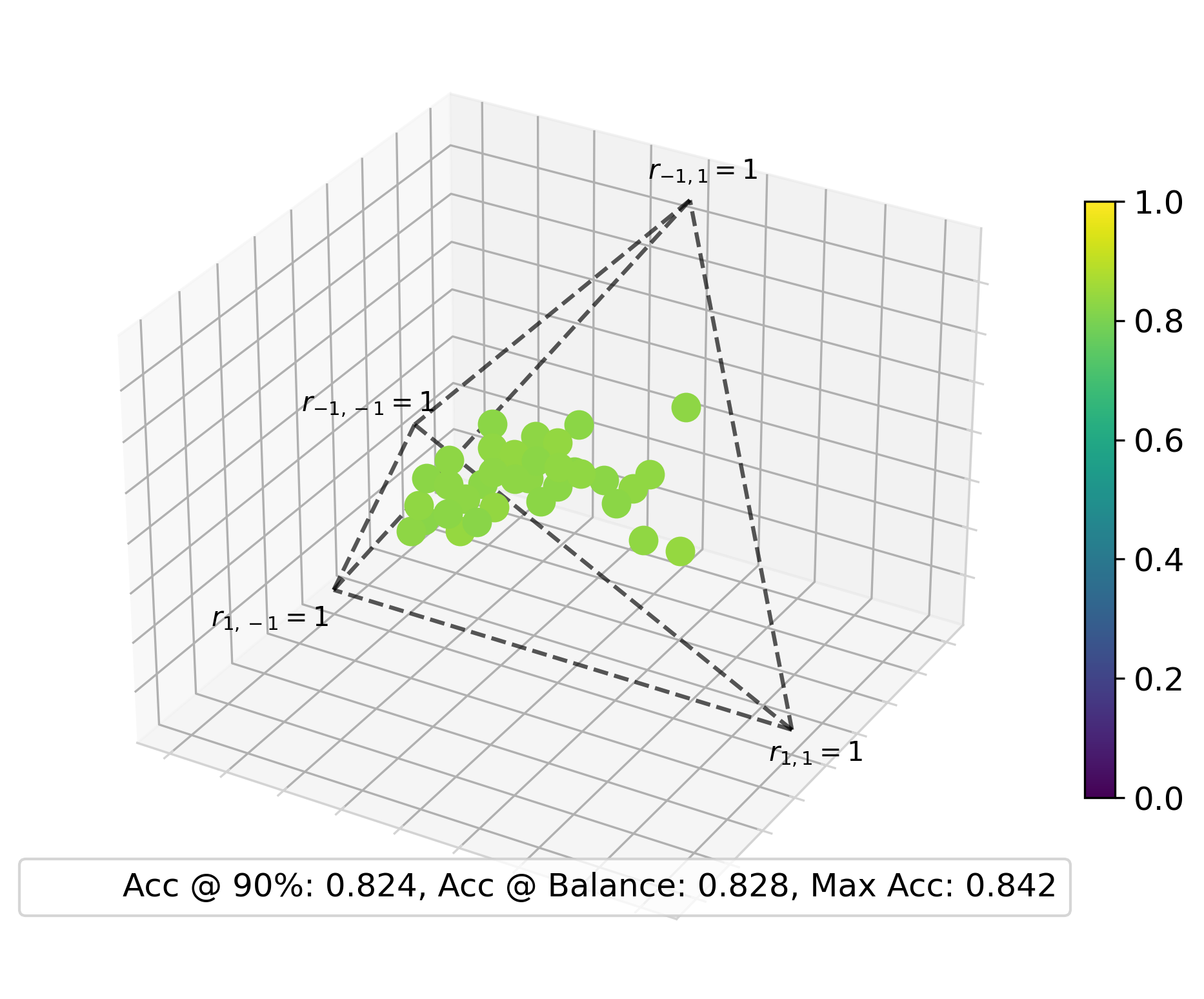}
        \caption{UrbanCars-syn}
    \end{subfigure}
    \hfill
    \begin{subfigure}{0.24\linewidth}
        \includegraphics[width=\linewidth]{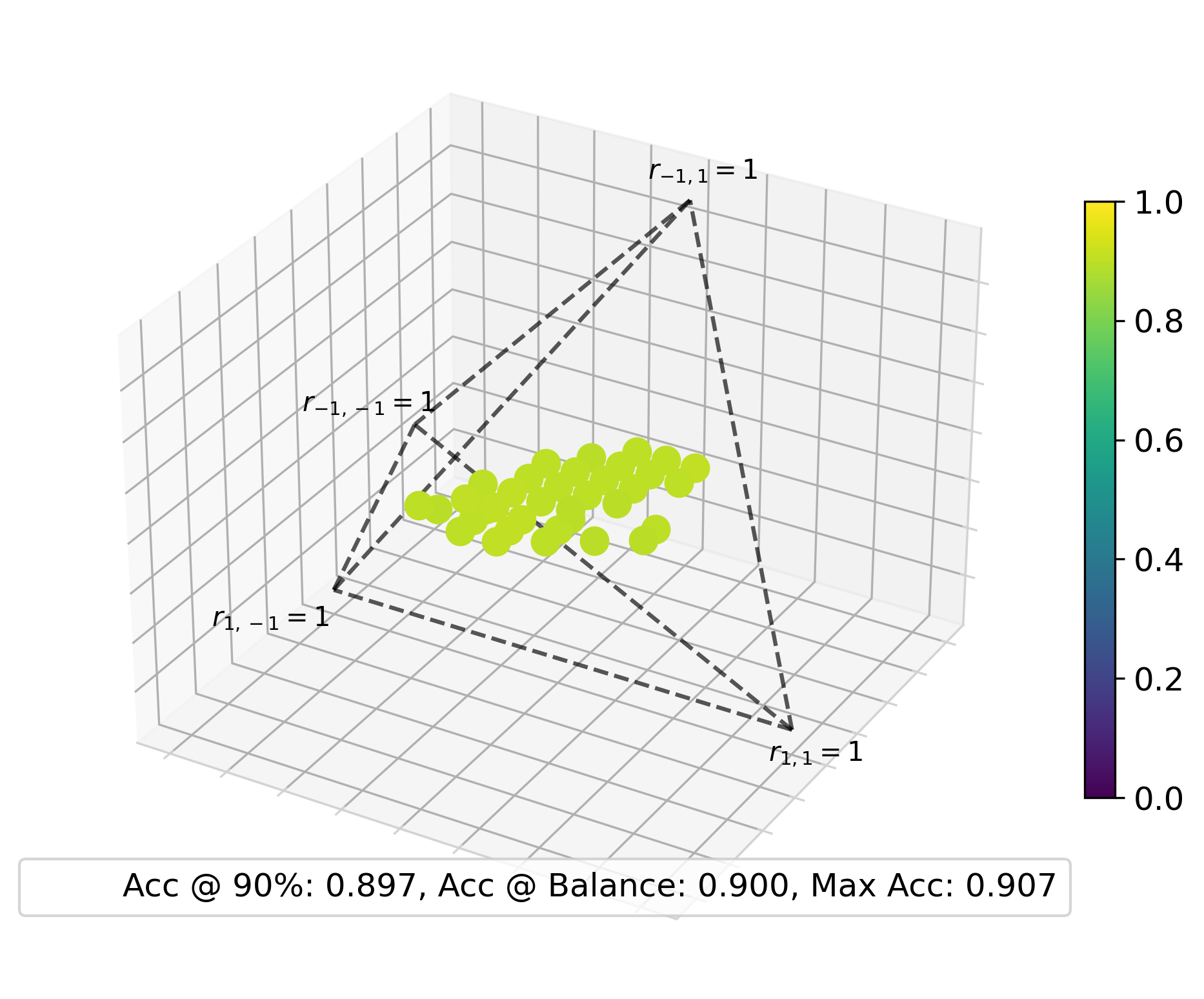}
        \caption{CelebA-subset}
    \end{subfigure}
    \hfill
    \begin{subfigure}{0.24\linewidth}
        \includegraphics[width=\linewidth]{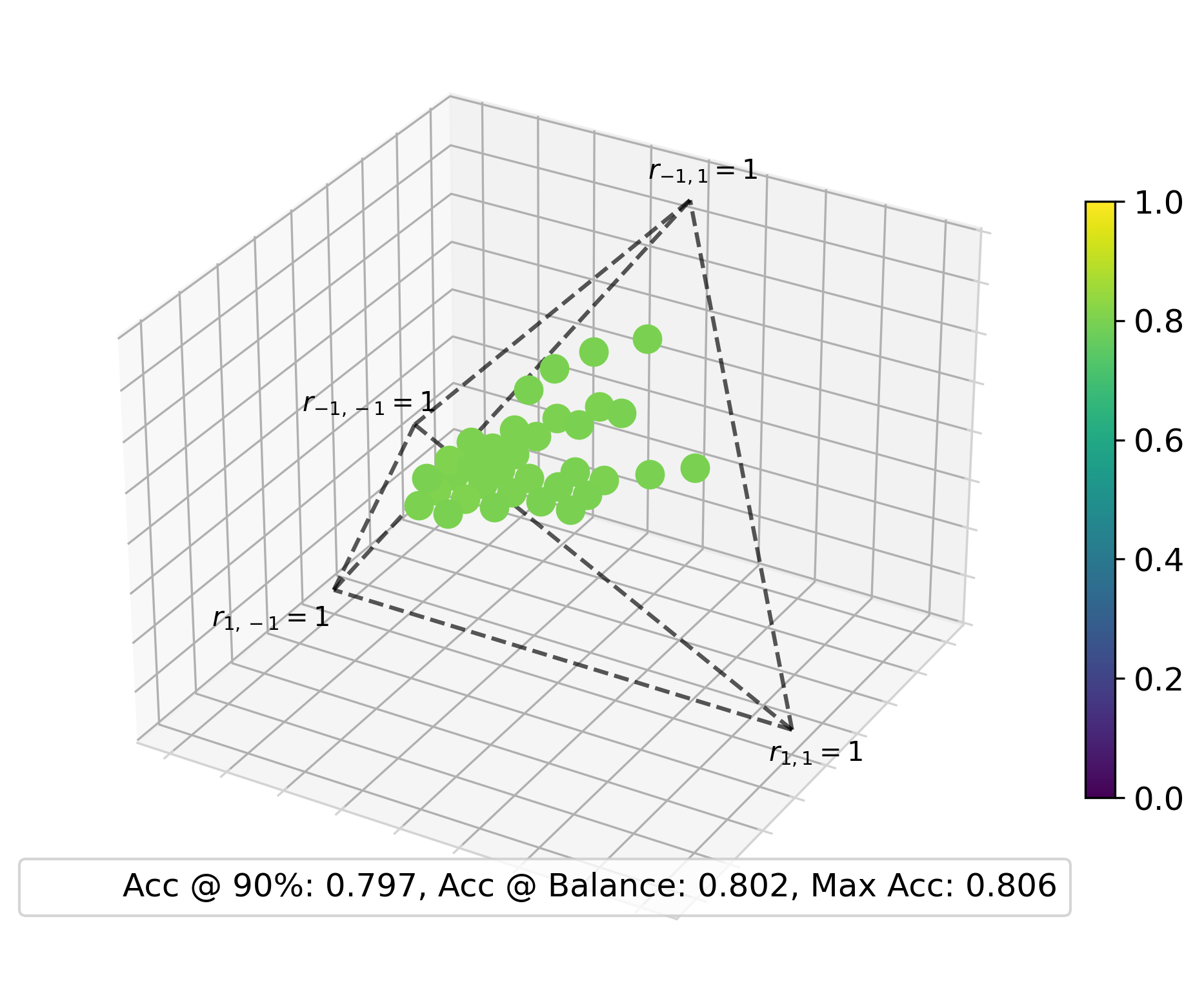}
        \caption{CivilComments-subset}
    \end{subfigure}
    \caption{{\bf Optimality of test performance across multiple configurations.} Among $7^3=343$ training data configurations, those achieving test performance above the 90th percentile are visualized within the 3-simplex. The results confirm that multiple configurations can achieve test performance comparable to or exceeding that of the balanced configuration, consistent with Figure~\ref{fig:max_acc_thr}.}
    \label{fig:max_acc_emp}
    \vspace{-2mm}
\end{figure*}

    \subsection{Existence of Direction Divergence}
    
    We first test whether Grad and Bal can lead to different empirical outcomes across training configurations. We sample a coarse grid of $7^3=343$ configurations by varying $(\alpha_+,\alpha_-,\beta)$ from $0.125$ to $0.875$ with stride $0.125$ for each dataset.
    For each configuration, we construct the corresponding training subset, update the configuration for 10 steps along either Grad or Bal, and evaluate the resulting empirical adjusted accuracy.
    For each initial configuration $\vp=(\alpha_+,\alpha_-,\beta)$, we measure the empirical accuracy gap of Grad over Bal as
        $A(\vp_{\mathrm{Grad}}) - A(\vp_{\mathrm{Bal}})$,
    where $\vp_{\mathrm{Grad}}$ and $\vp_{\mathrm{Bal}}$ denote the configurations obtained after 10 update steps along Grad and Bal, respectively. We visualize this gap in the 3-simplex in Figure~\ref{fig:emp-sim}, with lighter colors indicating larger gains from Grad.
    
    The empirical patterns are consistent with the theoretical prediction in Figure~\ref{fig:sim}: regions near the edges connecting $r_{-1,-1}$ to $r_{-1,1}$ and $r_{1,-1}$ to $r_{1,1}$ show larger Grad--Bal gaps. This confirms that there are many configurations where moving directly toward balance is not the most effective update, and where Grad provides a more efficient direction for improving robustness.

    \subsection{Efficiency When Directions Diverge}
    
    We next evaluate whether the divergence between Grad and Bal translates into more efficient data allocation. We select three examples from the theoretical simplex where Grad and Bal have both low directional similarity and a substantial predicted performance gap, which excludes cases where the directions align or perform similarly. Their initial configurations and direction comparisons are provided in Appendix~\ref{appd:simplex}.
    Starting from each initial configuration, we update the subgroup proportions along different directions for 5 to 25 steps. Each updated configuration defines a new target training distribution. To simulate data allocation under a fixed
    budget, we add or remove the same total number of samples according to the target subgroup proportions, train on the resulting dataset, and evaluate test performance. Thus, the change in accuracy, as the number of update steps increases, measures how efficiently each direction uses the data allocation budget.
    Figure~\ref{fig:allocation} reports adjusted accuracy across allocation steps. Each experiment is repeated with three random seeds, and we show the mean with min--max error bars. Across datasets and initial configurations, Grad yields the
    fastest performance improvement, demonstrating that the derived direction can guide data collection more efficiently than moving directly toward balance. We provide the statistical tests in Appendix~\ref{appd:stats-test}.

    \subsection{Optimality in Multiple Configurations}
    
    Finally, we test whether strong performance is limited to the balanced configuration. Using the same $7^3=343$ training configurations, we train and evaluate models for each configuration and visualize those whose adjusted accuracy is above the 90th percentile in Figure~\ref{fig:max_acc_emp}. For each dataset, we also mark the 90th-percentile accuracy, the balanced-configuration accuracy, and the maximum accuracy.
    The high-performing configurations are not concentrated only at the balanced center. Instead, many non-balanced configurations achieve accuracy comparable to, and sometimes higher than, the balanced configuration. This empirical pattern is consistent with the theoretical visualization in Figure~\ref{fig:max_acc_thr}, and supports our optimality claim: robust performance can be achieved by multiple subgroup configurations, not only by full balance. Additional visualizations with higher percentile thresholds are provided in the Appendix~\ref{appd:opt} to further support the claim.

\begin{wraptable}{r}{0.5\linewidth}
\vspace{-2mm}
\centering
\caption{
{\bf Grad-guided resampling on Example 2 of Waterbirds-syn}. Vanilla over-/under-sampling targets the balanced configuration, while Grad-guided resampling targets the alternative optimal configurations.
}
\label{tb:use-case}

\small
\resizebox{\linewidth}{!}{%
\begin{tabular}{lcc}
    \toprule
    Method & Adjusted Acc. & WGA \\
    \midrule
    Original & 0.766 & 0.376 \\
    Over-sampling & 0.768 & 0.366 \\
    Over-sampling + Grad & 0.821 & 0.541 \\
    Under-sampling & 0.777 & 0.837 \\
    Under-sampling + Grad & 0.901 & 0.848 \\
    \bottomrule
\end{tabular}%
}

\vspace{-2mm}
\end{wraptable}

    \subsection{Application to Grad-Guided Resampling}
    Our main application is data allocation under a fixed annotation budget, where Grad indicates how to adjust the subgroups for more effective improvement in subpopulation robustness. We further show that it can be integrated with standard resampling methods. In vanilla over-sampling or under-sampling, the target distribution is typically the
    fully balanced configuration. In contrast, Grad-guided resampling targets one of the identified optimal configurations obtained by moving along the derived Grad direction.
    
    We evaluate this application on the Example 2 configuration of Waterbirds-syn in Figure~\ref{fig:allocation}. We set the total training size to 9,000 so that each subgroup has enough samples for resampling, and compare vanilla resampling with Grad-guided resampling. As shown in Table~\ref{tb:use-case}, replacing the balanced target with the Grad-induced target improves both adjusted accuracy and worst-group accuracy, especially for over-sampling. 
    We further provide the subgroup statistics in Appendix~\ref{appd:app} to show that the Grad-guided over/under-sampling tends to utilize more original data and maintain better performance across multiple metrics.
    This suggests that Grad can serve as a drop-in guide for existing data rebalancing techniques.

    \subsection{Application to Last-layer Retraining Methods}\label{sec:llr}
    Last-layer retraining methods provide a complementary application as their second stage often uses a balanced held-out subset~\cite{kirichenko2023last,labonte2023towards}. We replace only this stage-two subgroup composition with the Grad-derived allocation, keep its total size fixed, and leave the remaining components of each method unchanged. Table~\ref{tab:llr} reports results for DFR~\cite{kirichenko2023last}, SELF~\cite{labonte2023towards}, GSR~\cite{qiao2025group}, and AFR~\cite{qiu2023simple}. Across Waterbirds and CelebA, the imbalanced Grad allocation retains the WGA of DFR and SELF and improves or retains GSR and AFR. The purpose is not to replace these algorithms, but to show that exact balance can be relaxed when minority examples are costly while preserving their robustness benefits.

\begin{table}[t]
    \begin{center}
    \caption{{\bf Comparison of worst-group and adjusted accuracy (\%) under different stage-two data allocation strategies.} Results are reported for last-layer retraining using the original stage-two allocation and the Grad allocation.}
    \vspace{2mm}
    \label{tab:llr}
    \small
    \setlength\tabcolsep{3pt}
    \begin{tabular}{llcccc}
        \toprule
        & & \multicolumn{2}{c}{Waterbirds} & \multicolumn{2}{c}{CelebA} \\
        \cmidrule(lr){3-4}\cmidrule(lr){5-6}
        Method & Metric & Original & Grad & Original & Grad \\
        \midrule
        \multirow{2}{*}{DFR}
            & WGA & 89.30{\scriptsize $\pm$0.55} & 89.10{\scriptsize $\pm$0.75} & 87.49{\scriptsize $\pm$1.27} & 87.36{\scriptsize $\pm$0.13} \\
            & AA  & 91.37{\scriptsize $\pm$0.45} & 91.60{\scriptsize $\pm$0.19} & 90.87{\scriptsize $\pm$0.82} & 90.39{\scriptsize $\pm$0.33} \\
        \midrule
        \multirow{2}{*}{SELF}
            & WGA & 88.28{\scriptsize $\pm$1.37} & 88.53{\scriptsize $\pm$0.76} & 82.59{\scriptsize $\pm$2.31} & 82.18{\scriptsize $\pm$3.81} \\
            & AA  & 91.50{\scriptsize $\pm$0.26} & 91.74{\scriptsize $\pm$0.26} & 90.52{\scriptsize $\pm$0.57} & 90.74{\scriptsize $\pm$0.72} \\
        \midrule
        \multirow{2}{*}{GSR}
            & WGA & 91.90{\scriptsize $\pm$1.42} & 91.96{\scriptsize $\pm$0.95} & 85.74{\scriptsize $\pm$2.57} & 86.67{\scriptsize $\pm$0.96} \\
            & AA  & 94.00{\scriptsize $\pm$0.39} & 93.79{\scriptsize $\pm$0.37} & 89.86{\scriptsize $\pm$0.25} & 90.23{\scriptsize $\pm$0.31} \\
        \midrule
        \multirow{2}{*}{AFR}
            & WGA & 86.90{\scriptsize $\pm$3.47} & 88.27{\scriptsize $\pm$0.24} & 80.74{\scriptsize $\pm$2.10} & 83.84{\scriptsize $\pm$0.22} \\
            & AA  & 91.95{\scriptsize $\pm$1.17} & 92.50{\scriptsize $\pm$0.76} & 89.05{\scriptsize $\pm$0.75} & 89.89{\scriptsize $\pm$0.26} \\
        \bottomrule
    \end{tabular}
    \end{center}
    \vspace{-2mm}
\end{table}

\section{Discussion}

    This work revisits whether fully balanced subgroup proportions are necessary for subpopulation robustness. Our analysis shows that robust performance can be achieved by a broader set of training configurations, even including imbalanced ones. We further derive gradient-based directions for adjusting subgroup proportions, and show that these directions can diverge from the path toward full balance. Experiments on vision and language benchmarks confirm that Grad-guided data allocation can improve performance more efficiently than directly enforcing balance. Overall, our work provides a principled method to strategically shape subgroup distributions for robust generalization.

{\bf Broad Impacts.}
    This work presents a data-centric perspective on achieving robust generalization, and our analysis guides the construction of datasets for better subpopulation robustness in downstream model training and deployment, particularly helpful in high-stakes domains where balance is infeasible to achieve. Although we focus on data-side effects without direct comparison with model-based methods to keep the theoretical insights clear, integrating our data-optimization principles with existing model-side approaches is a promising direction for real-world deployment. In Appendix~\ref{appd:gdro}, we provide an example showing that integrating our data allocation with Group DRO~\cite{sagawa2019distributionally} and Regularization~\cite{wang2024effect} can further improve performance.

{\bf Limitations and Future Work.}
    One limitation lies in the feature separability estimation. Although the estimation generalizes well across four datasets, it can be inaccurate in practice, especially in noisy data or unstable optimization. We partially address this limitation by introducing the Sign direction, which is more robust to separability-estimation error. We included an empirical study on the robustness to estimation error in Appendix~\ref{appd:est-error}.
    Another limitation is scalability: while the theoretical framework covers multiple attributes, the complexity of calculating intermediate derivatives increases quickly as the number of subgroups grows. Furthermore, current multi-attribute benchmarks are limited in scope. We thus focus our analysis on binary attribute settings in the main paper, provide a demonstration of 3-attribute in Appendix~\ref{appd:n3}, and leave further analysis to future work.
    Finally, our global analysis optimizes AA rather than WGA. The experiments show that the resulting direction improves WGA in the studied settings, and a local WGA direction can be estimated. However, a general theoretical connection between the two objectives remains open.

{\bf Acknowledgment.} This work was partially supported by NSF IIS \#2146091, IIS \#2345235, and USDA \#2023-67021-41368.

{
\small

    \bibliographystyle{plain}
    \bibliography{neurips_2026}

@inproceedings{qiao2025group,
  title={Group-robust sample reweighting for subpopulation shifts via influence functions},
  author={Qiao, Rui and Wu, Zhaoxuan and Wang, Jingtan and Koh, Pang Wei and Low, Bryan Kian Hsiang},
  booktitle={International Conference on Learning Representations},
  volume={2025},
  pages={36959--36980},
  year={2025}
}

@article{labonte2023towards,
  title={Towards last-layer retraining for group robustness with fewer annotations},
  author={LaBonte, Tyler and Muthukumar, Vidya and Kumar, Abhishek},
  journal={Advances in Neural Information Processing Systems},
  volume={36},
  pages={11552--11579},
  year={2023}
}

@inproceedings{
kirichenko2023last,
title={Last Layer Re-Training is Sufficient for Robustness to Spurious Correlations},
author={Polina Kirichenko and Pavel Izmailov and Andrew Gordon Wilson},
booktitle={The Eleventh International Conference on Learning Representations },
year={2023},
url={https://openreview.net/forum?id=Zb6c8A-Fghk}
}

@article{izmailov2022feature,
  title={On feature learning in the presence of spurious correlations},
  author={Izmailov, Pavel and Kirichenko, Polina and Gruver, Nate and Wilson, Andrew G},
  journal={Advances in Neural Information Processing Systems},
  volume={35},
  pages={38516--38532},
  year={2022}
}

@inproceedings{qiu2023simple,
  title={Simple and fast group robustness by automatic feature reweighting},
  author={Qiu, Shikai and Potapczynski, Andres and Izmailov, Pavel and Wilson, Andrew Gordon},
  booktitle={International Conference on Machine Learning},
  pages={28448--28467},
  year={2023},
  organization={PMLR}
}

@inproceedings{li2023whac,
  title={A whac-a-mole dilemma: Shortcuts come in multiples where mitigating one amplifies others},
  author={Li, Zhiheng and Evtimov, Ivan and Gordo, Albert and Hazirbas, Caner and Hassner, Tal and Ferrer, Cristian Canton and Xu, Chenliang and Ibrahim, Mark},
  booktitle={Proceedings of the IEEE/CVF Conference on Computer Vision and Pattern Recognition},
  pages={20071--20082},
  year={2023}
}

@inproceedings{krause20133d,
  title={3d object representations for fine-grained categorization},
  author={Krause, Jonathan and Stark, Michael and Deng, Jia and Fei-Fei, Li},
  booktitle={Proceedings of the IEEE international conference on computer vision workshops},
  pages={554--561},
  year={2013}
}

@inproceedings{sagawa2019distributionally,
title={Distributionally robust neural networks for group shifts: On the importance of regularization for worst-case generalization},
author={Sagawa, Shiori and Koh, Pang Wei and Hashimoto, Tatsunori B and Liang, Percy},
booktitle={International Conference on Learning Representations},
year={2020}
}

@inproceedings{lin2017focal,
  title={Focal loss for dense object detection},
  author={Lin, Tsung-Yi and Goyal, Priya and Girshick, Ross and He, Kaiming and Doll{\'a}r, Piotr},
  booktitle={Proceedings of the IEEE international conference on computer vision},
  pages={2980--2988},
  year={2017}
}

@inproceedings{cui2019class,
  title={Class-balanced loss based on effective number of samples},
  author={Cui, Yin and Jia, Menglin and Lin, Tsung-Yi and Song, Yang and Belongie, Serge},
  booktitle={Proceedings of the IEEE/CVF conference on computer vision and pattern recognition},
  pages={9268--9277},
  year={2019}
}

@inproceedings{zhang2022correct,
  title={Correct-N-Contrast: a Contrastive Approach for Improving Robustness to Spurious Correlations},
  author={Zhang, Michael and Sohoni, Nimit S and Zhang, Hongyang R and Finn, Chelsea and Re, Christopher},
  booktitle={International Conference on Machine Learning},
  pages={26484--26516},
  year={2022},
  organization={PMLR}
}

@inproceedings{sagawa2020investigation,
  title={An investigation of why overparameterization exacerbates spurious correlations},
  author={Sagawa, Shiori and Raghunathan, Aditi and Koh, Pang Wei and Liang, Percy},
  booktitle={International Conference on Machine Learning},
  pages={8346--8356},
  year={2020},
  organization={PMLR}
}

@inproceedings{wang2024effect,
  title={On the Effect of Key Factors in Spurious Correlation: A theoretical Perspective},
  author={Wang, Yipei and Wang, Xiaoqian},
  booktitle={International Conference on Artificial Intelligence and Statistics},
  pages={3745--3753},
  year={2024},
  organization={PMLR}
}

@inproceedings{lu2024neural,
  title={Neural Collapse Inspired Debiased Representation Learning for Min-max Fairness},
  author={Lu, Shenyu and Chai, Junyi and Wang, Xiaoqian},
  booktitle={Proceedings of the 30th ACM SIGKDD Conference on Knowledge Discovery and Data Mining},
  pages={2048--2059},
  year={2024}
}

@article{arjovsky2019invariant,
  title={Invariant risk minimization},
  author={Arjovsky, Martin and Bottou, L{\'e}on and Gulrajani, Ishaan and Lopez-Paz, David},
  journal={arXiv preprint arXiv:1907.02893},
  year={2019}
}

@inproceedings{yang2023change,
  title={Change is Hard: A Closer Look at Subpopulation Shift},
  author={Yang, Yuzhe and Zhang, Haoran and Katabi, Dina and Ghassemi, Marzyeh},
  booktitle={International Conference on Machine Learning},
  pages={39584--39622},
  year={2023},
  organization={PMLR}
}

@article{hardt2016equality,
  title={Equality of opportunity in supervised learning},
  author={Hardt, Moritz and Price, Eric and Srebro, Nati},
  journal={Advances in neural information processing systems},
  volume={29},
  year={2016}
}

@article{krawczyk2016learning,
  title={Learning from imbalanced data: open challenges and future directions},
  author={Krawczyk, Bartosz},
  journal={Progress in artificial intelligence},
  volume={5},
  number={4},
  pages={221--232},
  year={2016},
  publisher={Springer}
}

@article{mehrabi2021survey,
  title={A survey on bias and fairness in machine learning},
  author={Mehrabi, Ninareh and Morstatter, Fred and Saxena, Nripsuta and Lerman, Kristina and Galstyan, Aram},
  journal={ACM computing surveys (CSUR)},
  volume={54},
  number={6},
  pages={1--35},
  year={2021},
  publisher={ACM New York, NY, USA}
}

@inproceedings{liu2022self,
  title={Self-supervised Learning is More Robust to Dataset Imbalance},
  author={Liu, Hong and HaoChen, Jeff Z and Gaidon, Adrien and Ma, Tengyu},
  booktitle={International Conference on Learning Representations},
  year={2022}
}

@inproceedings{nagarajan2021understanding,
  title={Understanding the failure modes of out-of-distribution generalization},
  author={Nagarajan, Vaishnavh and Andreassen, Anders and Neyshabur, Behnam},
  booktitle={International Conference on Learning Representations},
  year={2021}
}

@inproceedings{borkan2019nuanced,
  title={Nuanced metrics for measuring unintended bias with real data for text classification},
  author={Borkan, Daniel and Dixon, Lucas and Sorensen, Jeffrey and Thain, Nithum and Vasserman, Lucy},
  booktitle={Companion proceedings of the 2019 world wide web conference},
  pages={491--500},
  year={2019}
}

@article{he2020deberta,
  title={Deberta: Decoding-enhanced bert with disentangled attention},
  author={He, Pengcheng and Liu, Xiaodong and Gao, Jianfeng and Chen, Weizhu},
  journal={arXiv preprint arXiv:2006.03654},
  year={2020}
}

@inproceedings{li2025let,
  title={Let samples speak: Mitigating spurious correlation by exploiting the clusterness of samples},
  author={Li, Weiwei and Liu, Junzhuo and Ren, Yuanyuan and Zheng, Yuchen and Liu, Yahao and Li, Wen},
  booktitle={Proceedings of the IEEE/CVF Conference on Computer Vision and Pattern Recognition},
  pages={15486--15496},
  year={2025}
}

@inproceedings{zheng2025neurontune,
  title={NeuronTune: Towards Self-Guided Spurious Bias Mitigation},
  author={Zheng, Guangtao and Ye, Wenqian and Zhang, Aidong},
  booktitle={International Conference on Machine Learning},
  pages={78454--78477},
  year={2025},
  organization={PMLR}
}

@inproceedings{devlin2019bert,
  title={Bert: Pre-training of deep bidirectional transformers for language understanding},
  author={Devlin, Jacob and Chang, Ming-Wei and Lee, Kenton and Toutanova, Kristina},
  booktitle={Proceedings of the 2019 conference of the North American chapter of the association for computational linguistics: human language technologies, volume 1 (long and short papers)},
  pages={4171--4186},
  year={2019}
}

@article{deng2023robust,
  title={Robust learning with progressive data expansion against spurious correlation},
  author={Deng, Yihe and Yang, Yu and Mirzasoleiman, Baharan and Gu, Quanquan},
  journal={Advances in neural information processing systems},
  volume={36},
  pages={1390--1402},
  year={2023}
}

@inproceedings{idrissi2022simple,
  title={Simple data balancing achieves competitive worst-group-accuracy},
  author={Idrissi, Badr Youbi and Arjovsky, Martin and Pezeshki, Mohammad and Lopez-Paz, David},
  booktitle={Conference on Causal Learning and Reasoning},
  pages={336--351},
  year={2022},
  organization={PMLR}
}

@article{mei2021generalization,
  title={The Generalization Error of Random Features Regression: Precise Asymptotics and the Double Descent Curve},
  author={Mei, Song and Montanari, Andrea},
  journal={Communications on Pure and Applied Mathematics},
  year={2021}
}

@article{montanari2019generalization,
  title={The generalization error of max-margin linear classifiers: Benign overfitting and high dimensional asymptotics in the overparametrized regime},
  author={Montanari, Andrea and Ruan, Feng and Sohn, Youngtak and Yan, Jun},
  journal={arXiv preprint arXiv:1911.01544},
  year={2019}
}

@inproceedings{ming2022impact,
  title={On the impact of spurious correlation for out-of-distribution detection},
  author={Ming, Yifei and Yin, Hang and Li, Yixuan},
  booktitle={Proceedings of the AAAI Conference on Artificial Intelligence},
  volume={36},
  number={9},
  pages={10051--10059},
  year={2022}
}

@inproceedings{yao2022improving,
  title={Improving out-of-distribution robustness via selective augmentation},
  author={Yao, Huaxiu and Wang, Yu and Li, Sai and Zhang, Linjun and Liang, Weixin and Zou, James and Finn, Chelsea},
  booktitle={International Conference on Machine Learning},
  pages={25407--25437},
  year={2022},
  organization={PMLR}
}

@article{wen2025elastic,
  title={Elastic Representation: Mitigating Spurious Correlations for Group Robustness},
  author={Wen, Tao and Wang, Zihan and Zhang, Quan and Lei, Qi},
  journal={arXiv preprint arXiv:2502.09850},
  year={2025}
}

@article{duchi2021learning,
  title={Learning models with uniform performance via distributionally robust optimization},
  author={Duchi, John C and Namkoong, Hongseok},
  journal={The Annals of Statistics},
  volume={49},
  number={3},
  pages={1378--1406},
  year={2021},
  publisher={Institute of Mathematical Statistics}
}

@article{zhang2023deep,
  title={Deep long-tailed learning: A survey},
  author={Zhang, Yifan and Kang, Bingyi and Hooi, Bryan and Yan, Shuicheng and Feng, Jiashi},
  journal={IEEE Transactions on Pattern Analysis and Machine Intelligence},
  volume={45},
  number={9},
  pages={10795--10816},
  year={2023},
  publisher={IEEE}
}

@article{he2009learning,
  title={Learning from imbalanced data},
  author={He, Haibo and Garcia, Edwardo A},
  journal={IEEE Transactions on knowledge and data engineering},
  volume={21},
  number={9},
  pages={1263--1284},
  year={2009},
  publisher={Ieee}
}

@article{johnson2019survey,
  title={Survey on deep learning with class imbalance},
  author={Johnson, Justin M and Khoshgoftaar, Taghi M},
  journal={Journal of big data},
  volume={6},
  number={1},
  pages={1--54},
  year={2019},
  publisher={Springer}
}

@book{quinonero2022dataset,
  title={Dataset shift in machine learning},
  author={Qui{\~n}onero-Candela, Joaquin and Sugiyama, Masashi and Schwaighofer, Anton and Lawrence, Neil D},
  year={2022},
  publisher={Mit Press}
}

@article{geirhos2020shortcut,
  title={Shortcut learning in deep neural networks},
  author={Geirhos, Robert and Jacobsen, J{\"o}rn-Henrik and Michaelis, Claudio and Zemel, Richard and Brendel, Wieland and Bethge, Matthias and Wichmann, Felix A},
  journal={Nature Machine Intelligence},
  volume={2},
  number={11},
  pages={665--673},
  year={2020},
  publisher={Nature Publishing Group UK London}
}

@inproceedings{koh2021wilds,
  title={Wilds: A benchmark of in-the-wild distribution shifts},
  author={Koh, Pang Wei and Sagawa, Shiori and Marklund, Henrik and Xie, Sang Michael and Zhang, Marvin and Balsubramani, Akshay and Hu, Weihua and Yasunaga, Michihiro and Phillips, Richard Lanas and Gao, Irena and others},
  booktitle={International conference on machine learning},
  pages={5637--5664},
  year={2021},
  organization={PMLR}
}

@inproceedings{kearns2018preventing,
  title={Preventing fairness gerrymandering: Auditing and learning for subgroup fairness},
  author={Kearns, Michael and Neel, Seth and Roth, Aaron and Wu, Zhiwei Steven},
  booktitle={International conference on machine learning},
  pages={2564--2572},
  year={2018},
  organization={PMLR}
}

@inproceedings{wu2023discover,
  title={Discover and cure: Concept-aware mitigation of spurious correlation},
  author={Wu, Shirley and Yuksekgonul, Mert and Zhang, Linjun and Zou, James},
  booktitle={International Conference on Machine Learning},
  pages={37765--37786},
  year={2023},
  organization={PMLR}
}

@article{codella2019skin,
  title={Skin lesion analysis toward melanoma detection 2018: A challenge hosted by the international skin imaging collaboration (isic)},
  author={Codella, Noel and Rotemberg, Veronica and Tschandl, Philipp and Celebi, M Emre and Dusza, Stephen and Gutman, David and Helba, Brian and Kalloo, Aadi and Liopyris, Konstantinos and Marchetti, Michael and others},
  journal={arXiv preprint arXiv:1902.03368},
  year={2019}
}

@inproceedings{liu2015deep,
  title={Deep learning face attributes in the wild},
  author={Liu, Ziwei and Luo, Ping and Wang, Xiaogang and Tang, Xiaoou},
  booktitle={Proceedings of the IEEE international conference on computer vision},
  pages={3730--3738},
  year={2015}
}

@inproceedings{he2016deep,
  title={Deep residual learning for image recognition},
  author={He, Kaiming and Zhang, Xiangyu and Ren, Shaoqing and Sun, Jian},
  booktitle={Proceedings of the IEEE conference on computer vision and pattern recognition},
  pages={770--778},
  year={2016}
}

@inproceedings{dosovitskiyimage,
  title={An Image is Worth 16x16 Words: Transformers for Image Recognition at Scale},
  author={Dosovitskiy, Alexey and Beyer, Lucas and Kolesnikov, Alexander and Weissenborn, Dirk and Zhai, Xiaohua and Unterthiner, Thomas and Dehghani, Mostafa and Minderer, Matthias and Heigold, Georg and Gelly, Sylvain and others},
  booktitle={International Conference on Learning Representations},
  year={2021},
}

@article{wah2011caltech,
  title={The caltech-ucsd birds-200-2011 dataset},
  author={Wah, Catherine and Branson, Steve and Welinder, Peter and Perona, Pietro and Belongie, Serge},
  year={2011},
  publisher={California Institute of Technology}
}

@article{zhou2017places,
  title={Places: A 10 million image database for scene recognition},
  author={Zhou, Bolei and Lapedriza, Agata and Khosla, Aditya and Oliva, Aude and Torralba, Antonio},
  journal={IEEE transactions on pattern analysis and machine intelligence},
  volume={40},
  number={6},
  pages={1452--1464},
  year={2017},
  publisher={IEEE}
}

@inproceedings{rolf2021representation,
  title={Representation matters: Assessing the importance of subgroup allocations in training data},
  author={Rolf, Esther and Worledge, Theodora T and Recht, Benjamin and Jordan, Michael},
  booktitle={International conference on machine learning},
  pages={9040--9051},
  year={2021},
  organization={PMLR}
}

@article{schrouff2024mind,
  title={Mind the graph when balancing data for fairness or robustness},
  author={Schrouff, Jessica and Bellot, Alexis and Rannen-Triki, Amal and Malek, Alan and Albuquerque, Isabela and Gretton, Arthur and D'Amour, Alexander and Chiappa, Silvia},
  journal={Advances in Neural Information Processing Systems},
  volume={37},
  pages={29913--29947},
  year={2024}
}

@article{alloula2025representation,
  title={Representation Invariance and Allocation: When Subgroup Balance Matters},
  author={Alloula, Anissa and Jones, Charles and Wakefield-Skorniewska, Zuzanna and Quinzan, Francesco and Papie{\.z}, Bart{\l}omiej},
  journal={arXiv preprint arXiv:2512.09496},
  year={2025}
}

@inproceedings{claucich2025fairness,
  title={Fairness of Deep Ensembles: On the interplay between per-group task difficulty and under-representation},
  author={Claucich, Estanislao and Hooker, Sara and Milone, Diego H and Ferrante, Enzo and Echeveste, Rodrigo},
  booktitle={Proceedings of the 2025 ACM Conference on Fairness, Accountability, and Transparency},
  pages={3138--3147},
  year={2025}
}

}

\appendix

\newpage
\appendix
\onecolumn

\section{Proof of Lemma 1}

    \noindent\textbf{Lemma 1. (Bayesian Group Accuracy)} \textit{For a subgroup specified by attributes $\va\in\{\pm1\}^N$, the group accuracy of
    a linear classifier $(\vw,b)$ is
    \begin{align}
        Acc_{\va}(\vw, b) = \frac{1}{2}\Big(1+\operatorname{erf}\big((\frac{\vmu_\va^T\vw + a_1 b}{\sqrt{2\vw^T\Sigma \vw}}\big)\Big)
    \end{align}
    where $\vw, b$ is the Bayesian optimal classifier solved based on training data statistics $(\vmu, \Sigma, \valpha, \beta)$, $a_n$ denotes the value of $n$-th attribute for this subgroup, and $m_n$ is the feature separability of the $n$-th feature.}

    \begin{proof}
        As we have $N$ attributes $a_1,\cdots,a_N$, there are totally $2^N$ groups, denoted as the binary attributes of each group $\va$. Let $r_\va$ denote the proportion of this subgroup, $\vmu_\va = [a_1 \vmu_1, \cdots,  a_N \vmu_N]$ is the corresponding conditional mean of the features for the training data. 
        The joint distribution of the data conditioned on $y$ can be written as
        \begin{equation}
        \begin{aligned}
        & p(\boldsymbol{z} \mid y=1)=\prod_{n=1}^N p\left(z_n \mid y=1\right) \\
        = & \frac{1}{\sqrt{(2 \pi)^d|\Sigma|}} \sum_{\va \in\{-1,1\}^N}\left\{r_\va \exp \left(-\frac{1}{2}\left(\boldsymbol{z}-\vmu_{\va}\right)^T \Sigma^{-1}\left(\boldsymbol{z}-\vmu_{\va}\right)^T\right)\right\} \\
        & p(\boldsymbol{z} \mid y=-1)=\prod_{n=1}^N p\left(z_n \mid y=-1\right) \\
        = & \frac{1}{\sqrt{(2 \pi)^d|\Sigma|}} \sum_{\va \in\{-1,1\}^N}\left\{r_\va \exp \left(-\frac{1}{2}\left(\boldsymbol{z} + \vmu_{\va}\right)^T \Sigma^{-1}\left(\boldsymbol{z} + \vmu_{\va}\right)^T\right)\right\}
        \end{aligned}
        \end{equation}
    
        Similar to \cite{wang2024effect}, the accuracy can be decomposed as the sum of true positive and true negative rates. Let $\Omega(\vz)=\{\vw \mid \vw^T \vz + b > 0 \}$ be the half-space of the positive prediction and the true positive rate is
        
        \begin{equation}
        \begin{aligned}
        & \int_{\Omega(\vz)} \frac{1}{\sqrt{(2 \pi)^d|\Sigma|}} \sum_{\va \in\{-1,1\}^N}\left\{r_\va \exp \left(-\frac{1}{2}\left(\vz-\vmu_{\va}\right)^T \Sigma^{-1}\left(\vz-\vmu_{\va}\right)^T\right)\right\} \mathrm{d}\vz \\
        =& \frac{1}{\sqrt{(2\pi)^d|\Sigma|}}\sum_{\va\in\{-1, 1\}^N}\bigg\{r_{\va}\int_{\{\vz|\vw^T\vz+b>0\}}\exp\Big(-\frac{1}{2}(\vz - \vmu_\va)^T\Sigma^{-1}(\vz - \vmu_\va)\Big)\bigg\}\rd\vz\\
        =& \frac{1}{\sqrt{(2\pi)^d|\Sigma|}}\sum_{\va\in\{-1, 1\}^N}\bigg\{r_{\va}\sqrt{|\Sigma|}\sqrt{(2\pi)^{d-1}}\sqrt{\frac{\pi}{2}}\big(1 - \operatorname{erf}(-\frac{\vmu_\va^T\vw + b}{\sqrt{2\vw^T\Sigma\vw}})\big)\\
        = & \frac{1}{2} \Big( 1 + \sum_{\va \in\{-1,1\}^N}\Big[r_\va \operatorname{erf} \Big(\frac{\vmu_\va^T\vw + b}{\sqrt{2\vw^T\Sigma\vw}}\Big)\Big] \Big)
        \end{aligned}
        \end{equation}
    \end{proof}

    Similarly, the true negative rate can be written as
    
    \begin{equation}
    \begin{aligned}
    & \int_{\sR^d \backslash \Omega(\vz)} \frac{1}{\sqrt{(2 \pi)^d|\Sigma|}} \sum_{\va \in\{-1,1\}^N}\left\{r_\va \exp \left(-\frac{1}{2}\left(\vz + \vmu_{\va}\right)^T \Sigma^{-1}\left(\vz + \vmu_{\va}\right)^T\right)\right\} \mathrm{d}\vz\\
    = & 1 - \int_{\Omega(\vz)} \frac{1}{\sqrt{(2 \pi)^d|\Sigma|}} \sum_{\va \in\{-1,1\}^N}\left\{r_\va \exp \left(-\frac{1}{2}\left(\vz + \vmu_{\va}\right)^T \Sigma^{-1}\left(\vz + \vmu_{\va}\right)^T\right)\right\} \mathrm{d}\vz\\
    = & \frac{1}{2} \Big( 1 - \sum_{\va \in\{-1,1\}^N}\Big[r_\va \operatorname{erf} \Big(\frac{-\vmu_\va^T\vw + b}{\sqrt{2\vw^T\Sigma\vw}}\Big)\Big] \Big)\\
    = & \frac{1}{2} \Big( 1 + \sum_{\va \in\{-1,1\}^N}\Big[r_\va \operatorname{erf} \Big(\frac{\vmu_\va^T\vw - b}{\sqrt{2\vw^T\Sigma\vw}}\Big)\Big] \Big)
    \end{aligned}
    \end{equation}

    Putting them together, we have the accuracy as
    \begin{equation}
        \begin{aligned}
            Acc(\vw, b) = & \frac{1}{2}\left(\int_{\Omega(\vw)} p(\vz \mid y=1) \mathrm{d} \vz+\int_{\mathbb{R}^d \backslash \Omega(\vw)} p(\vz \mid y=-1) \mathrm{d} \vz\right) \\
            =& \frac{1}{2} \Big( 1 + \sum_{\va\in\{\pm1\}^N}\Big[r_\va\operatorname{erf}\big(\frac{\vmu_\va ^T\vw + a_1 b}{\sqrt{2\vw^T\Sigma\vw}}\big)\Big] \Big) \\
        \end{aligned}
    \end{equation}
    with $a_1 \equiv y$. As $r_{\va}$ is the group proportion, the group accuracy thus is
    \begin{align}
        Acc_{\va}(\vw, b) = \frac{1}{2}\Big(1+\operatorname{erf}\big((\frac{\vmu_\va^T\vw + a_1 b}{\sqrt{2\vw^T\Sigma \vw}}\big)\Big)
    \end{align}

\section{Solution to Bayesian Optimal Classifier}
        The Bayesian optimal classifier has $\frac{\partial Acc}{\partial \vw}=0$ and $\frac{\partial Acc}{\partial b}=0$ on the training data, therefore, we have
        \begin{align}
            \frac{\partial acc}{\partial \vw} = (\Sigma\vw\vw^T-\vw^T\Sigma\vw I)\bigg\{&\sum_{\va\in\{\pm1\}^2}\Big[r_{\va}\exp\big(-\frac{(\vmu_\va^T\vw + a_1 b)^2}{2\vw^T\Sigma\vw}\big)\Big]\vmu_\va\bigg\}=0\\
        \end{align}
        We also have 
        \begin{align}
            (\Sigma\vw\vw^T)(\Sigma\vw) = (\Sigma\vw)(\vw^T\Sigma\vw) = (\vw^T\Sigma\vw)\Sigma\vw
        \end{align}
        therefore, $\exists \eta\in\sR$ s.t. 
        \begin{align}
            \Sigma\vw =& \eta\bigg\{\sum_{\va\in\{\pm1\}^2}\Big[r_{\va}\exp\big(-\frac{(\vmu_\va^T\vw + b)^2}{2\vw^T\Sigma\vw}\big)\Big]\vmu_\va\bigg\} \\
        \end{align}
        which indicates that the weights $w_i$ are collinear with $\Sigma_i^{-1}\mu_i$.
        
        Thus we can denote the weights of the Bayesian optimal classifier with $\vw_n=\eta_n \Sigma_n^{-1}\vmu_n$. Focusing on $N=2$, we can further denote the coefficients as 
        \begin{align}
            \tau = \eta_2 / \eta_1
            \label{eq:tau}
        \end{align}
        and the following equation by substituting the $\eta_1, \eta_2$ with terms from the Bayesian optimal classifier:
        \begin{align}
            \eta_1 =& \alpha_+\beta\exp\big(-\frac{(\vmu_\va^T\vw + b)^2}{2\vw^T\Sigma\vw}\big) + (1 - \alpha_+)\beta\exp\big(-\frac{(\vmu_\va^T\vw + b)^2}{2\vw^T\Sigma\vw}\big) \\
            & + (1-\alpha_-) (1 - \beta)\exp\big(-\frac{(\vmu_\va^T\vw - b)^2}{2\vw^T\Sigma\vw}\big) + \alpha_-(1-\beta)\exp\big(-\frac{(\vmu_\va^T\vw - b)^2}{2\vw^T\Sigma\vw}\big)\\
            \eta_2 =& \alpha_+\beta\exp\big(-\frac{(\vmu_\va^T\vw + b)^2}{2\vw^T\Sigma\vw}\big) - (1 - \alpha_+)\beta\exp\big(-\frac{(\vmu_\va^T\vw + b)^2}{2\vw^T\Sigma\vw}\big) \\
            & - (1-\alpha_-)(1-\beta)\exp\big(-\frac{(\vmu_\va^T\vw - b)^2}{2\vw^T\Sigma\vw}\big) + \alpha_-(1-\beta)\exp\big(-\frac{(\vmu_\va^T\vw - b)^2}{2\vw^T\Sigma\vw}\big)
        \end{align}
        
        In addition, we have another equation
        \begin{align}
            \frac{\partial Acc}{\partial b} =& \frac{1}{2}\sum_{\va\in\{\pm1\}^2}\Big[r_{\va}\exp\big(-\frac{(\vmu_\va^T\vw + a_1b)^2}{2\vw^T\Sigma\vw}\big)\Big] \\
            =&\alpha_+\beta\exp\big(-\frac{(\vmu_\va^T\vw + b)^2}{2\vw^T\Sigma\vw}\big) + (1 - \alpha_+)\beta\exp\big(-\frac{(\vmu_\va^T\vw + b)^2}{2\vw^T\Sigma\vw}\big) \\
            & - (1-\alpha_-) (1 - \beta)\exp\big(-\frac{(\vmu_\va^T\vw - b)^2}{2\vw^T\Sigma\vw}\big) - \alpha_-(1-\beta)\exp\big(-\frac{(\vmu_\va^T\vw - b)^2}{2\vw^T\Sigma\vw}\big)=0
            \label{eq:tau2}
        \end{align}

        Taking Eq.~\ref{eq:tau} and \ref{eq:tau2} as an equation system, we are able to solve for the numerical solution to $\tau, b$. With the solved classifier as $[\Sigma_1^{-1}\vmu_1, \tau \Sigma_2^{-1}\vmu_2], b$, we have the group accuracy simplified as
        \begin{align}
            Acc_{\va}(\vw, b) =& \frac{1}{2}\Big(1+\operatorname{erf}\big((\frac{\vmu_\va^T\vw + a_1 b}{\sqrt{2\vw^T\Sigma \vw}}\big)\Big) \\
            =& \frac{1}{2} \Big(1 + \operatorname{erf}{(\frac{a_1 m_1 + \tau a_2 m_2 + a_1 b}{\sqrt{2(m_1 + \tau^2 m_2)}})}\Big)
            \label{eq:acc-appd}
        \end{align}

        With the above notation for group accuracy, combining Eq.~\ref{eq:tau} and Eq.~\ref{eq:tau2}, the equation system to solve $\tau, b$ can be simplified as:

        \begin{equation}
            \left\{
            \begin{aligned}
                & \tau = \frac{\gamma_1 + \gamma_4 - \gamma_2 - \gamma_3}{\gamma_1 + \gamma_4 + \gamma_2 + \gamma_3} \\
                & \gamma_1 + \gamma_2 - (\gamma_3 + \gamma_4) = 0
            \end{aligned}
            \right.
            \label{eq:solve}
        \end{equation}

        where $\gamma$s are a series of functions of data parameters $m_1, m_2, \valpha, \beta$:
        \begin{equation}
            \left\{
            \begin{aligned}
                \gamma_1 &= \alpha_+ \beta \exp(- \frac{(m_1 + \tau m_2 + b)^2}{2(m_1 + \tau^2 m_2)}) \\
                \gamma_2 &= (1 - \alpha_+) \beta \exp(- \frac{(m_1 - \tau m_2 + b)^2}{2(m_1 + \tau^2 m_2)}) \\
                \gamma_3 &= (1 - \alpha_-) (1 - \beta) \exp(- \frac{(-m_1 + \tau m_2 + b)^2}{2(m_1 + \tau^2 m_2)}) = (1 - \alpha_-) (1 - \beta) \exp(- \frac{(m_1 - \tau m_2 - b)^2}{2(m_1 + \tau^2 m_2)})\\
                \gamma_4 &= \alpha_- (1 - \beta) \exp(- \frac{(-m_1 - \tau m_2 + b)^2}{2(m_1 + \tau^2 m_2)}) = \alpha_- (1 - \beta) \exp(- \frac{(m_1 + \tau m_2 - b)^2}{2(m_1 + \tau^2 m_2)})\\
            \end{aligned}
            \right.
        \end{equation}
        
        With the solved $\tau, b$, we can substitute them into Eq.~\ref{eq:acc-appd} to compute the group accuracy and further compute the adjusted accuracy for the test data as $\frac{1}{4}\sum_{\va \in\{-1,1\}^2}Acc_{\va}$.

    \section{Derivative of Adjusted Accuracy}~\label{appd:proposition1}

    With the adjusted accuracy $A = \frac{1}{4}\sum_{\va \in\{\pm1\}^2}Acc_{\va}$, we want to compute $\partial A / \partial p$, with $p \in \{\alpha_+, \alpha_-, \beta\}$.
    We denote the following notations to simplify the equations of $\gamma_i$:
    \begin{equation}
        \left\{
        \begin{aligned}
            D &= D(\tau) = m_1 + \tau^2 m_2\\
            N_1 &= N_1(\tau, b) = m_1 + \tau m_2 + b\\
            N_2 &= N_2(\tau, b) = m_1 - \tau m_2 + b\\
            N_3 &= N_3(\tau, b) = m_1 - \tau m_2 - b\\
            N_4 &= N_4(\tau, b) = m_1 + \tau m_2 - b\\
        \end{aligned}
        \right.
    \end{equation}
    
    Therefore, $\gamma$s can be written as:
    \begin{equation}
        \left\{
        \begin{aligned}
            \gamma_1 &= \alpha_+ \beta \exp(- \frac{N_1^2}{2D}) \\
            \gamma_2 &= (1 - \alpha_+) \beta \exp(- \frac{N_2^2}{2D}) \\
            \gamma_3 &= (1 - \alpha_-) (1 - \beta) \exp(- \frac{N_3^2}{2D})\\
            \gamma_4 &= \alpha_- (1 - \beta) \exp(- \frac{N_4^2}{2D}) \\
        \end{aligned}
        \right.
    \end{equation}

    Computing the partial derivative of the adjusted accuracy with respect to $p \in \{\alpha_+, \alpha_-, \beta\}$:
    \begin{equation}
        \begin{aligned}
            \frac{\partial A}{\partial p} &= \frac{1}{4}\sum_{\va \in \gA}\frac{\partial A_{\va}}    {\partial p} \\
            &= \frac{1}{4}\sum_{\va \in \gA}\frac{d A_{\va}}{d f} (\frac{\partial f}{\partial \tau} \frac{\partial \tau}{\partial p} + \frac{\partial f}{\partial b} \frac{\partial b}{\partial p})\\
            &= \frac{1}{4}\sum_{\va \in \gA}\frac{1}{\sqrt{\pi}} \exp(-f^2) (\frac{\partial f}{\partial \tau} \frac{\partial \tau}{\partial p} + \frac{\partial f}{\partial b} \frac{\partial b}{\partial p})\\
            &= \frac{1}{4}\sum_{\va \in \gA}\frac{1}{\sqrt{\pi}} \exp(-f^2) (\frac{m_2(a_2 m_1 - \tau a_1 (m_1 + b))}{D\sqrt{2D}} \frac{\partial \tau}{\partial p} + \frac{a_1}{\sqrt{2D}} \frac{\partial b}{\partial p})\\
        \end{aligned}
        \label{eq:grad}
    \end{equation}
    where we have 
    \begin{equation}
        \begin{aligned}
            f(\tau, b) &=\frac{a_1 m_1 + \tau a_2 m_2 + a_1 b}{\sqrt{2(m_1 + \tau^2 m_2)}}\\
            &=\frac{a_1 m_1 + \tau a_2 m_2 + a_1 b}{\sqrt{2D}}   
        \end{aligned}
    \end{equation}
    
    In Eq.~\ref{eq:grad}, given data parameters $m_1, m_2, \vmu, \Sigma$, we are able to solve the numerical solution for $\tau, b$ with Eq.~\ref{eq:solve}, and the unknown variables are only $\frac{\partial \tau}{\partial p}$ and $\frac{\partial b}{\partial p}$.

    Considering the equation system that solves $\tau, b$ as
    \begin{equation}
        \left\{
        \begin{aligned}
            F_1(\tau, b; \alpha_+, \alpha_-, \beta) &= \tau (\gamma_1 + \gamma_4 + \gamma_2 + \gamma_3) - (\gamma_1 + \gamma_4 - \gamma_2 - \gamma_3) = 0 \\
            &= (\tau - 1)(\gamma_1 + \gamma_4) + (\tau + 1)(\gamma_2 + \gamma_3) = 0\\
            F_2(\tau, b; \alpha_+, \alpha_-, \beta) &= \gamma_1 + \gamma_2 - \gamma_4 - \gamma_3 = 0
        \end{aligned}
        \right.
    \end{equation}

    We compute the derivatives for the equation system and use the equation system of the derivatives to solve $\frac{\partial \tau}{\partial p}$ and $\frac{\partial b}{\partial p}$:    
    \begin{equation}
        \left\{
        \begin{aligned}
            & \frac{\partial F_1}{\partial \tau} \frac{\partial \tau}{\partial p} + \frac{\partial F_1}{\partial b} \frac{\partial b}{\partial p} + \frac{\partial F_1}{\partial p} = 0 \\
            & \frac{\partial F_2}{\partial \tau} \frac{\partial \tau}{\partial p} + \frac{\partial F_2}{\partial b} \frac{\partial b}{\partial p} + \frac{\partial F_2}{\partial p} = 0 \\
        \end{aligned}
        \right.
        \label{eq:dev}
    \end{equation}
    Except for $\frac{\partial \tau}{\partial p}$ and $\frac{\partial b}{\partial p}$, we have the following needed derivatives:
    \begin{equation}
        \begin{aligned}
            \frac{\partial F_1}{\partial \tau} &= (\tau - 1)(\frac{\partial \gamma_1}{\partial \tau} + \frac{\partial \gamma_4}{\partial \tau}) + (\tau + 1)(\frac{\partial \gamma_2}{\partial \tau} + \frac{\partial \gamma_3}{\partial \tau}) + (\gamma_1 + \gamma_4 + \gamma_2 + \gamma_3)\\
            &= \frac{(\tau - 1)m_2}{D^2}[\gamma_1 N_1(\tau N_1 - D) + \gamma_4 N_4(\tau N_4 - D)] + \frac{(\tau + 1)m_2}{D^2}[\gamma_2 N_2(\tau N_2 - D) + \gamma_3 N_3(\tau N_3 - D)]\\
            &+ (\gamma_1 + \gamma_4 + \gamma_2 + \gamma_3)\\
            \frac{\partial F_2}{\partial \tau} &= \frac{\partial \gamma_1}{\partial \tau} + \frac{\partial \gamma_2}{\partial \tau} - \frac{\partial \gamma_3}{\partial \tau} - \frac{\partial \gamma_4}{\partial \tau} \\
            &= \frac{m_2}{D^2}[\gamma_1 N_1(\tau N_1 - D) + \gamma_2 N_2(\tau N_2 - D) - \gamma_3 N_3(\tau N_3 - D) - \gamma_4 N_4(\tau N_4 - D)] \\
            \frac{\partial F_1}{\partial b} &= (\tau - 1)(\frac{\partial \gamma_1}{\partial b} + \frac{\partial \gamma_4}{\partial b}) + (\tau + 1)(\frac{\partial \gamma_2}{\partial b} + \frac{\partial \gamma_3}{\partial b}) \\
            &= \frac{(\tau - 1)(-\gamma_1 N_1 + \gamma_4 N_4) + (\tau + 1)(-\gamma_2 N_2 + \gamma_3 N_3)}{D} \\
            \frac{\partial F_2}{\partial b} &= \frac{\partial \gamma_1}{\partial b} + \frac{\partial \gamma_2}{\partial b} - \frac{\partial \gamma_3}{\partial b} - \frac{\partial \gamma_4}{\partial b} \\
            &= \frac{-\gamma_1 N_1 -\gamma_2 N_2 - \gamma_3 N_3 - \gamma_4 N_4}{D} \\
            \frac{\partial F_1}{\partial \beta} &= (\tau - 1)(\frac{\partial \gamma_1}{\partial \beta} + \frac{\partial \gamma_4}{\partial \beta}) + (\tau + 1)(\frac{\partial \gamma_2}{\partial \beta} + \frac{\partial \gamma_3}{\partial \beta}) \\
            &= (\tau - 1)[\alpha_+ \exp(- \frac{N_1^2}{2D}) - \alpha_- \exp(- \frac{N_4^2}{2D})] + (\tau + 1)[(1 - \alpha_+)\exp(- \frac{N_2^2}{2D}) - (1 - \alpha_-)\exp(- \frac{N_3^2}{2D})]\\
            \frac{\partial F_2}{\partial \beta} &= \frac{\partial \gamma_1}{\partial \beta} + \frac{\partial \gamma_2}{\partial \beta} - \frac{\partial \gamma_4}{\partial \beta} - \frac{\partial \gamma_3}{\partial \beta}\\
             &= \alpha_+ \exp(- \frac{N_1^2}{2D}) + (1 - \alpha_+)\exp(- \frac{N_2^2}{2D}) + \alpha_- \exp(- \frac{N_4^2}{2D}) + (1 - \alpha_-)\exp(- \frac{N_3^2}{2D})\\
            \frac{\partial F_1}{\partial \alpha_+} &= (\tau - 1)\frac{\partial \gamma_1}{\partial \alpha_+} + (\tau + 1)\frac{\partial \gamma_2}{\partial \alpha_+} = (\tau - 1) \beta \exp(- \frac{N_1^2}{2D}) - (\tau + 1) \beta \exp(- \frac{N_2^2}{2D}) \\
            \frac{\partial F_2}{\partial \alpha_+} &= \frac{\partial \gamma_1}{\partial \alpha_+} + \frac{\partial \gamma_2}{\partial \alpha_+} = \beta \exp(- \frac{N_1^2}{2D}) - \beta \exp(- \frac{N_2^2}{2D}) \\
            \frac{\partial F_1}{\partial \alpha_-} &= (\tau - 1)\frac{\partial \gamma_4}{\partial \alpha_-} + (\tau + 1)\frac{\partial \gamma_3}{\partial \alpha_-} = (\tau - 1) (1 - \beta) \exp(- \frac{N_4^2}{2D}) - (\tau + 1) (1 - \beta) \exp(- \frac{N_3^2}{2D}) \\
            \frac{\partial F_2}{\partial \alpha_-} &= -\frac{\partial \gamma_4}{\partial \alpha_-} - \frac{\partial \gamma_3}{\partial \alpha_-} = -(1 - \beta) \exp(- \frac{N_4^2}{2D}) + (1 - \beta) \exp(- \frac{N_3^2}{2D}) \\
        \end{aligned}
    \end{equation}

\section{Dataset Information}~\label{appd:dataset}
    Following the previous work~\cite{yang2023change}, we evaluate our theoretical insights on three image datasets and one text dataset, whose group configurations exhibit mixed subpopulation shifts. The image datasets include Waterbirds~\cite{sagawa2019distributionally}, UrbanCars~\cite{li2023whac} and CelebA~\cite{liu2015deep}; and the text dataset is CivilComments~\cite{borkan2019nuanced}. 
    To enable an empirical analysis of comprehensive training configurations, we construct subsets for CelebA and CivilComments, denoted as CelebA-subset and CivilComments-subset. CivilComments has 7 values for the spurious attribute, and following \cite{yang2023change}, we use the coarse labeling that considers the spurious attributes as binary: with or without identities.
    We sampled 2000 and 10000 data for the CelebA-subset and CivilComments-subset, respectively. For the experiments of data allocation in Section 4.3, we add 1000/5000 data for every 5 allocation steps, resulting in 7000 and 35000 after a total of 25 allocation steps. The data allocation starts with a fixed initial subset, and adds/removes samples to make the resulting new subset align with the new number of samples and the new configurations updated by different directions.
    
    As Waterbirds and UrbanCars are synthesized from images of foregrounds and backgrounds, we adopt a similar scheme to generate variants that allow flexibility in group configurations. We refer to them as Waterbirds-syn and UrbanCars-syn. For Waterbirds, we generate the variant using segmented birds as foregrounds from the Caltech-UCSD Birds-200-2011 (CUB) dataset~\cite{wah2011caltech} and backgrounds from the Places dataset~\cite{zhou2017places}. Similarly, for UrbanCars, we use foregrounds from Stanford Cars~\cite{krause20133d} and backgrounds from Places~\cite{zhou2017places}. For experiments in Sections 4.2 and 4.4, we create the variants according to the configurations sampled from the 3-simplex coordinates. Regarding the setup in Section 4.3, we fix the initial configuration by fixing the sampled foregrounds and backgrounds, and add/remove a corresponding number of samples to create new variants for data allocation. We add 500 samples for every 5 allocation steps, which results in 3500 in total at the last step for UrbanCars-syn. The Waterbirds-syn has a total of 9000 data to vary the configurations due to its challenging nature.

    For these datasets, we train an ERM on the standard datasets to obtain the group accuracy on test data and estimate the feature separability that is used for direction estimation. With the training configuration $(\alpha_+, \alpha_-, \beta)$ and four empirical group accuracy, we solve the numerical solution of $m_1, m_2, \tau, b$ from an equation system of four group-accuracy equations.
    The estimated separability is provided in the following table:

    \begin{table}[h]
    \begin{center}
    \caption{Estimated Separability for Datasets}    
    \setlength\tabcolsep{3pt}
    \resizebox{0.5\linewidth}{!}{
    \begin{tabular}{lcccc}
    \toprule
    & Waterbirds-syn & UrbanCars-syn & CelebA-subset & CivilComments-subset \\
    \midrule
    $m_1$ & 2.142 & 0.567 & 1.903 & 0.977\\
    $m_2$ & 8.0 & 0.789 & 2.0 & 2.0\\
    \bottomrule
    \end{tabular}
   }
    \end{center}
    \end{table}

\section{Comparison to~\cite{wang2024effect}}\label{appd:related}

    While our work uses the analytical framework from \cite{wang2024effect}, our question, scope, and contributions are substantially different from it. In the context of spurious correlation, \cite{wang2024effect} analyzes feature separability as a key factor affecting performance and proposes a mitigation method. In contrast, we study a broader, data-centric question under subpopulation shifts: how to optimize subgroup configurations for robust performance.
    
    This shift in focus leads to three contributions beyond \cite{wang2024effect}. First, we identify a counterintuitive phenomenon of multiple optimal configurations, including skewed ones far from balance. Though \cite{wang2024effect} studies performance, it does not explicitly examine the role of data configuration, and its results in the spurious-correlation setting instead suggest that more balanced data is generally preferable. Second, we derive a principled allocation direction towards the most effective robust performance. Our analysis further shows that it can diverge from the balancing direction. This addresses an important question that is beyond the scope of \cite{wang2024effect}: Under its pure spurious-correlation setting, the resulting direction is degenerate, thus does not yield more effective guidance. Lastly, we translate our theoretically derived directions into a practical data-allocation method that can outperform simply moving toward balance across multiple datasets. 

\section{Training and Test Details}~\label{appd:exp_detail}
    Regarding the model structure, for image-based datasets, we chose the pretrained ResNet50~\cite{he2016deep} with a linear layer to map representation vectors to probabilities for two classes. For ResNet50, we use the SGD optimizer with a learning rate of 1e-3, a weight decay of 1e-4, and a momentum of 0.9. The batch size is set as 256, and the number of training steps for the basic setup is 401, 501, and 201 for Waterbirds-syn, UrbanCars-syn, and CelebA-subset, respectively. Regarding the CivilComments-subset, we used BERT~\cite{devlin2019bert}, applying AdamW optimizer with differential weight decay, a standard choice for fine-tuning BERT. The learning rate is set as 3e-5, and the weight decay is 0.01. A linear scheduler is applied without warmup. The batch size is 64 and the number of training steps is 251 for the basic subset. For the data allocation experiments, the number of training steps increases proportionally with the number of samples. The cross-entropy loss was applied for these experiments. The validation test data are adopted from the original datasets and we evaluate the adjusted accuracy or worst-group accuracy on them. The experiments are conducted on a single GPU, NVIDIA RTX A5000 (25GB memory) or NVIDIA GeForce RTX 3090 (25GB). The experiments on the same dataset are conducted on the same GPU.

\section{Implementation Details and Code Availability}~\label{appd:implementation}
    The corresponding mapping between the group proportions $r_{\va}$ and training configuration $(\alpha_+,\alpha_-,\beta)$ is described as
    \begin{equation}
        \begin{aligned}
            &\alpha_+ = \frac{r_{1,1}}{r_{1,1} + r_{1,-1}} \\
            &\alpha_- = \frac{r_{-1,-1}}{r_{-1,-1} + r_{-1,1}} \\
            &\beta = r_{1,1} + r_{1,-1} \\
        \end{aligned}
    \end{equation}

    For the directions of the gradient (Grad) or the gradient sign (Sign), we first estimate the parameters $\tau, b$ for the Bayesian optimal classifier using Eq.\ref{eq:solve} and compute the known derivatives in the equation system~\ref{eq:dev} with the estimated values for $\tau, b$. Further, we solve the key derivatives $\frac{\partial \tau}{\partial p}$ and $\frac{\partial b}{\partial p}$ and substitute them into Eq.\ref{eq:grad} to compute the gradients for each parameter. The direction for the Sign is then obtained via the signs of the gradients.
    Regarding the direction of balancing, we compute the Jacobian matrix $J$, the derivative of the group sizes $r_{\va} \in \sR^4$ with respect to each parameter as, 
    \begin{equation}
        \begin{aligned}
            J & = 
            \begin{pmatrix}
            \frac{\partial r_{1,1}}{\partial \alpha_+} & \frac{\partial r_{1,1}}{\partial \alpha_-} & \frac{\partial r_{1,1}}{\partial \beta} \\
            \frac{\partial r_{1,-1}}{\partial \alpha_+} & \frac{\partial r_{1,-1}}{\partial \alpha_-} & \frac{\partial r_{1,-1}}{\partial \beta} \\
            \frac{\partial r_{-1,1}}{\partial \alpha_+} & \frac{\partial r_{-1,1}}{\partial \alpha_-} & \frac{\partial r_{-1,1}}{\partial \beta} \\
            \frac{\partial r_{-1,-1}}{\partial \alpha_+} & \frac{\partial r_{-1,-1}}{\partial \alpha_-} & \frac{\partial r_{-1,-1}}{\partial \beta} \\
            \end{pmatrix}\\
            & = 
            \begin{pmatrix}
            \beta & 0 & \alpha_+ \\
            -\beta & 0 & 1 - \alpha_+ \\
            0 & -(1 - \beta) & -(1 - \alpha_-) \\
            0 & 1 - \beta & -\alpha_-
            \end{pmatrix}
        \end{aligned}
    \end{equation}
    And we map the direction towards the balance from the original group sizes to the direction represented by $\valpha, \beta$ using $J$:
    \begin{equation}
        \begin{aligned}
            J^T([0.25,0.25,0.25,0.25]^T - r_{\va})
        \end{aligned}
    \end{equation}
    Such an implementation aims to normalize the vectors of these directions in the parameter space to ensure a comparable scale in parameter updating. Algorithm~\ref{alg} summarizes how these directions are computed and used to update the training configuration.
    We have included the details for theoretical and empirical implementation in our main paper and appendix.

\section{3-simplex Visualization}\label{appd:simplex}
    
    \subsection{Cosine Similarity between Directions}

\begin{figure}[t!]
    \centering
    \begin{minipage}[t]{0.74\linewidth}
        \centering
        \begin{subfigure}{0.32\linewidth}
            \includegraphics[width=\linewidth]{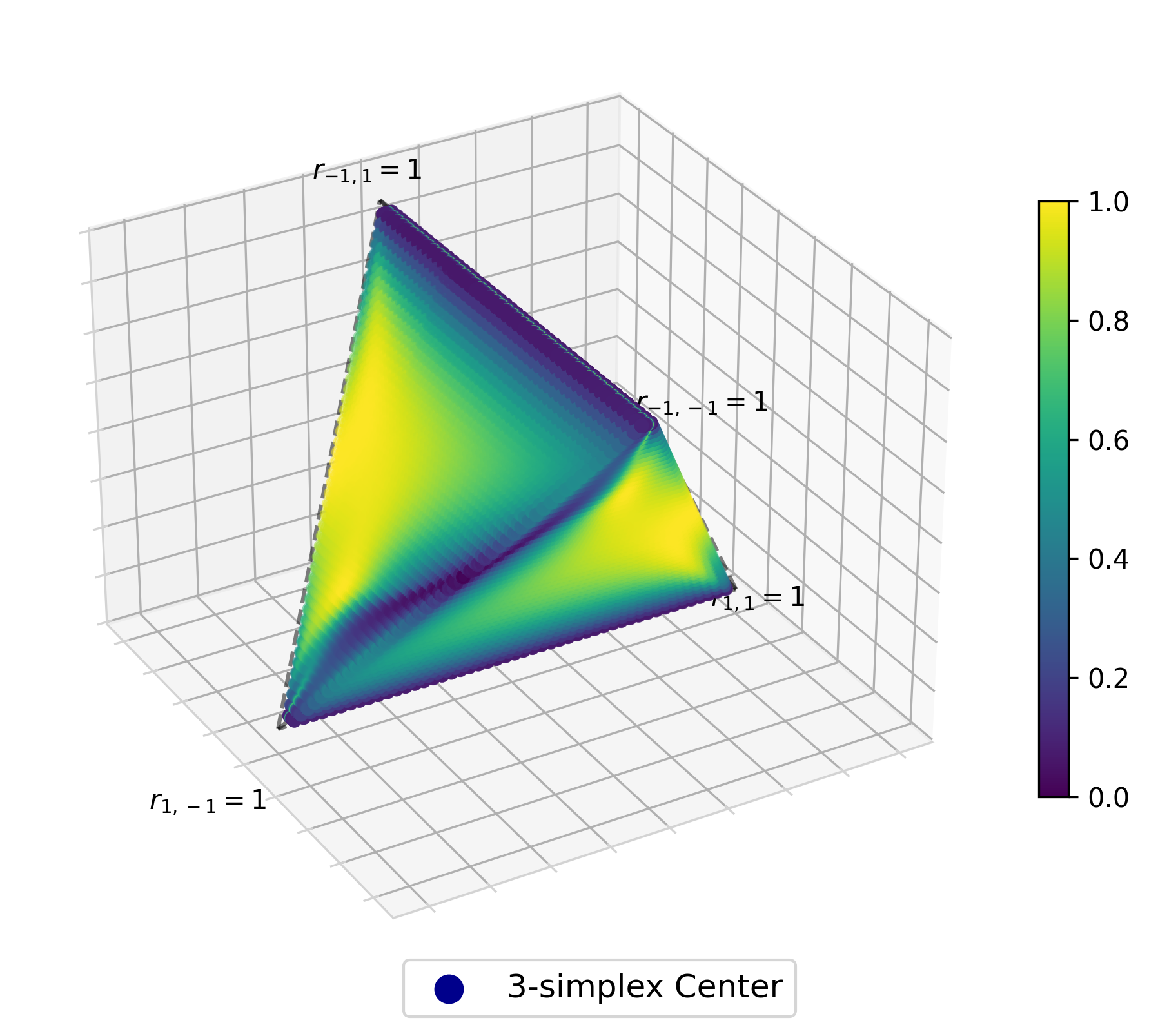}
        \end{subfigure}
        \hfill
        \begin{subfigure}{0.32\linewidth}
            \includegraphics[width=\linewidth]{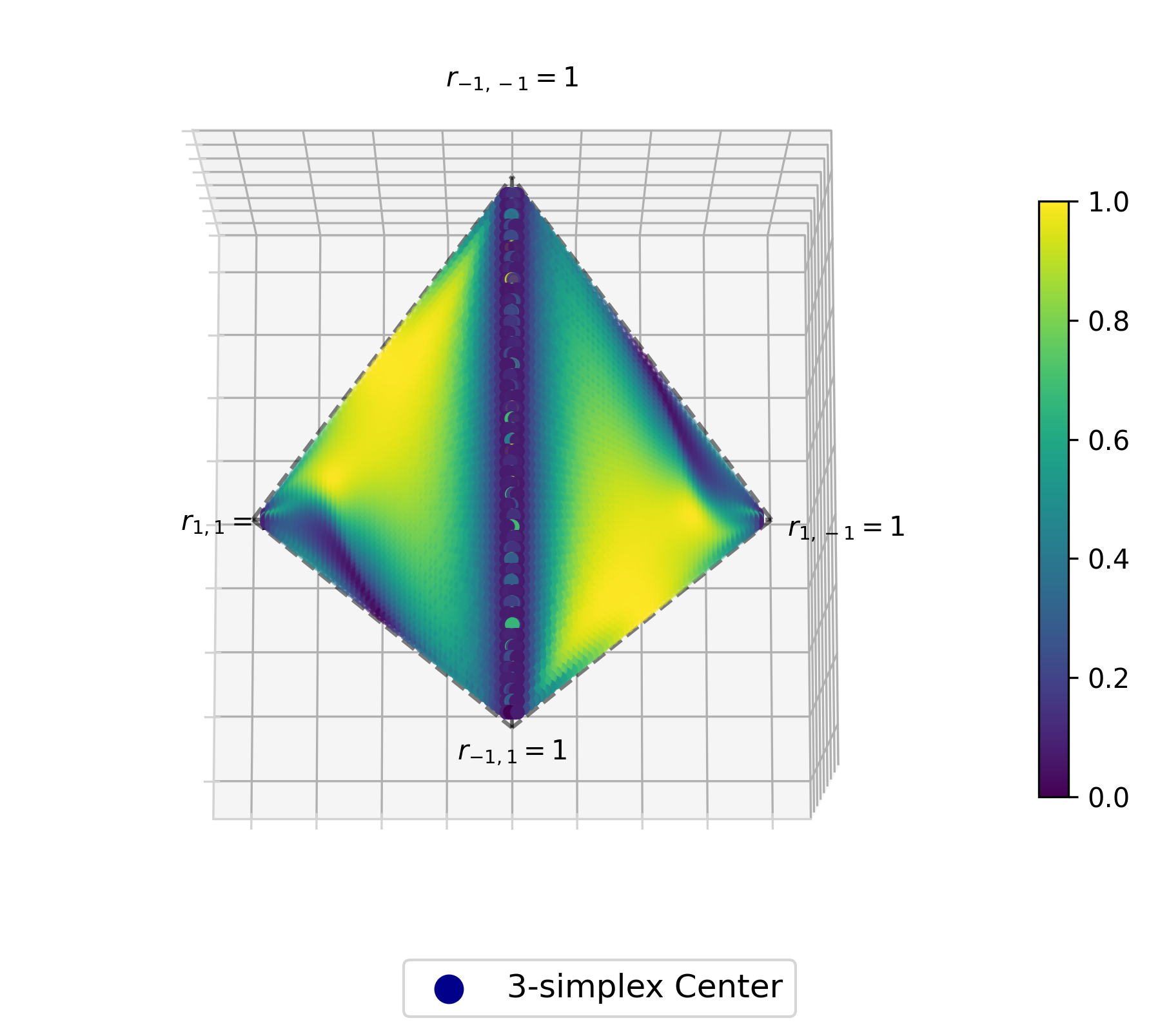}
        \end{subfigure}
        \hfill
        \begin{subfigure}{0.32\linewidth}
            \includegraphics[width=\linewidth]{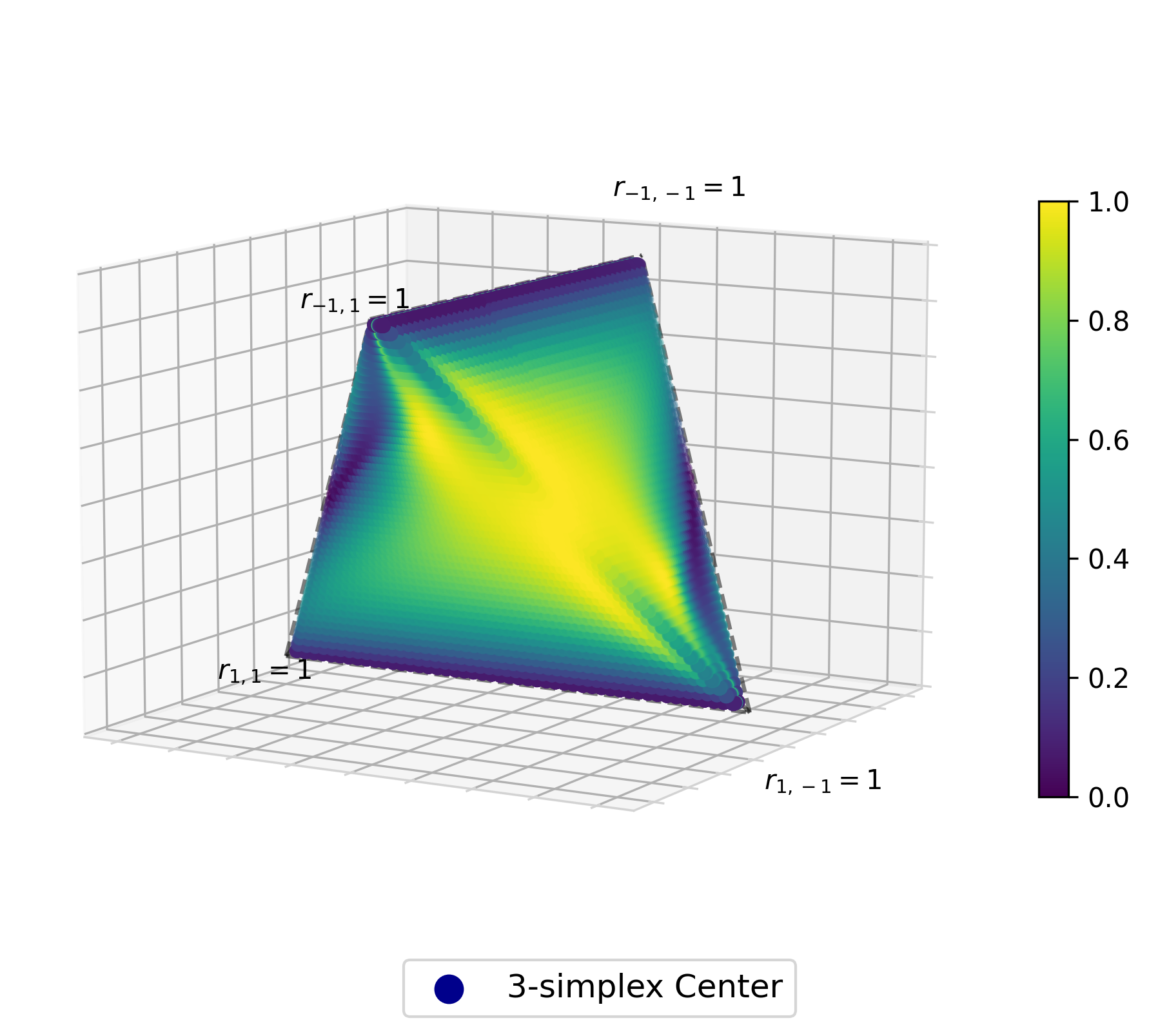}
        \end{subfigure}
        \caption{Different Views of the Cosine Similarity 3-simplex.}
        \label{appd:fig:simplex}
    \end{minipage}
    \hfill
    \begin{minipage}[t]{0.24\linewidth}
        \centering
        \includegraphics[width=\linewidth]{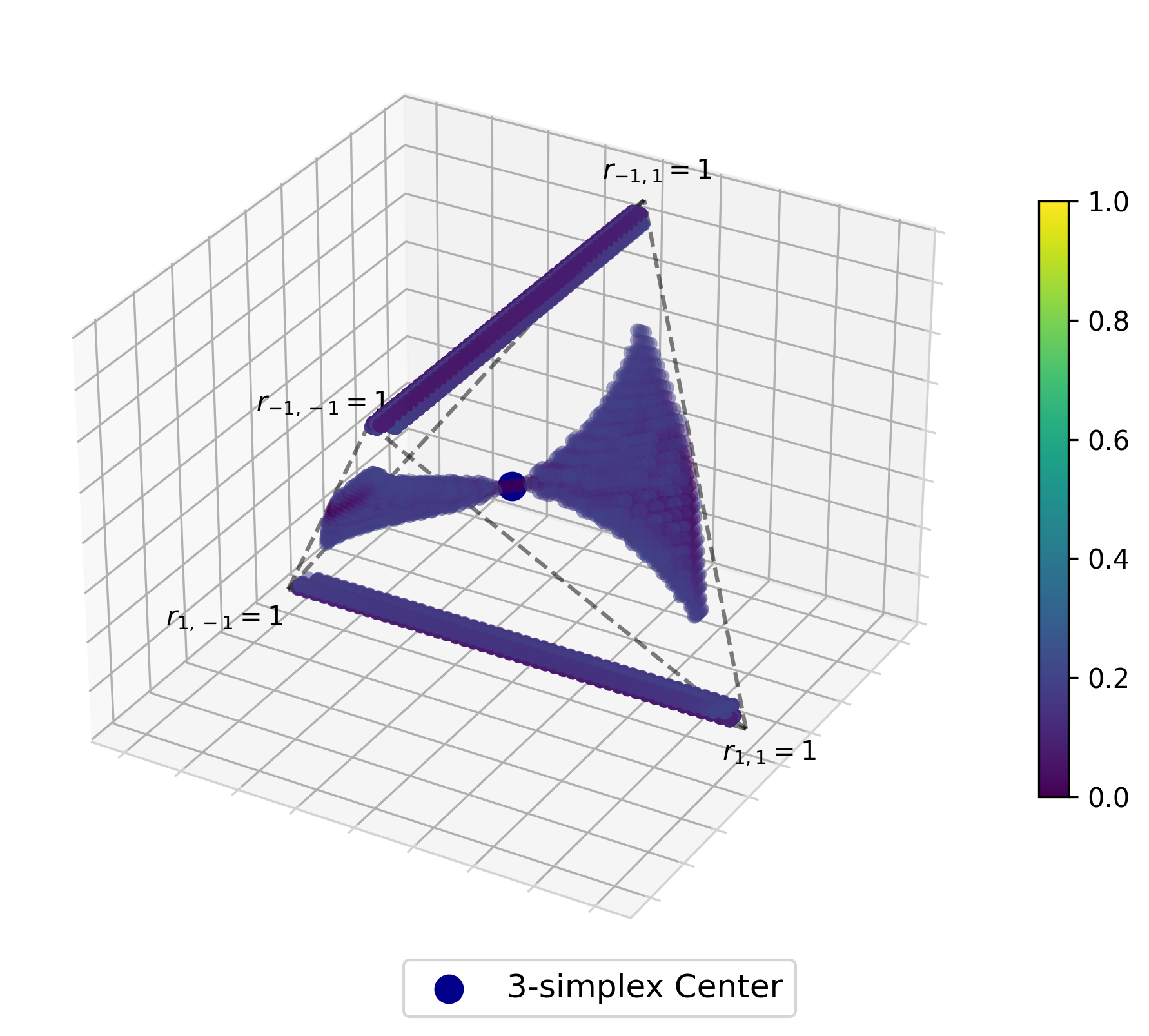}
        \caption{Areas of Low Grad-Bal Similarity.}
        \label{appd:fig:low-sim}
    \end{minipage}
\end{figure}

    We showed visualizations in Figure~\ref{fig:sim} that depict the similarity between directions estimated via Gradients (Grad) and Balancing (Bal). These are provided from two viewpoints to better capture the structure. We further present the same 3-simplex from various angles to highlight its symmetrical architecture in Figure~\ref{appd:fig:simplex}. For example, symmetric edges such as $r_{1,-1}=1, r_{1,1}=1$ and $r_{-1,-1}, r_{-1,1}$ are visible in Fig. 3, corresponding to cases where all data points share the same ground-truth label (1 or -1). The visualizations in Figure~\ref{appd:fig:simplex} primarily focus on views along these key edges.

    For the purpose of better visualization, we filter the similarity that is smaller than a threshold (0.2) to show the area where the two directions differ significantly in Figure~\ref{appd:fig:low-sim}. We observe that the configurations with low similarity between Grad and Bal are located near the two edges for severe class imbalance (the edge between $r_{-1,-1}=1$ and $r_{-1,1}=1$, and the edge between $r_{1,-1}=1$ and $r_{1,1}=1$), and two symmetric sectors where spurious correlation exhibits.

    \subsection{Initial Configurations for Data Allocation}
\begin{figure}
    \centering
    \includegraphics[width=0.3\linewidth]{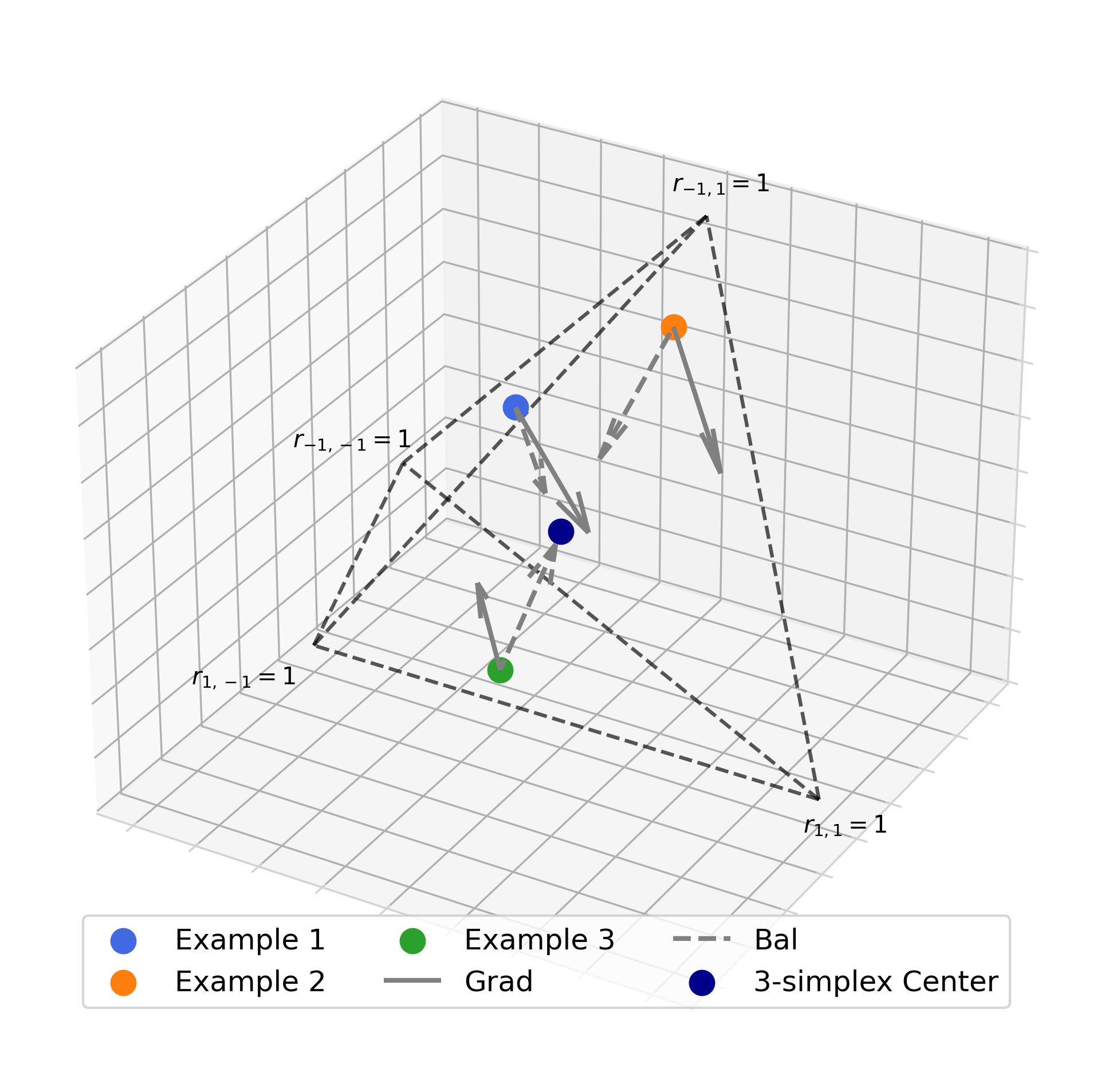}
    \caption{Comparison of the Grad and Bal directions for initial configurations in the 3-simplex.}
    \label{appd:fig:alloc}
\end{figure}

    We use three initial configurations for the experiments on data allocation. The detailed configurations are listed in the following table, and we visualize the directions of Grad and Bal along with the configurations in the 3-simplex in Figure~\ref {appd:fig:alloc}.

    \begin{table}[h]
    \begin{center}
    \caption{Initial Configurations for Data Allocation Experiments in Section 4.3}    
    \setlength\tabcolsep{3pt}
    \resizebox{0.3\linewidth}{!}{
    \begin{tabular}{lccc}
    \toprule
    & $\alpha_+$ & $\alpha_-$ & $\beta$ \\
    \midrule
    Example 1 & 0.848 & 0.695 & 0.059\\
    Example 2 & 0.976 & 0.184 & 0.128\\
    Example 3 & 0.397 & 0.883 & 0.862\\
    \bottomrule
    \end{tabular}
   }
    \end{center}
    \end{table}

    \subsection{Optimality in Multiple Configurations}~\label{appd:opt}
\begin{figure*}[h]
    \centering
    \begin{subfigure}{0.24\linewidth}
        \includegraphics[width=\linewidth]{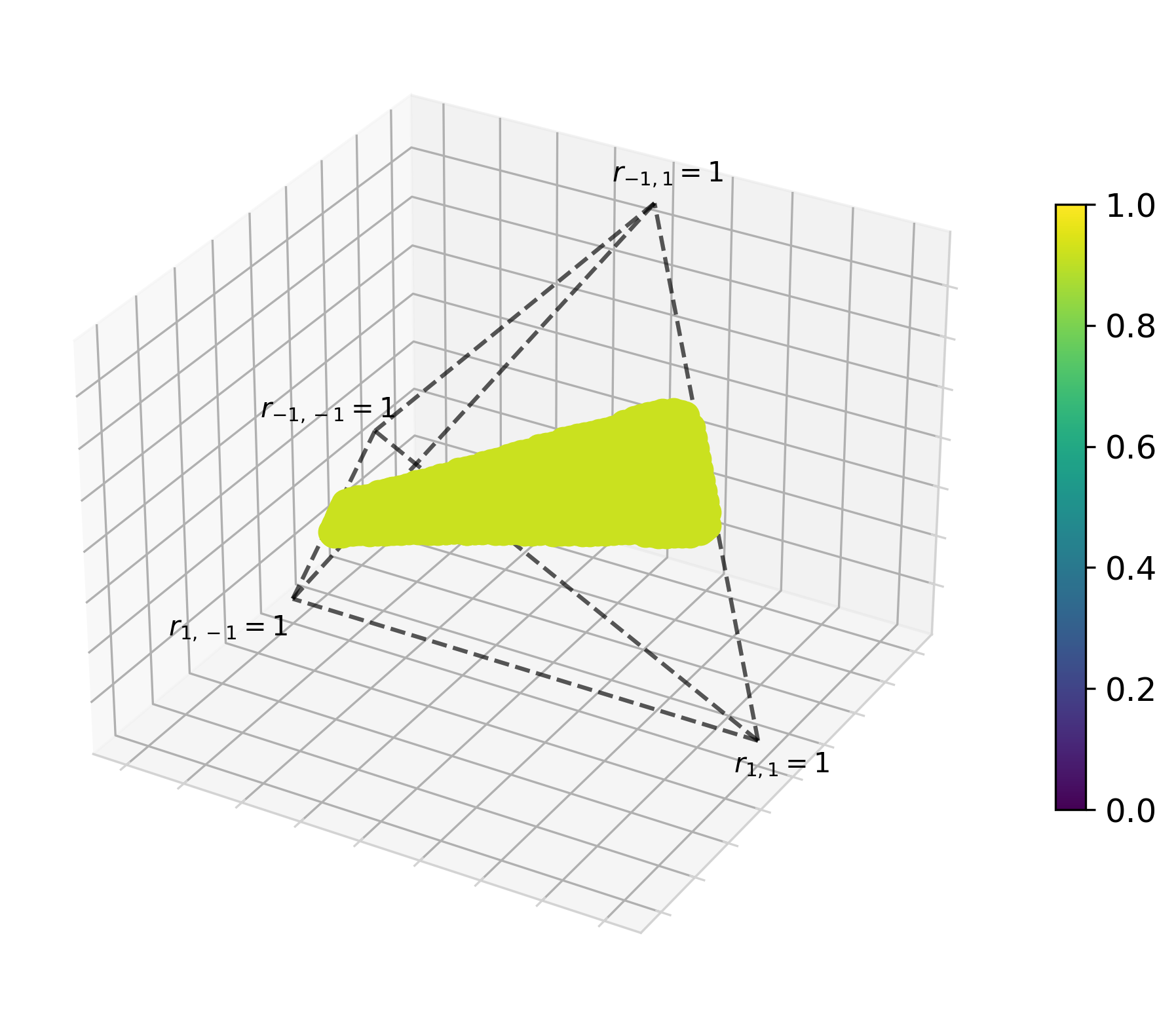}
        \caption{$n=50$, 95th percentile}
    \end{subfigure}
    \hfill
    \begin{subfigure}{0.24\linewidth}
        \includegraphics[width=\linewidth]{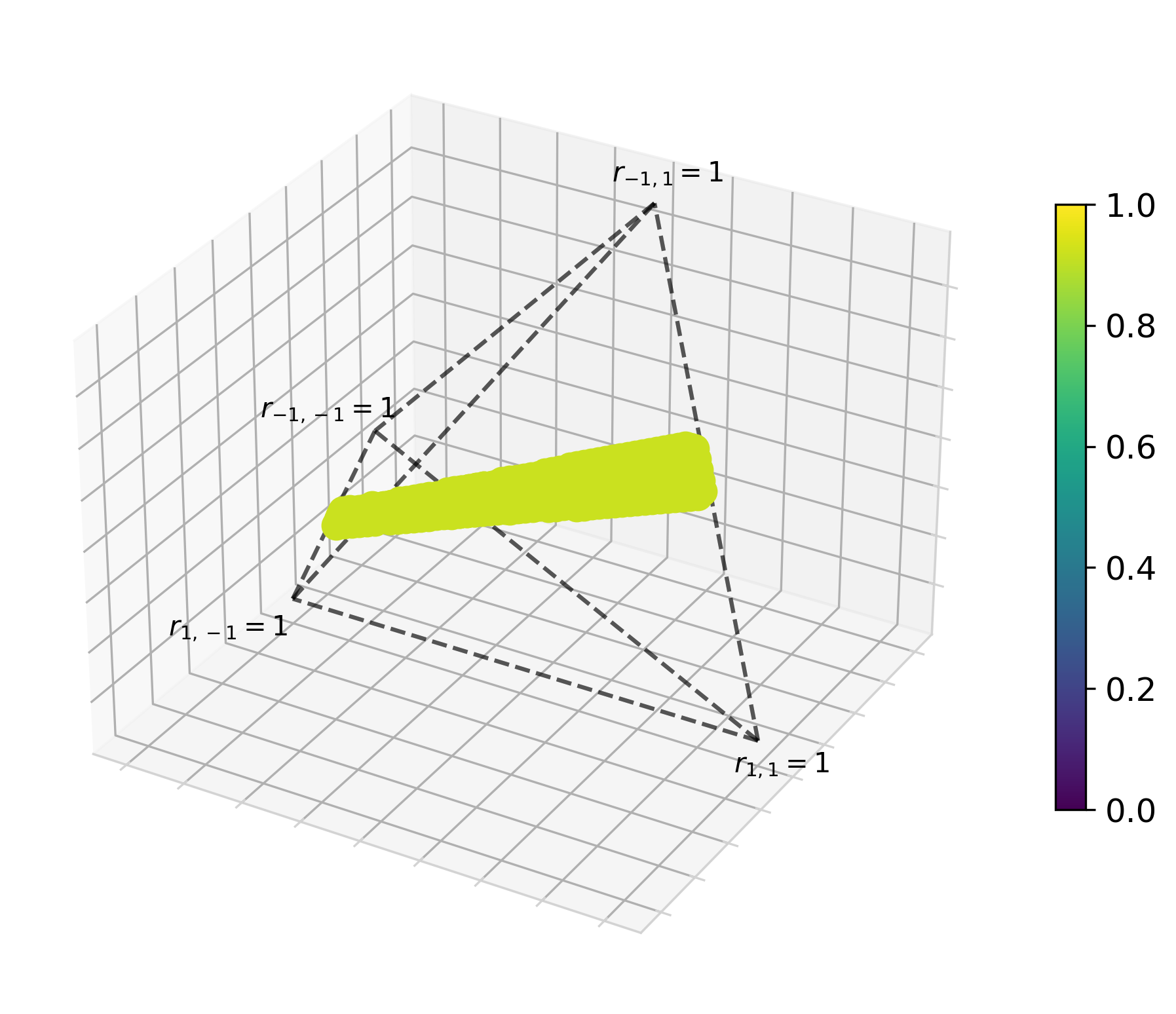}
        \caption{$n=50$, 99th percentile}
    \end{subfigure}
    \hfill
    \begin{subfigure}{0.24\linewidth}
        \includegraphics[width=\linewidth]{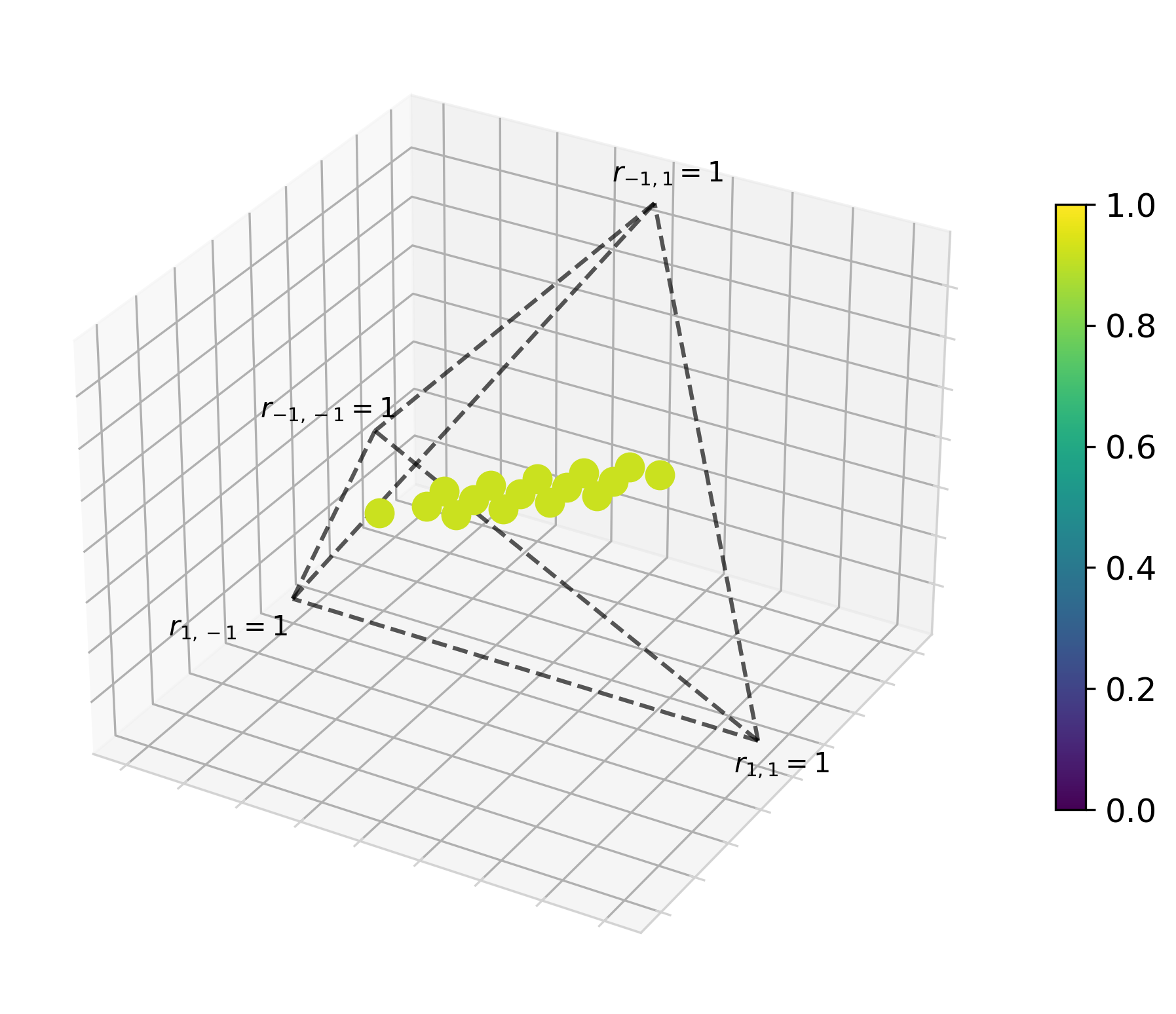}
        \caption{$n=7$, 95th percentile}
    \end{subfigure}
    \hfill
    \begin{subfigure}{0.24\linewidth}
        \includegraphics[width=\linewidth]{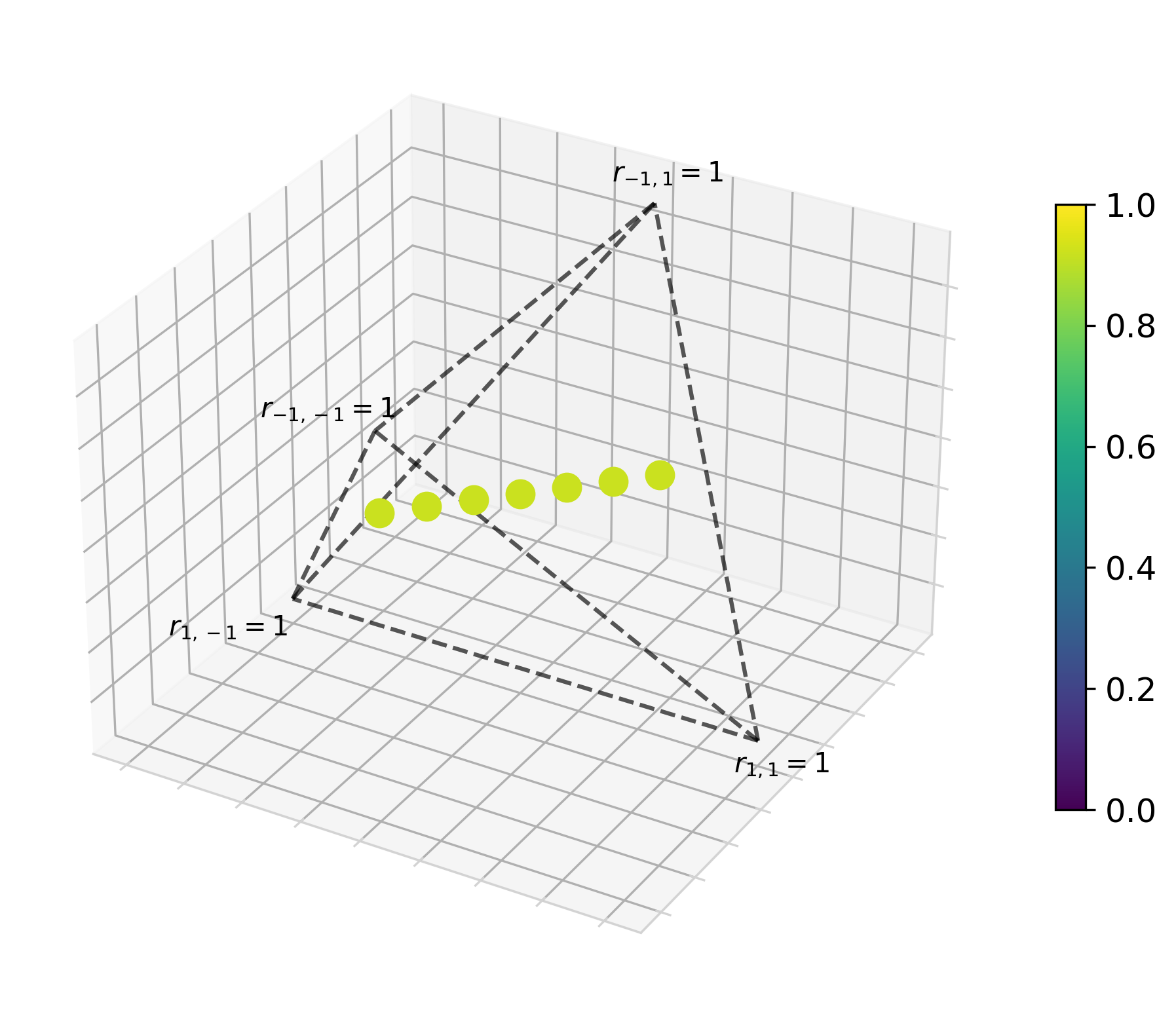}
        \caption{$n=7$, 99th percentile}
    \end{subfigure}
    \caption{Optimality of theoretical test performance.}
    \label{appd:fig:opt0}
\end{figure*}

\begin{figure*}[h]
    \centering
    \begin{subfigure}{0.24\linewidth}
        \includegraphics[width=\linewidth]{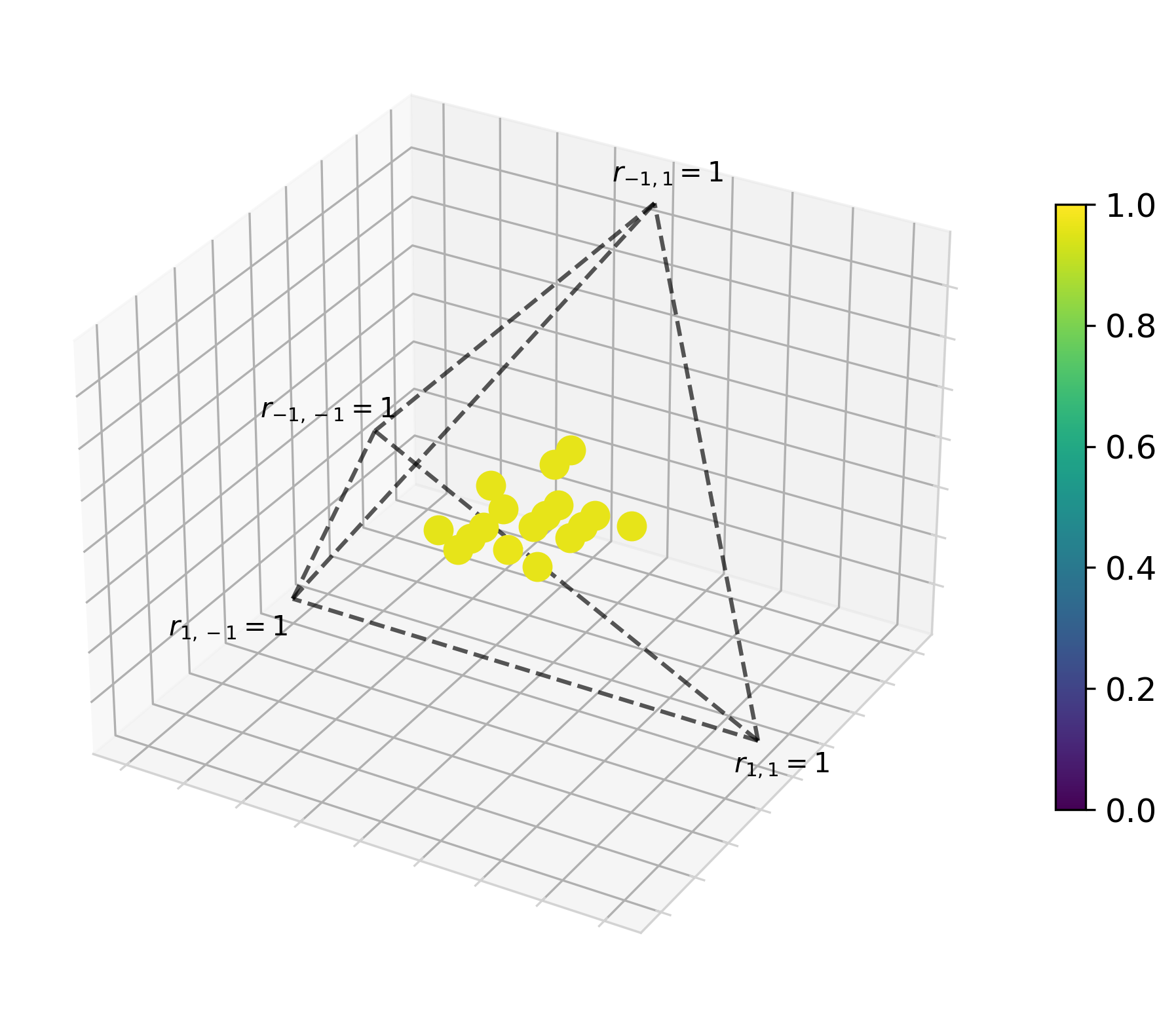}
        \caption{Waterbirds-syn}
    \end{subfigure}
    \hfill
    \begin{subfigure}{0.24\linewidth}
        \includegraphics[width=\linewidth]{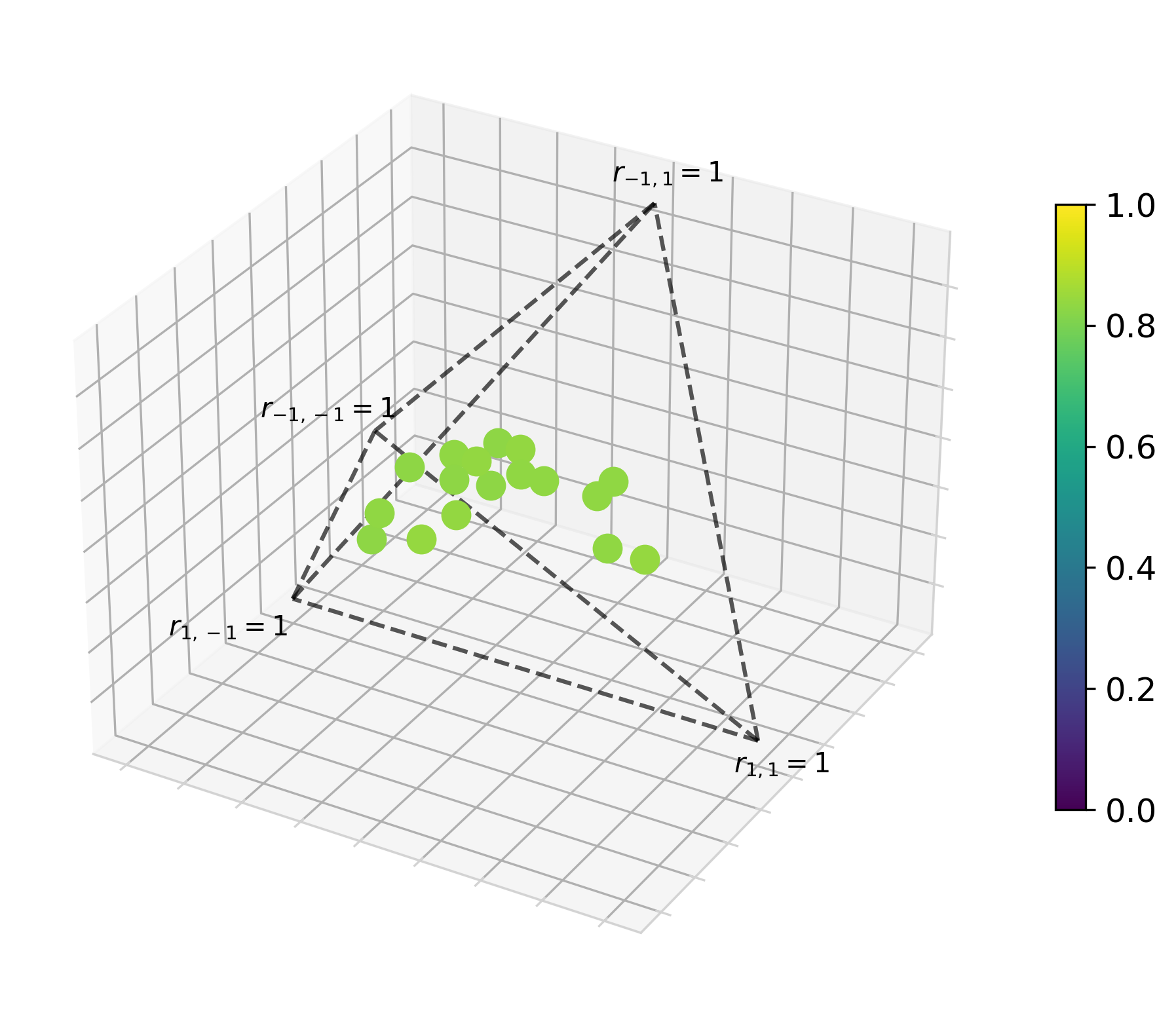}
        \caption{UrbanCars-syn}
    \end{subfigure}
    \hfill
    \begin{subfigure}{0.24\linewidth}
        \includegraphics[width=\linewidth]{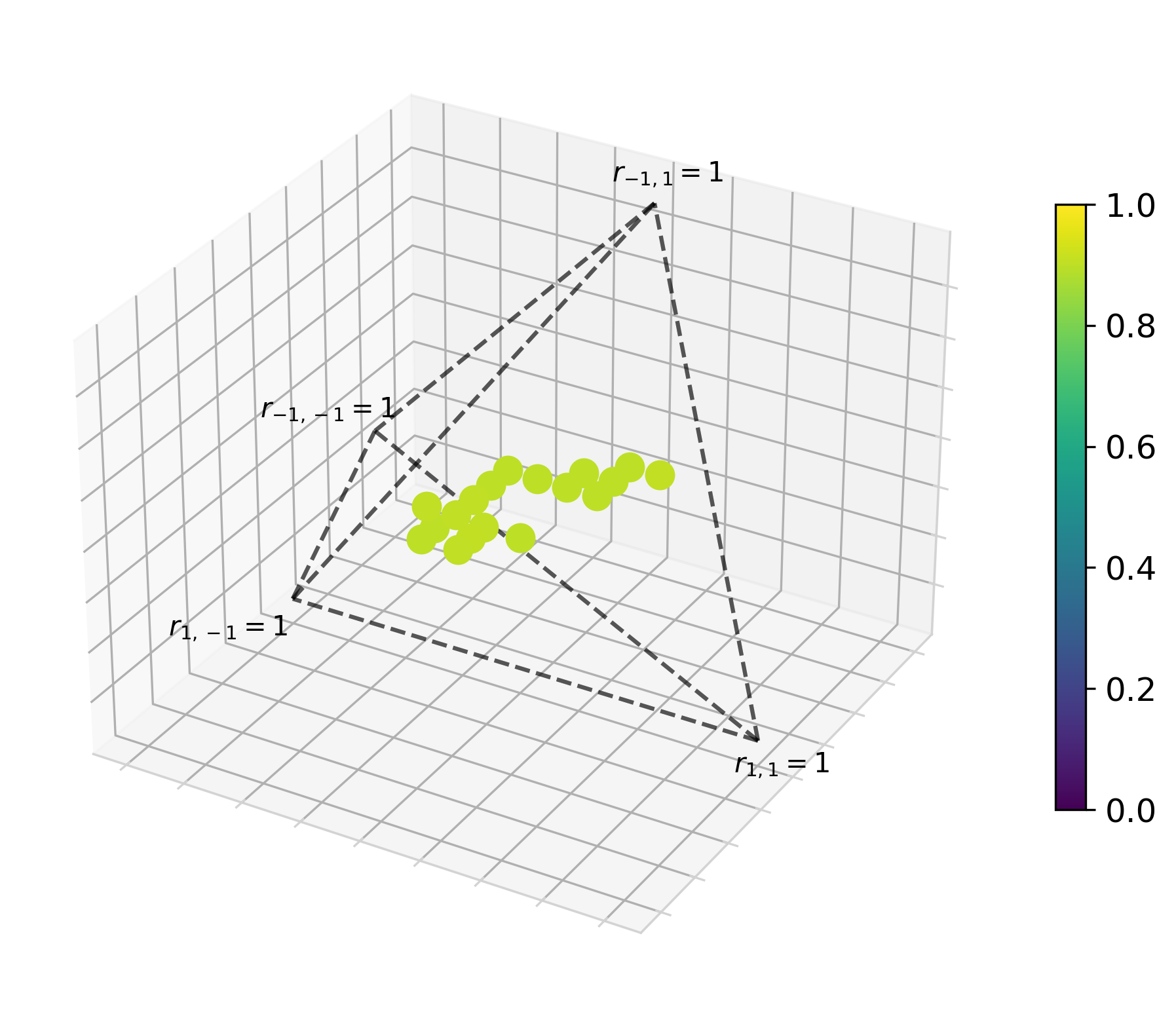}
        \caption{CelebA-subset}
    \end{subfigure}
    \hfill
    \begin{subfigure}{0.24\linewidth}
        \includegraphics[width=\linewidth]{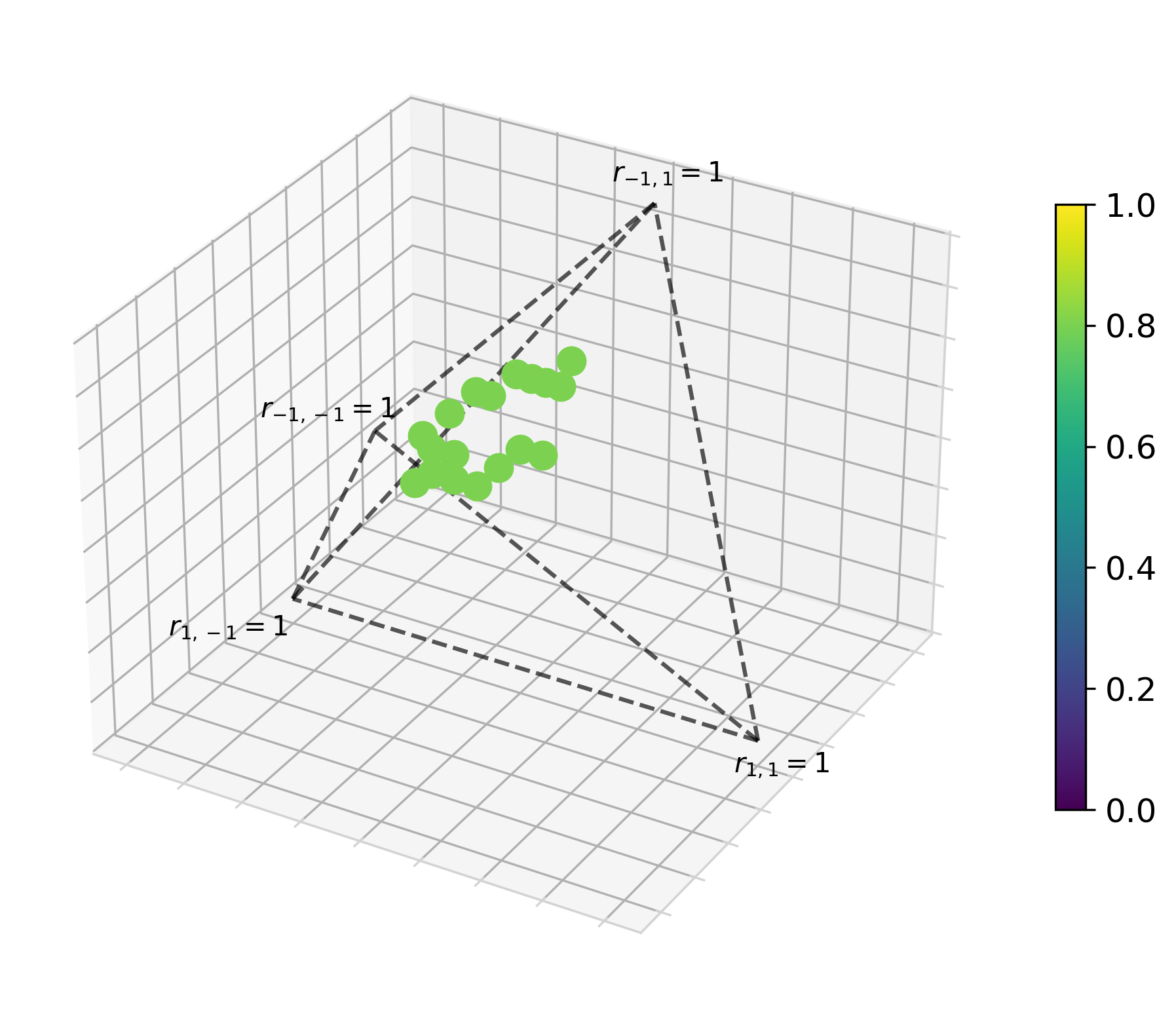}
        \caption{CivilComments-subset}
    \end{subfigure}
    \caption{Optimality of test performance (95th percentile) across multiple configurations.}
    \label{appd:fig:opt1}
\end{figure*}

\begin{figure*}[h]
    \centering
    \begin{subfigure}{0.24\linewidth}
        \includegraphics[width=\linewidth]{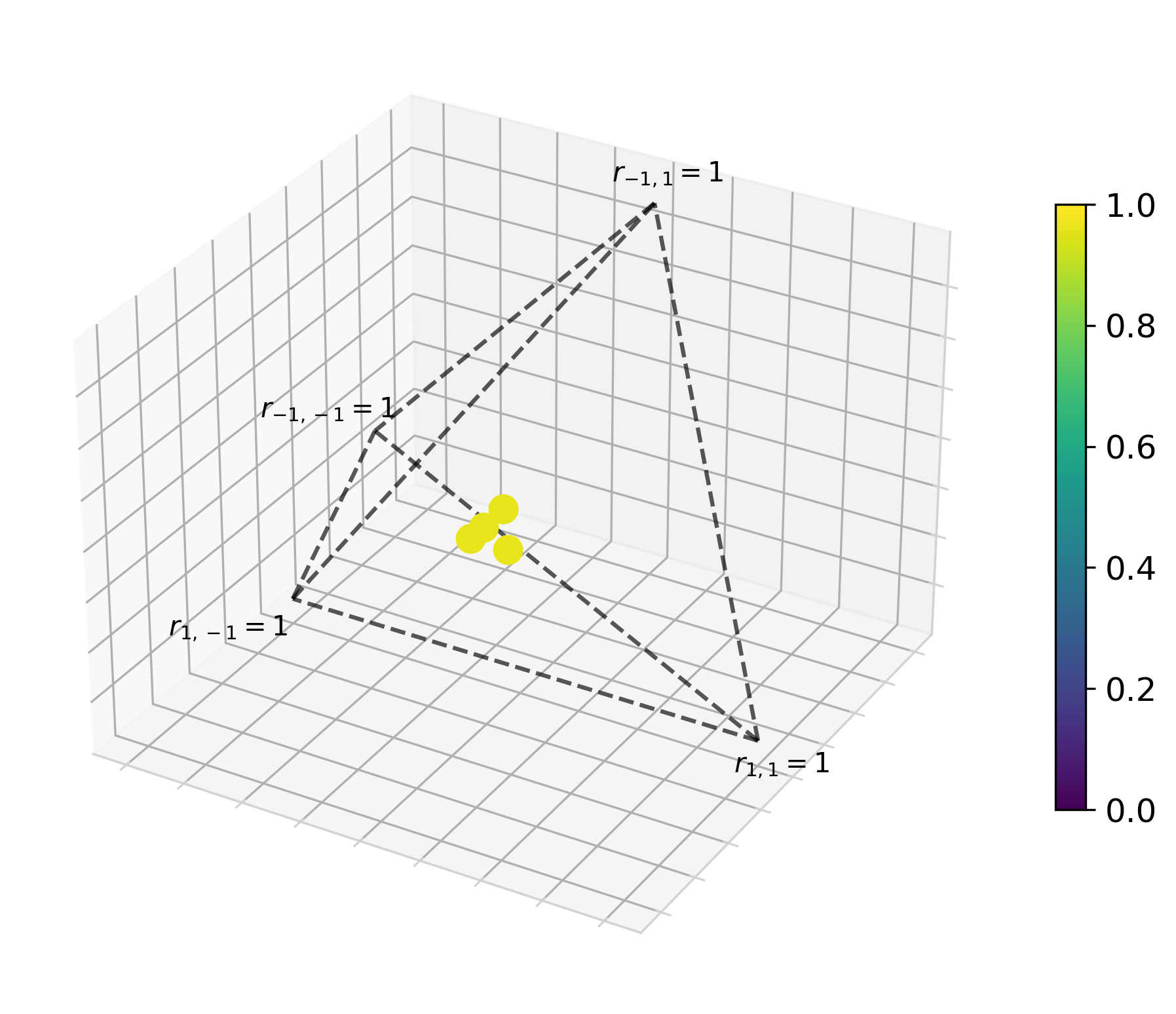}
        \caption{Waterbirds-syn}
    \end{subfigure}
    \hfill
    \begin{subfigure}{0.24\linewidth}
        \includegraphics[width=\linewidth]{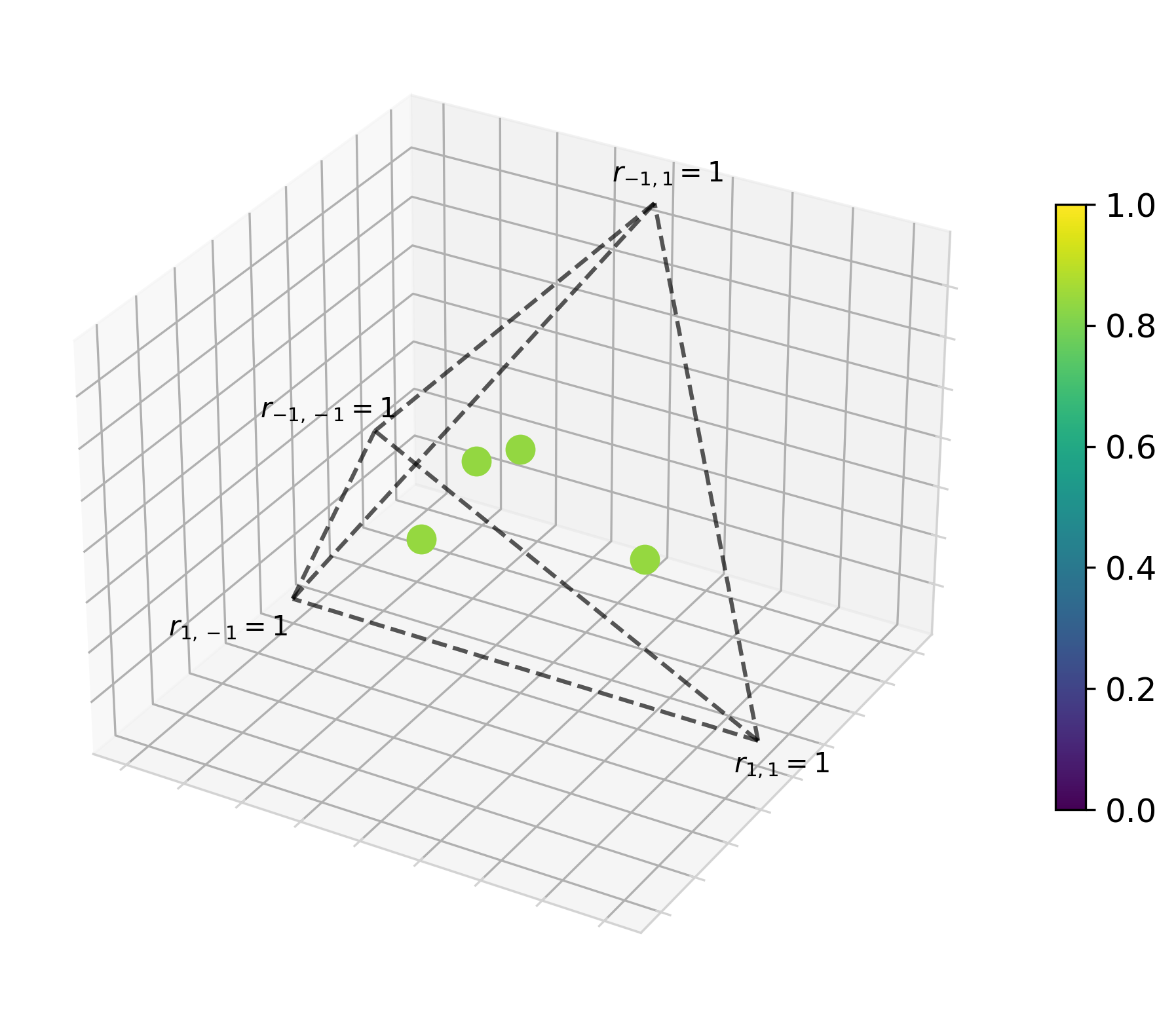}
        \caption{UrbanCars-syn}
    \end{subfigure}
    \hfill
    \begin{subfigure}{0.24\linewidth}
        \includegraphics[width=\linewidth]{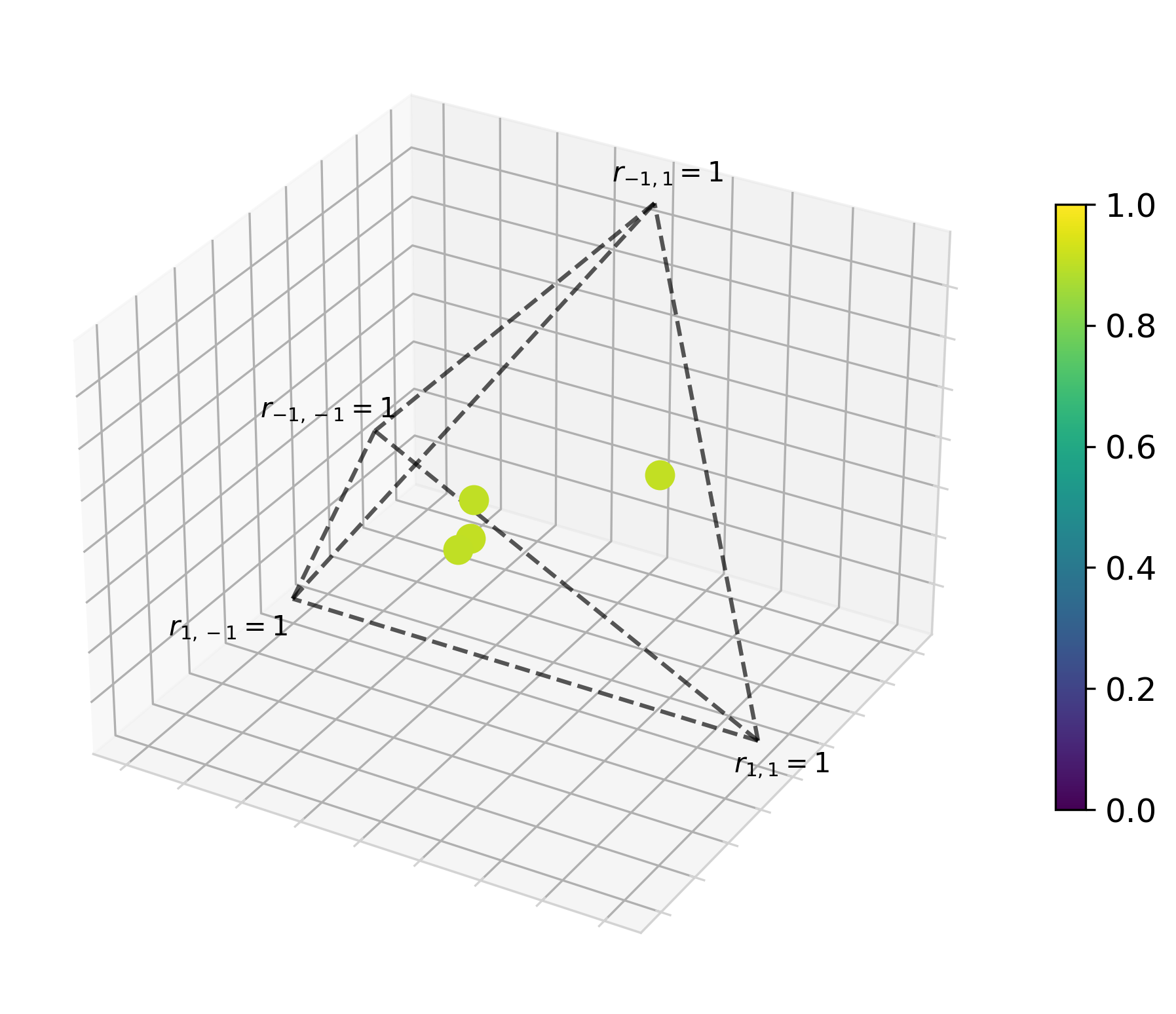}
        \caption{CelebA-subset}
    \end{subfigure}
    \hfill
    \begin{subfigure}{0.24\linewidth}
        \includegraphics[width=\linewidth]{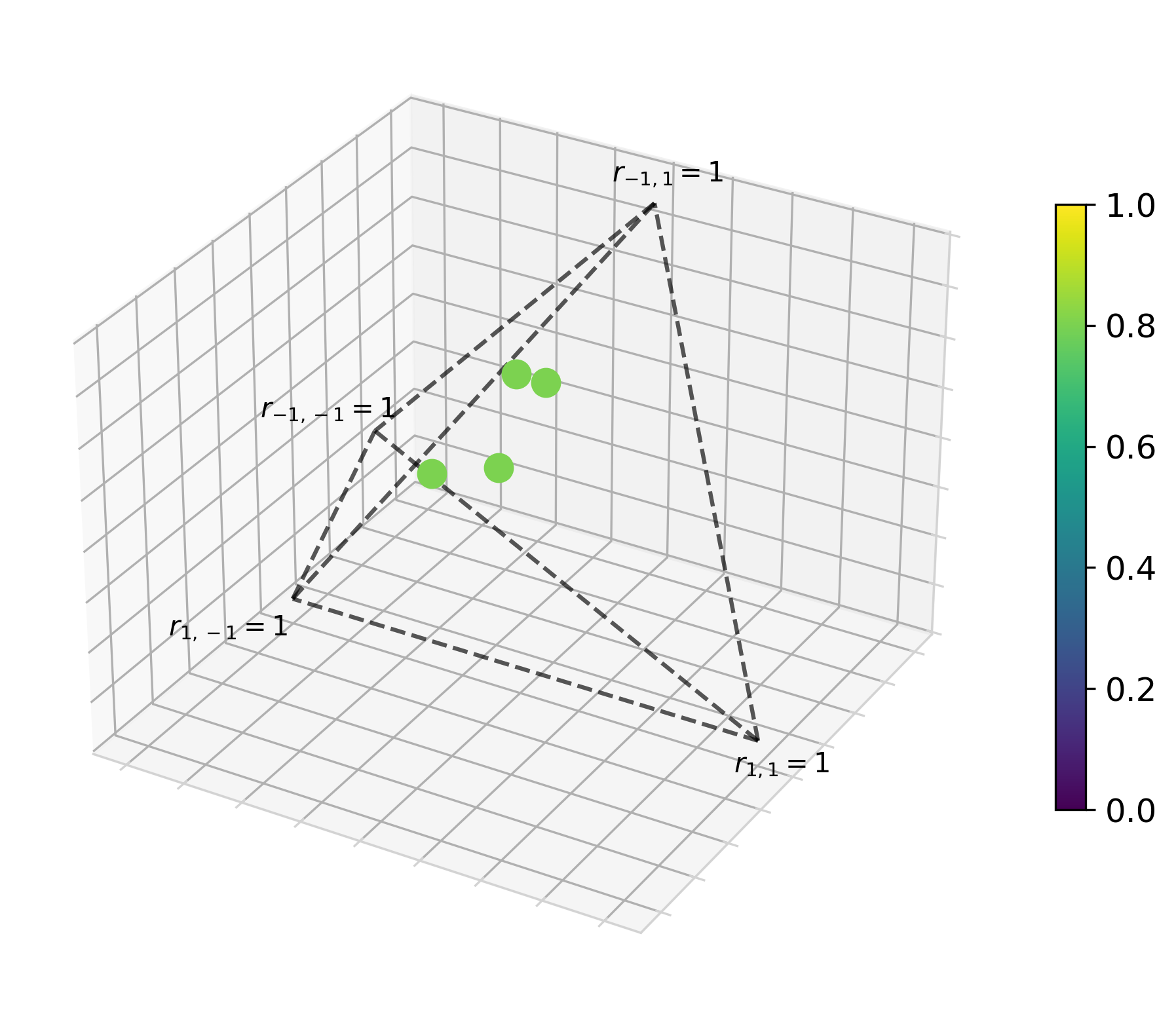}
        \caption{CivilComments-subset}
    \end{subfigure}
    \caption{Optimality of test performance (99th percentile) across multiple configurations.}
    \label{appd:fig:opt2}
\end{figure*}

    We display the results of visualizing configurations achieving 90th percentile or higher test accuracy in Section 4.4 to show the approximate region of those optimal configurations. This subsection includes the results of the 95th percentile and 99th percentile, for both theoretical results in Figure~\ref{appd:fig:opt0} and empirical results in Figures~\ref{appd:fig:opt1} and ~\ref{appd:fig:opt2}, to further show the existence of optimal configurations.

\section{Additional Results}
\subsection{Results with Worst-group Accuracy (WGA)}\label{appd:wga-results}
    In this work, subpopulation robustness refers to generalization under subpopulation shift from the training distribution to a different test distribution. For simplicity, we vary only the training distribution and evaluate performance on a balanced test distribution, making adjusted accuracy a natural robustness metric. We use adjusted accuracy in our theoretical analysis because it is positively correlated with WGA under subpopulation shifts~\cite{yang2023change} and, as a smooth average of group accuracies, enables the gradient-based analysis in Proposition~1. To address the potential tradeoff with WGA and the accuracy gap between best and worst groups, we additionally report WGA obtained from the experiments.

\begin{figure*}[h!]
    \centering

    \begin{subfigure}{0.24\linewidth}
        \includegraphics[width=\linewidth]{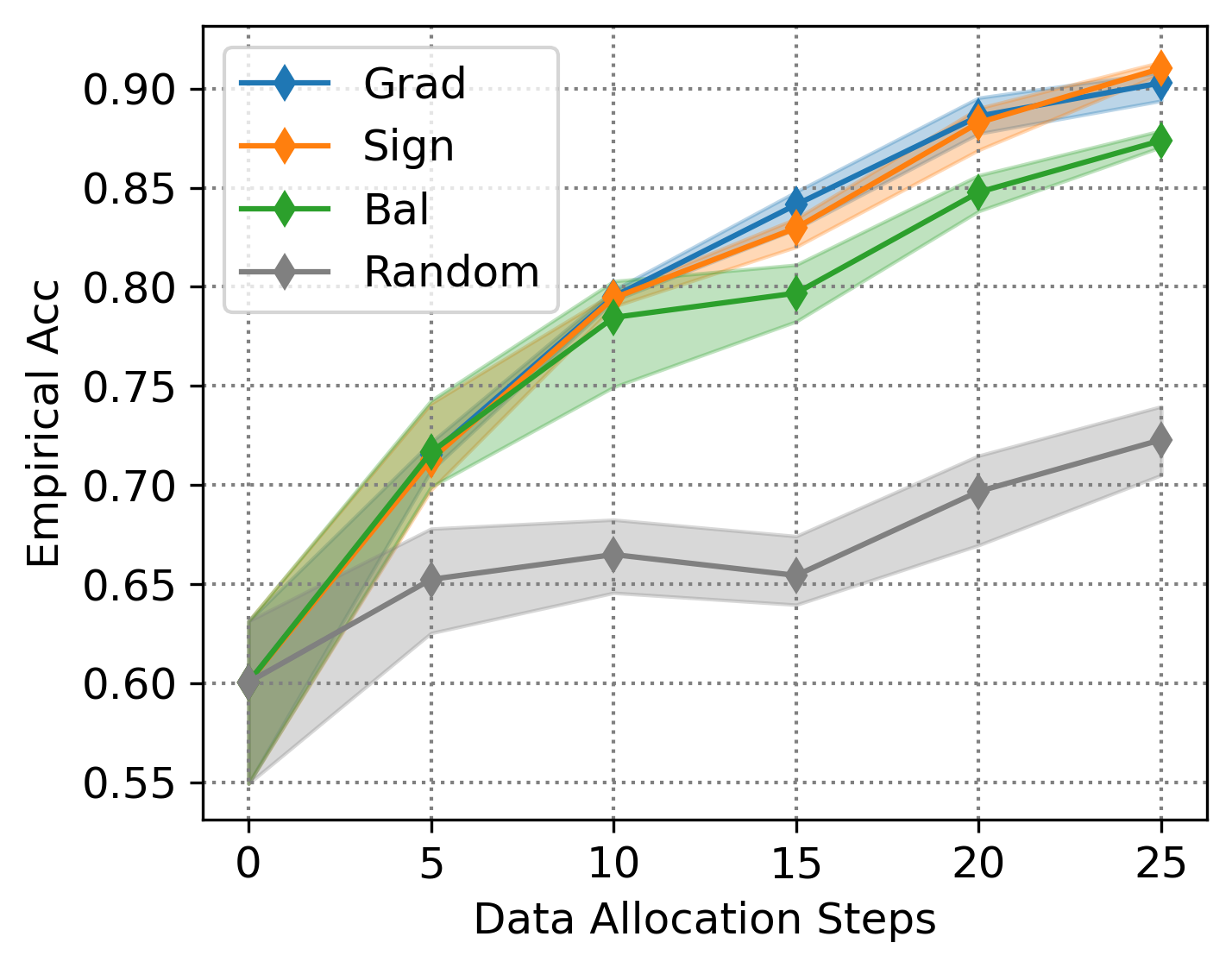}
    \end{subfigure}
    \hfill
    \begin{subfigure}{0.24\linewidth}
        \includegraphics[width=\linewidth]{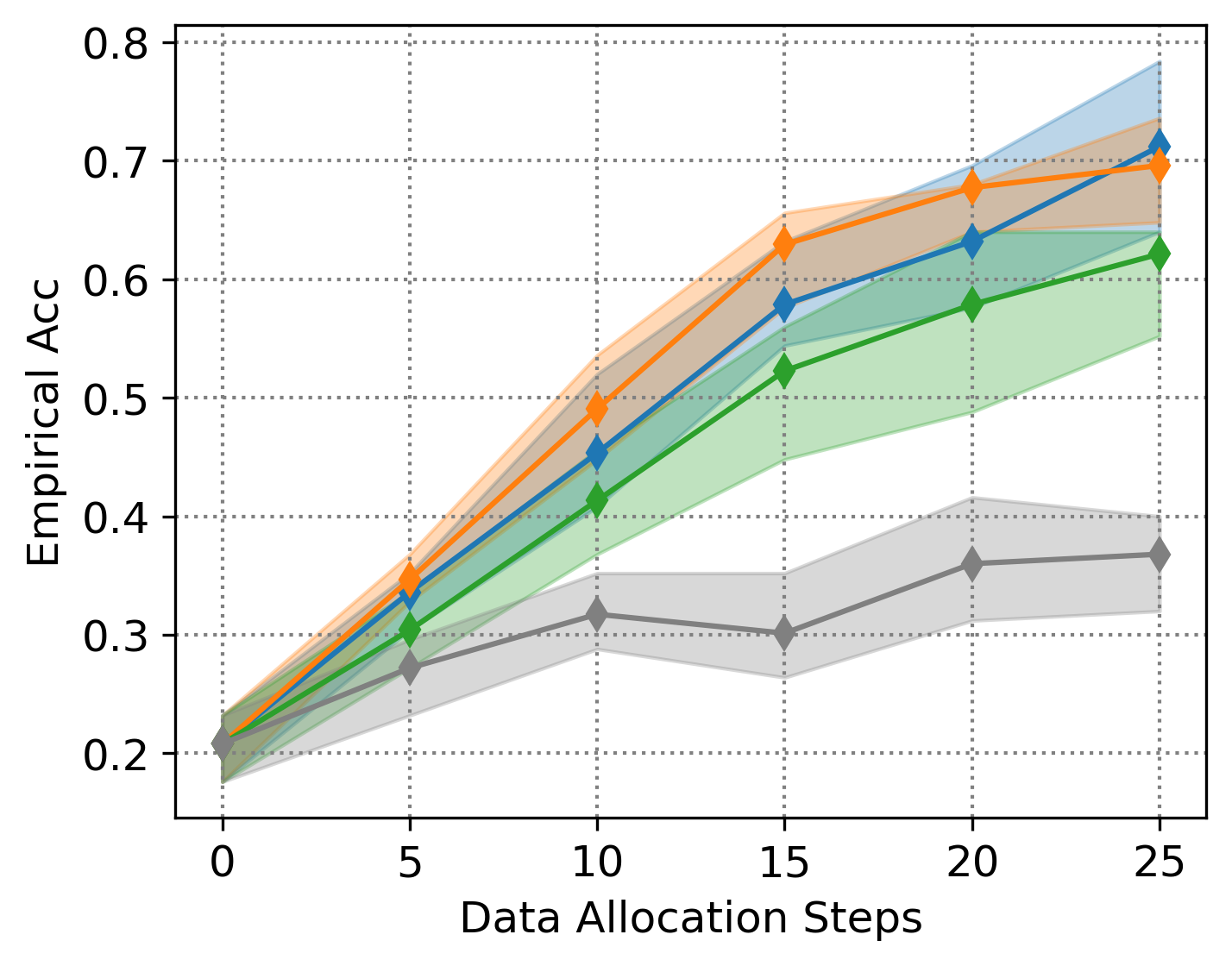}
    \end{subfigure}
    \hfill
    \begin{subfigure}{0.24\linewidth}
        \includegraphics[width=\linewidth]{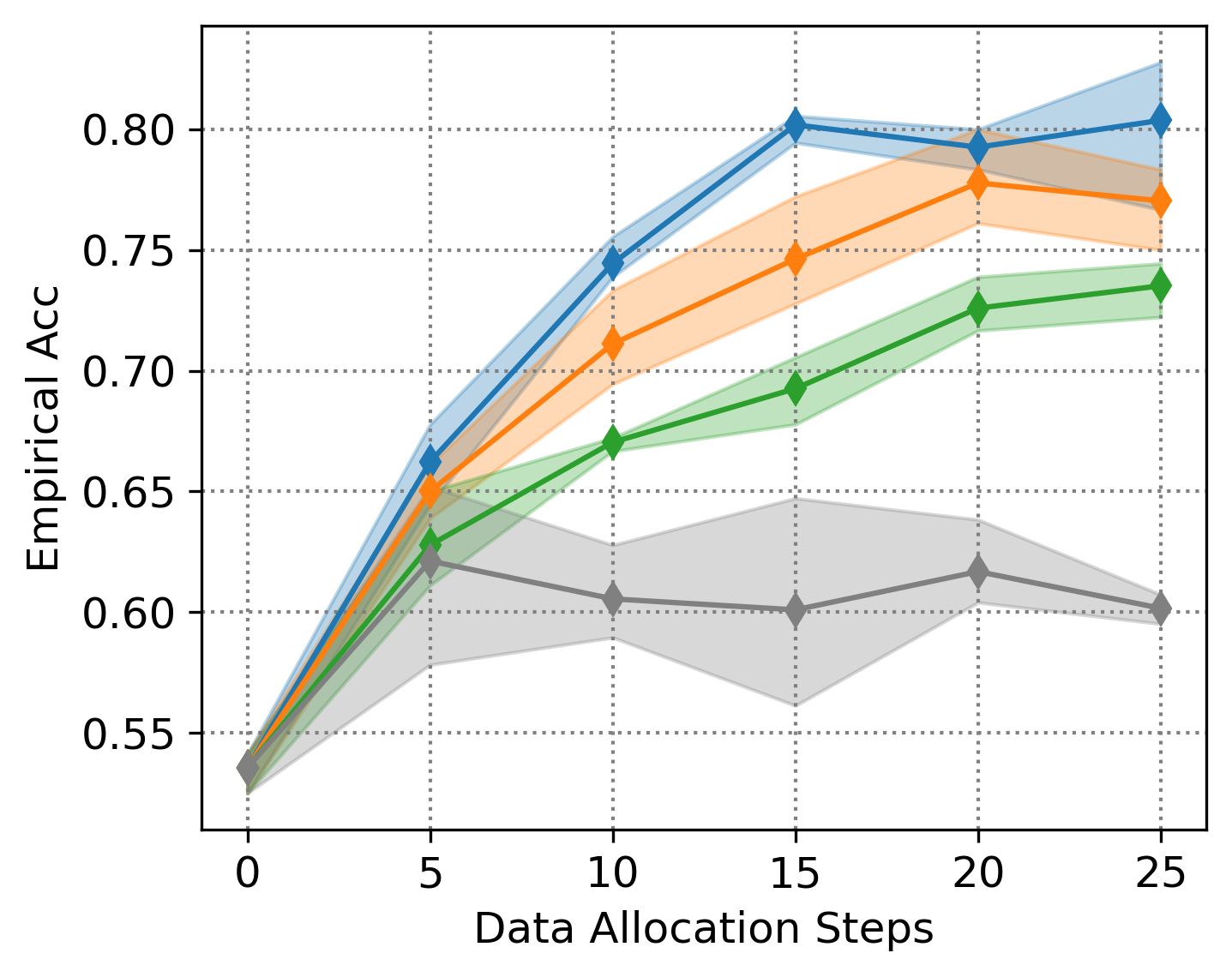}
    \end{subfigure}
    \hfill
    \begin{subfigure}{0.24\linewidth}
        \includegraphics[width=\linewidth]{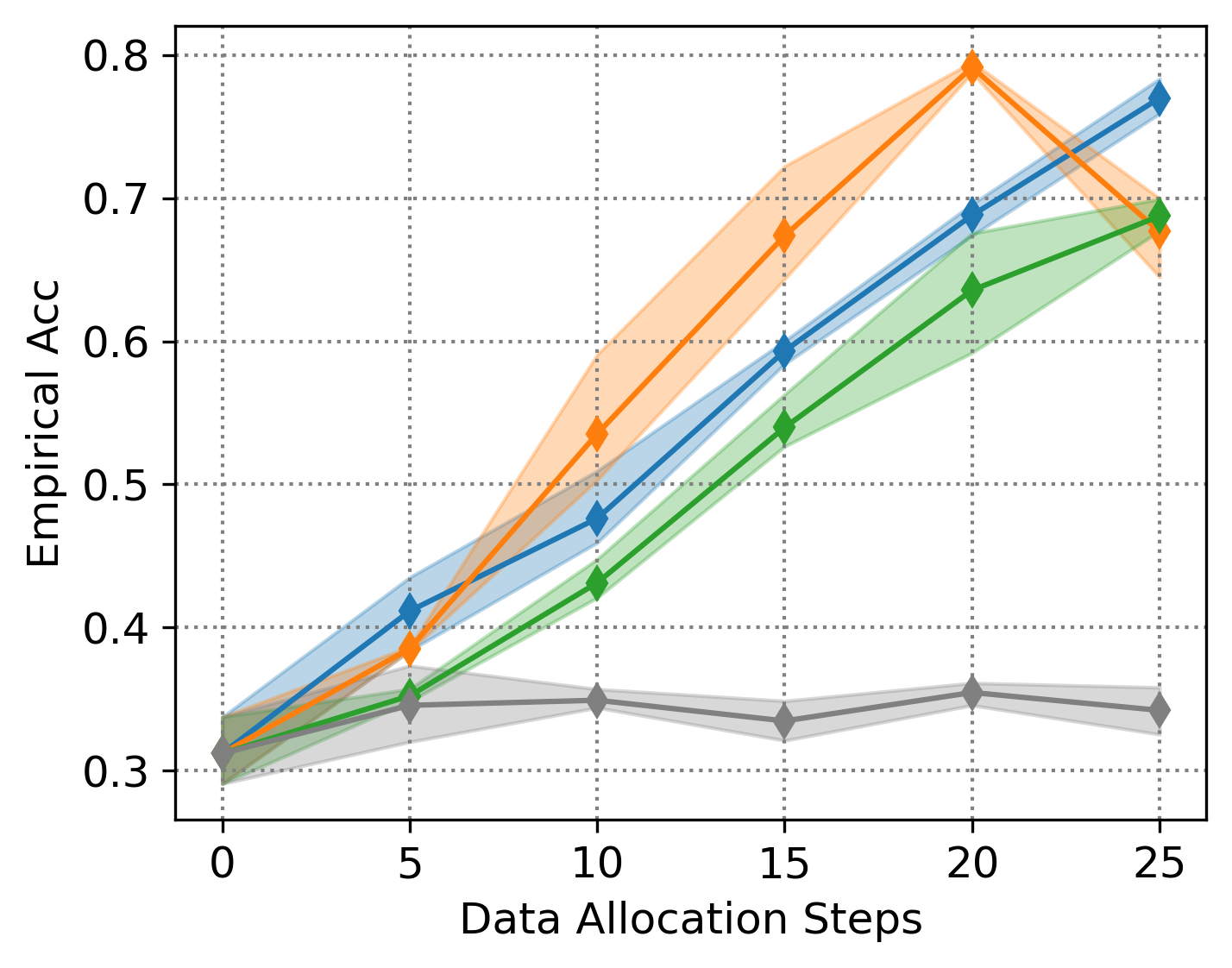}
    \end{subfigure}

    \vspace{1em}

    \begin{subfigure}{0.24\linewidth}
        \includegraphics[width=\linewidth]{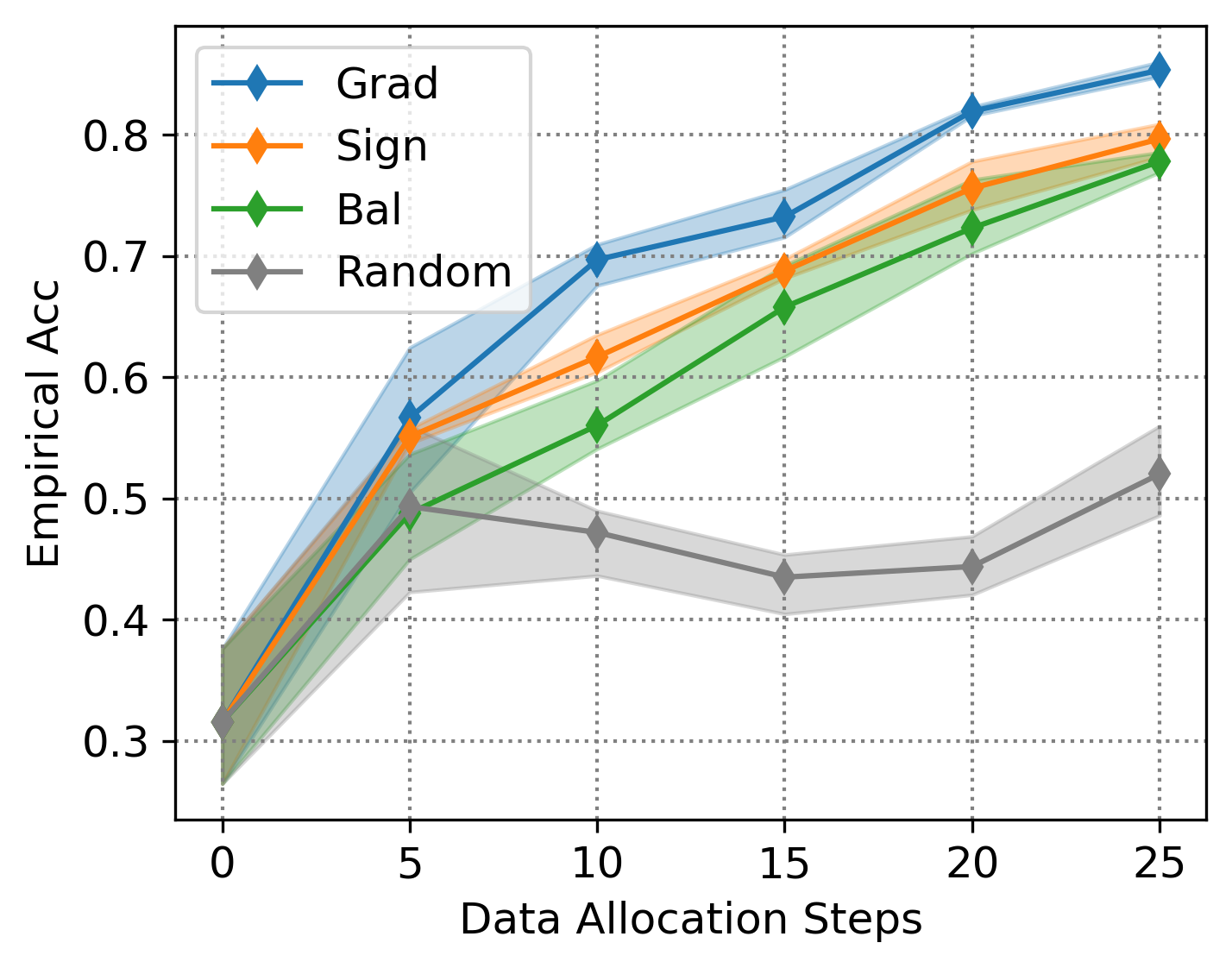}
    \end{subfigure}
    \hfill
    \begin{subfigure}{0.24\linewidth}
        \includegraphics[width=\linewidth]{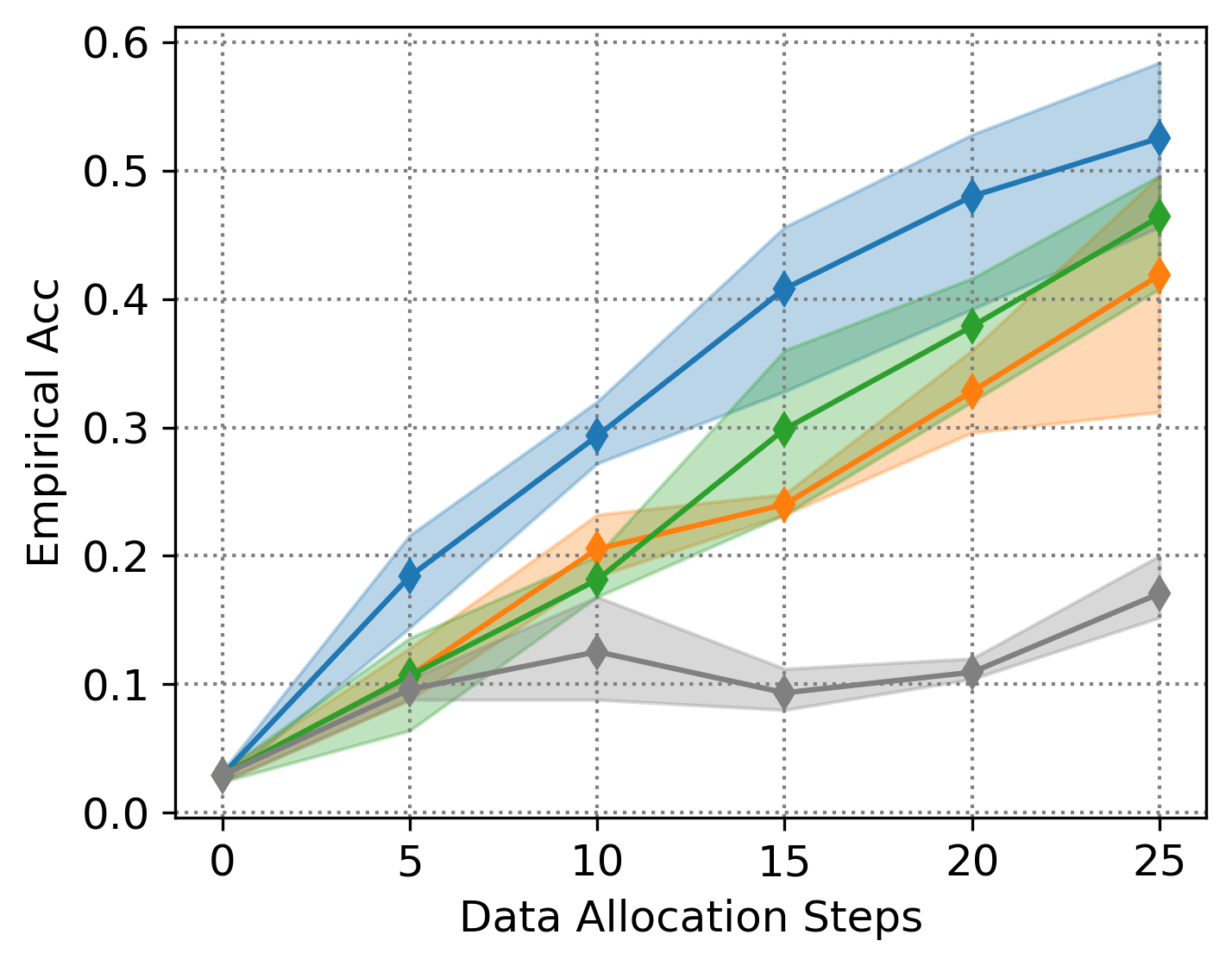}
    \end{subfigure}
    \hfill
    \begin{subfigure}{0.24\linewidth}
        \includegraphics[width=\linewidth]{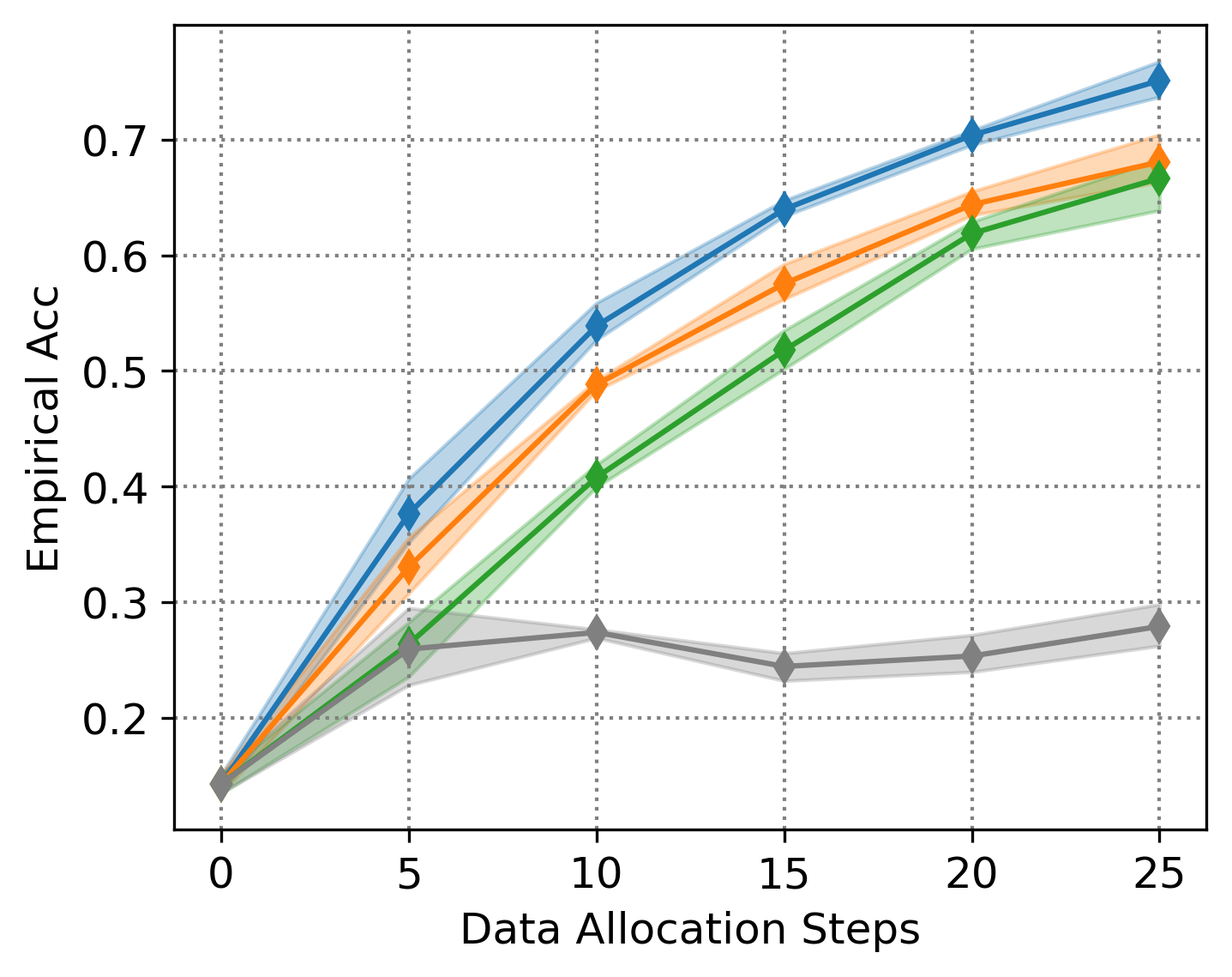}
    \end{subfigure}
    \hfill
    \begin{subfigure}{0.24\linewidth}
        \includegraphics[width=\linewidth]{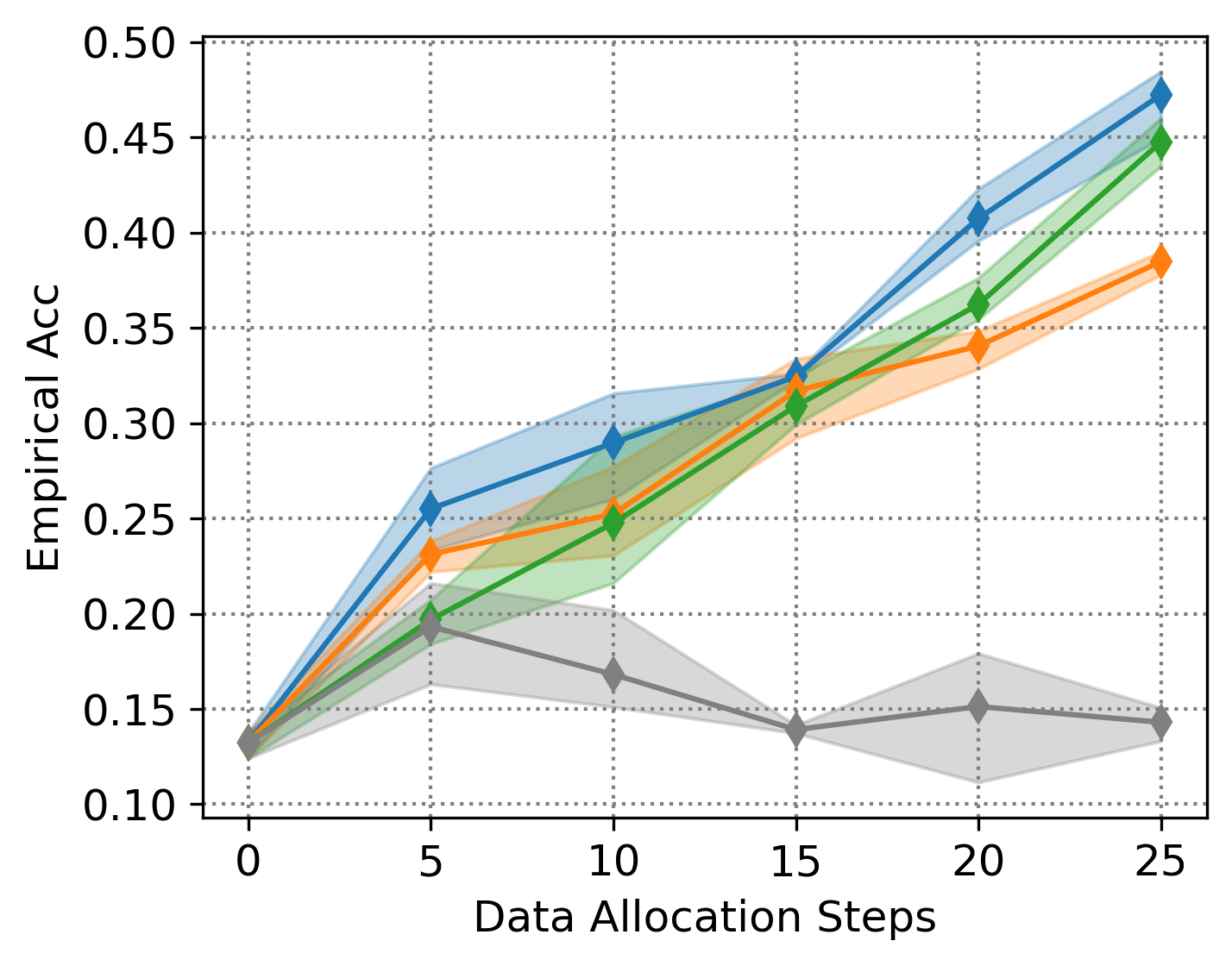}
    \end{subfigure}

    \vspace{1em}

    \begin{subfigure}{0.24\linewidth}
        \includegraphics[width=\linewidth]{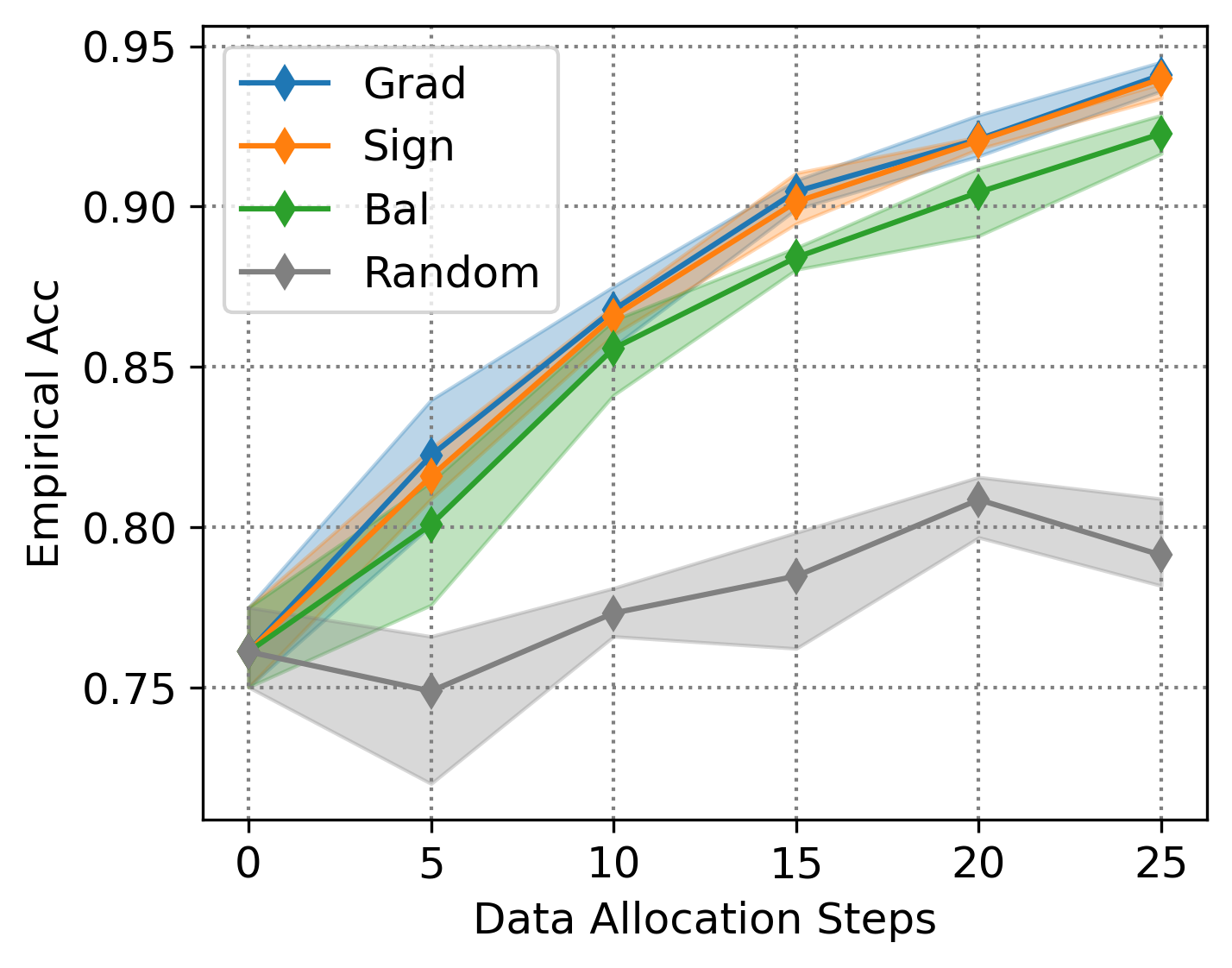}
        \caption{Waterbirds-syn}
    \end{subfigure}
    \hfill
    \begin{subfigure}{0.24\linewidth}
        \includegraphics[width=\linewidth]{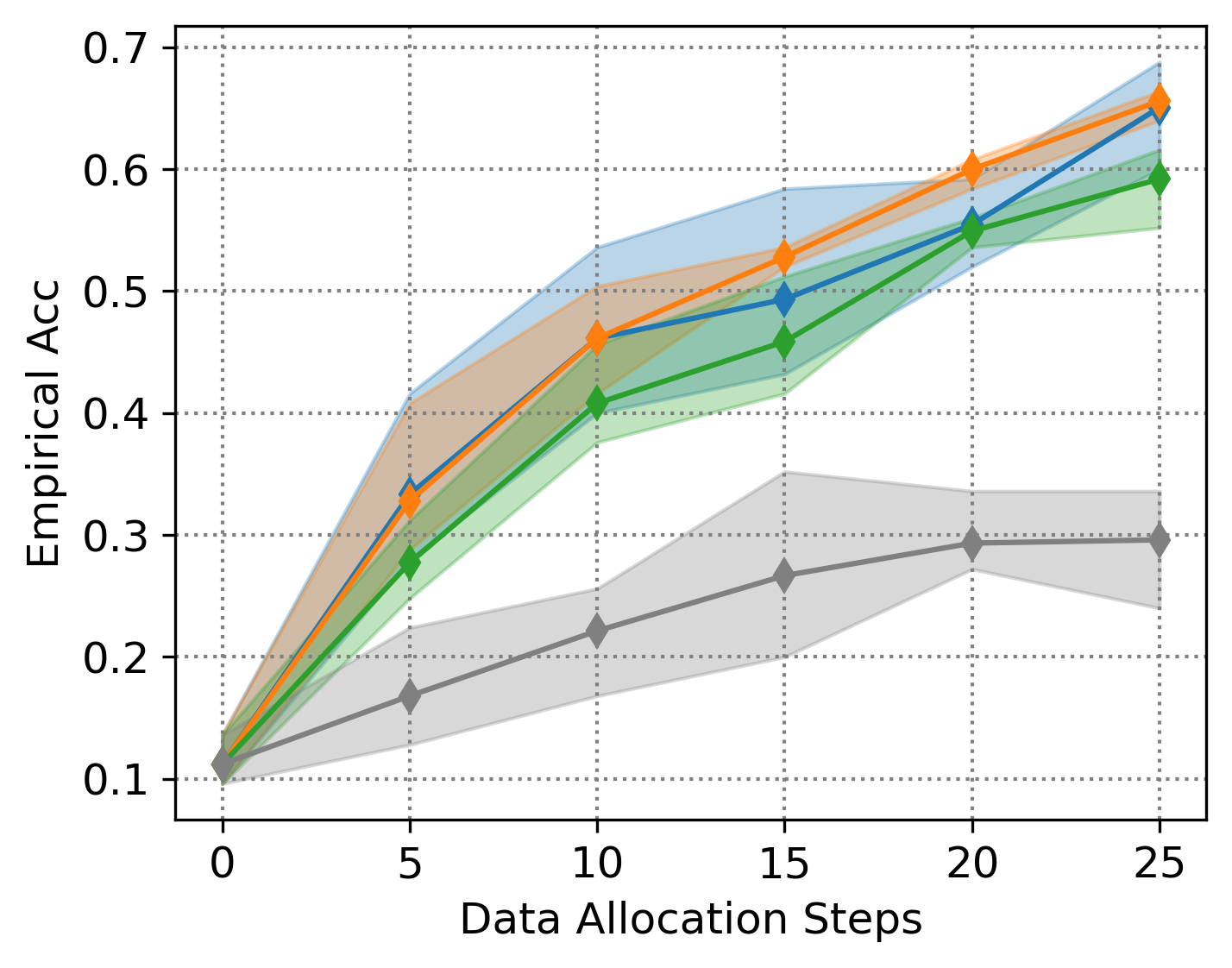}
        \caption{UrbanCars-syn}
    \end{subfigure}
    \hfill
    \begin{subfigure}{0.24\linewidth}
        \includegraphics[width=\linewidth]{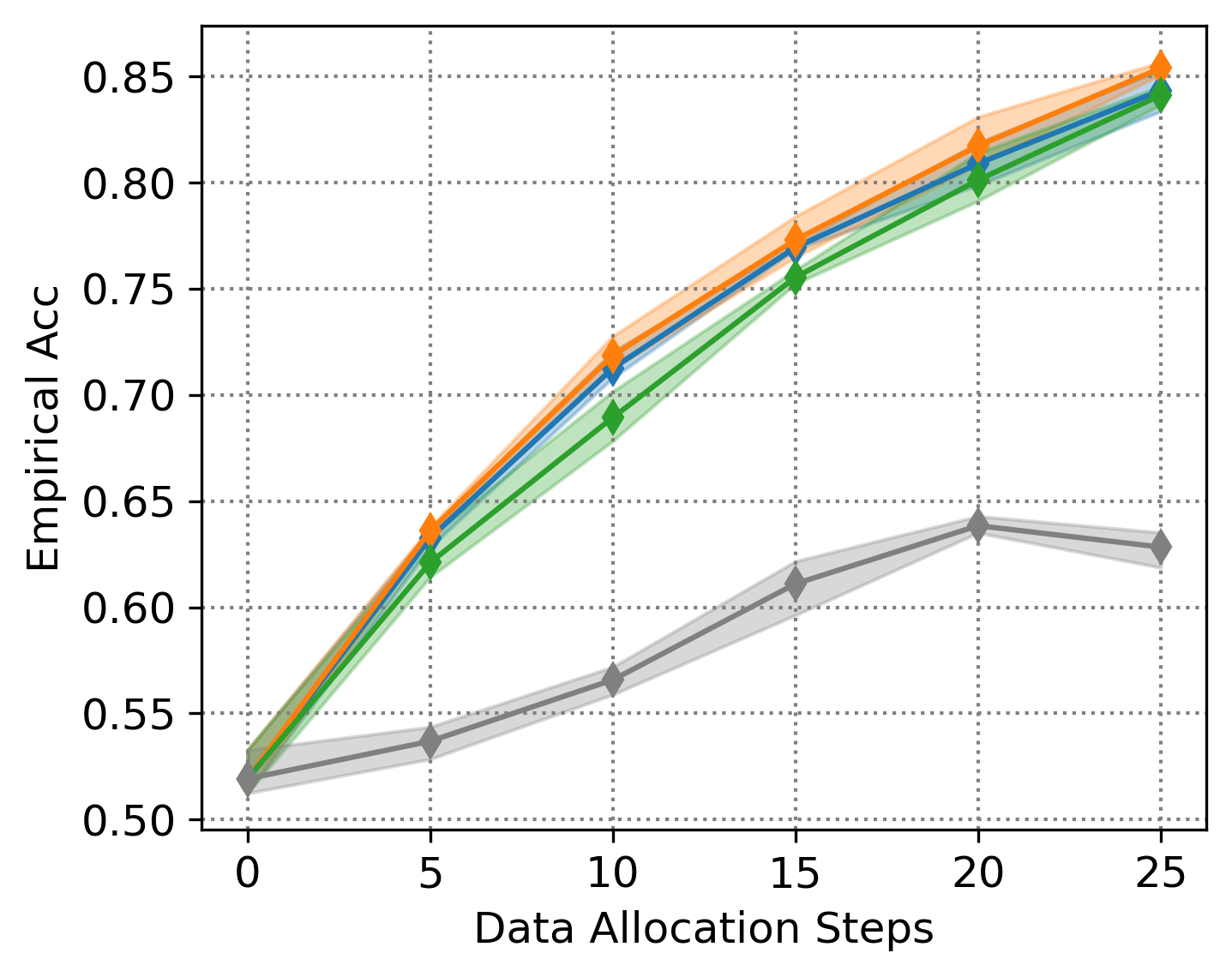}
        \caption{CelebA-subset}
    \end{subfigure}
    \hfill
    \begin{subfigure}{0.24\linewidth}
        \includegraphics[width=\linewidth]{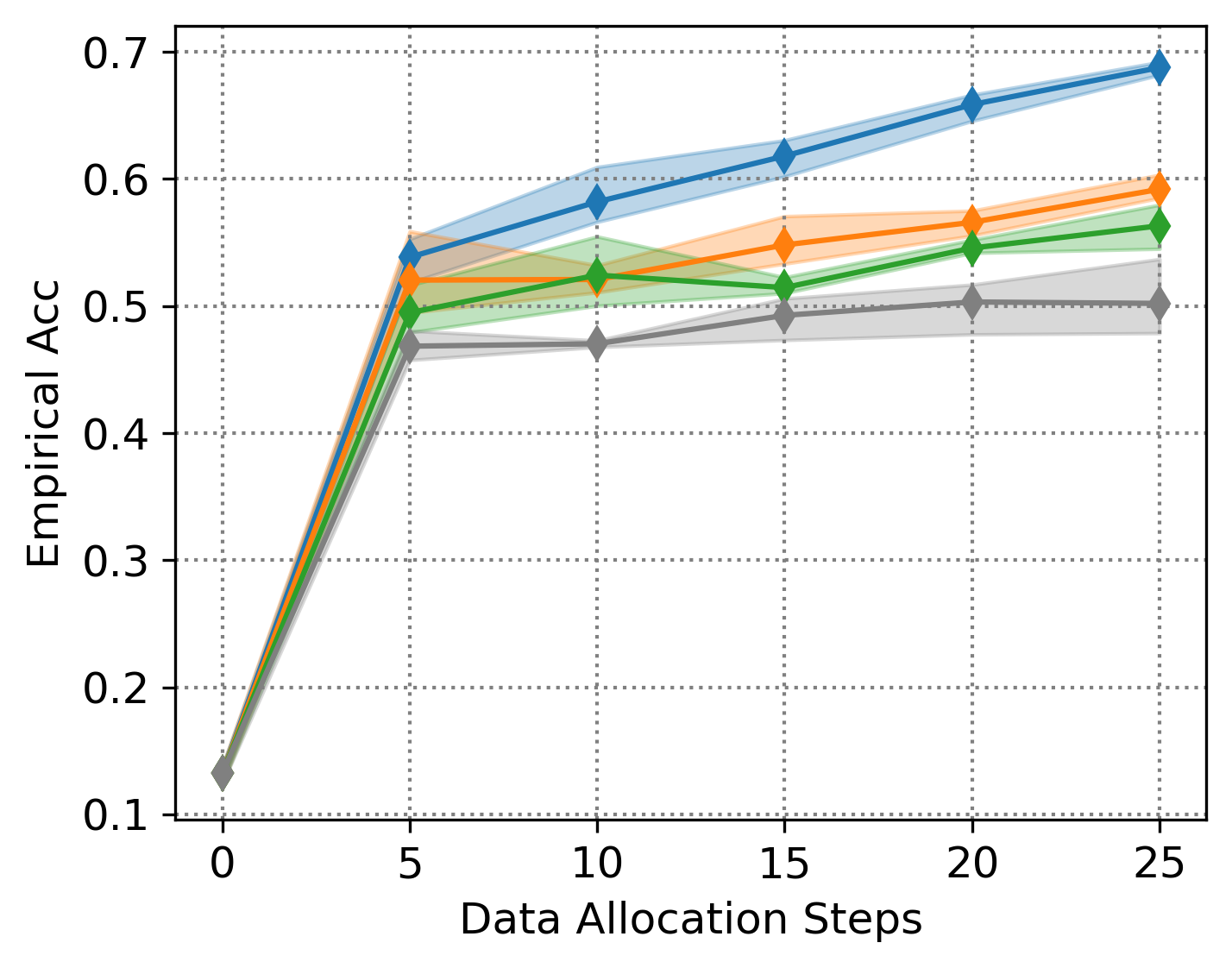}
        \caption{CivilComments-subset}
    \end{subfigure}

    \caption{Efficiency of WGA of data allocation along directions across datasets and initial configurations.}
    \vspace{-3mm}
\end{figure*}

\begin{figure*}[h]
    \centering
    \begin{subfigure}{0.24\linewidth}
        \includegraphics[width=\linewidth]{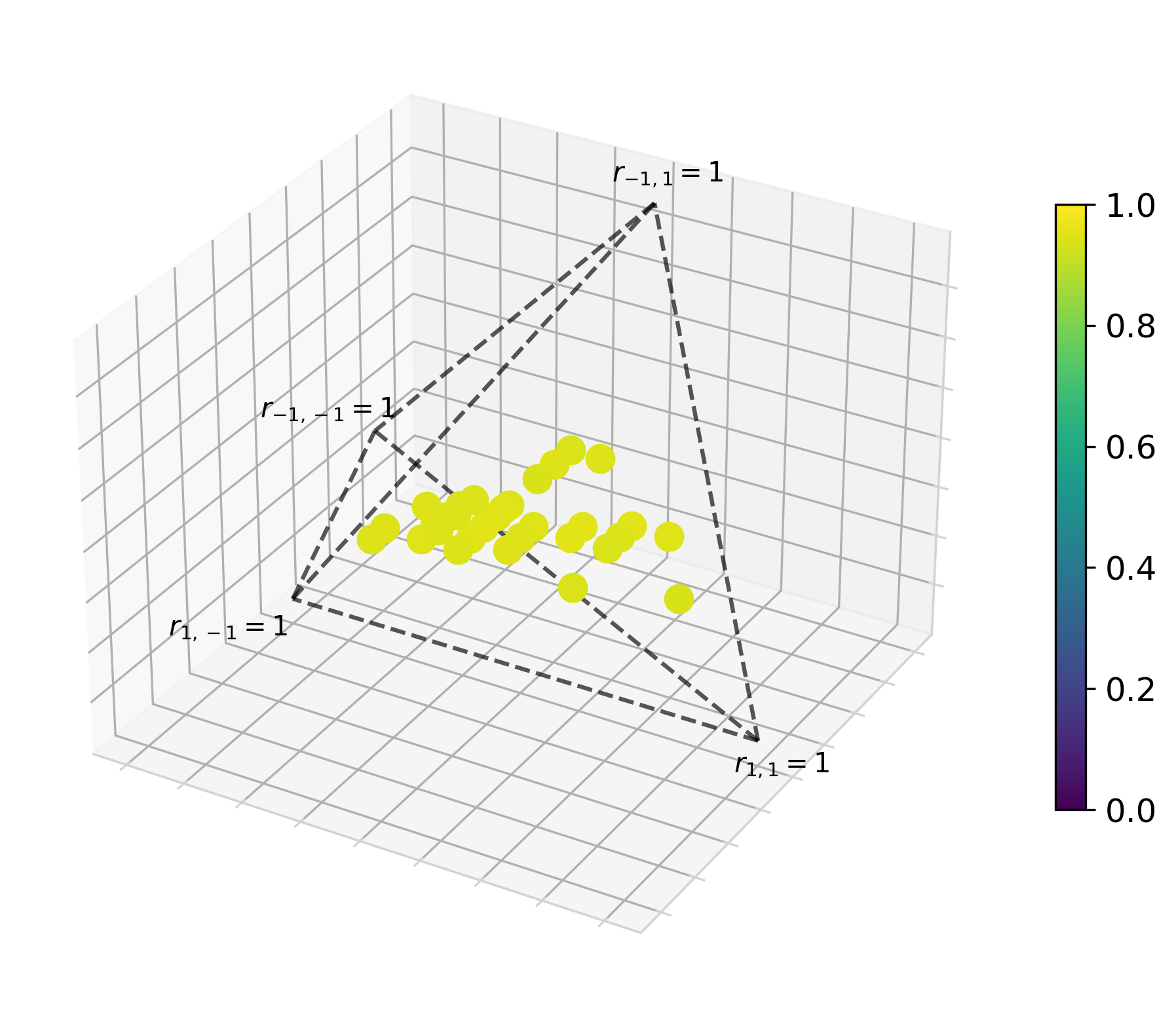}
        \caption{Waterbirds-syn}
    \end{subfigure}
    \hfill
    \begin{subfigure}{0.24\linewidth}
        \includegraphics[width=\linewidth]{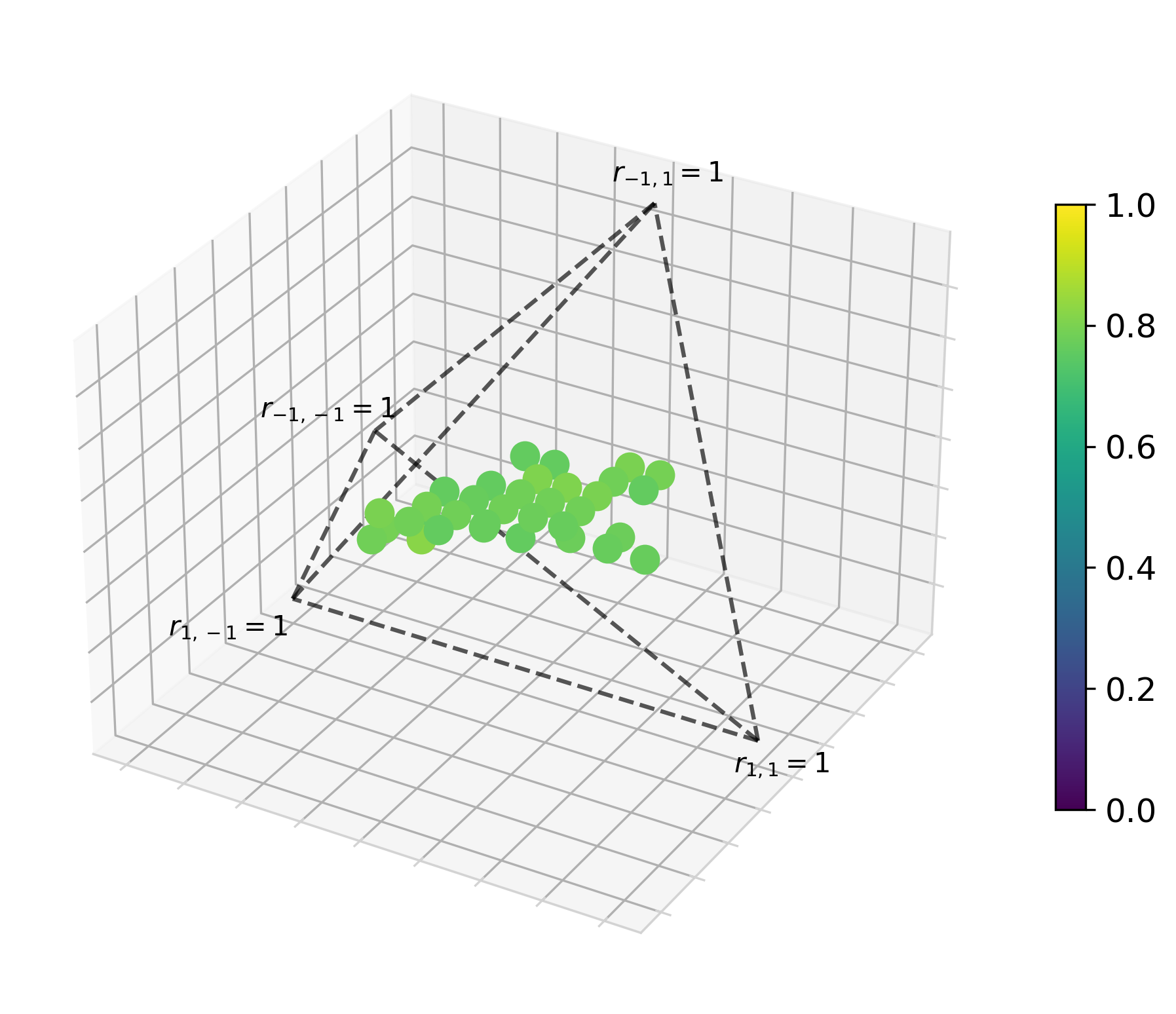}
        \caption{UrbanCars-syn}
    \end{subfigure}
    \hfill
    \begin{subfigure}{0.24\linewidth}
        \includegraphics[width=\linewidth]{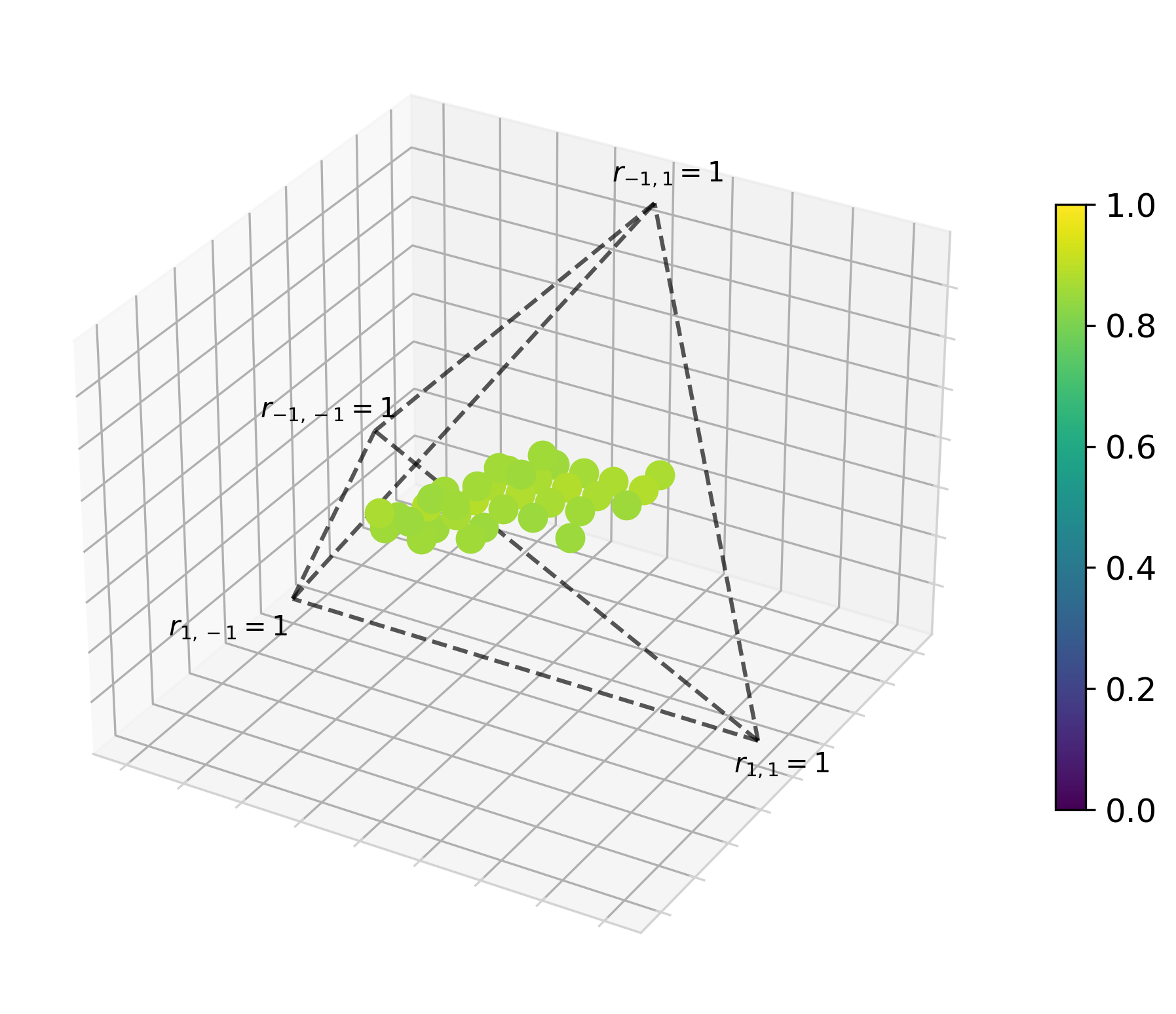}
        \caption{CelebA-subset}
    \end{subfigure}
    \hfill
    \begin{subfigure}{0.24\linewidth}
        \includegraphics[width=\linewidth]{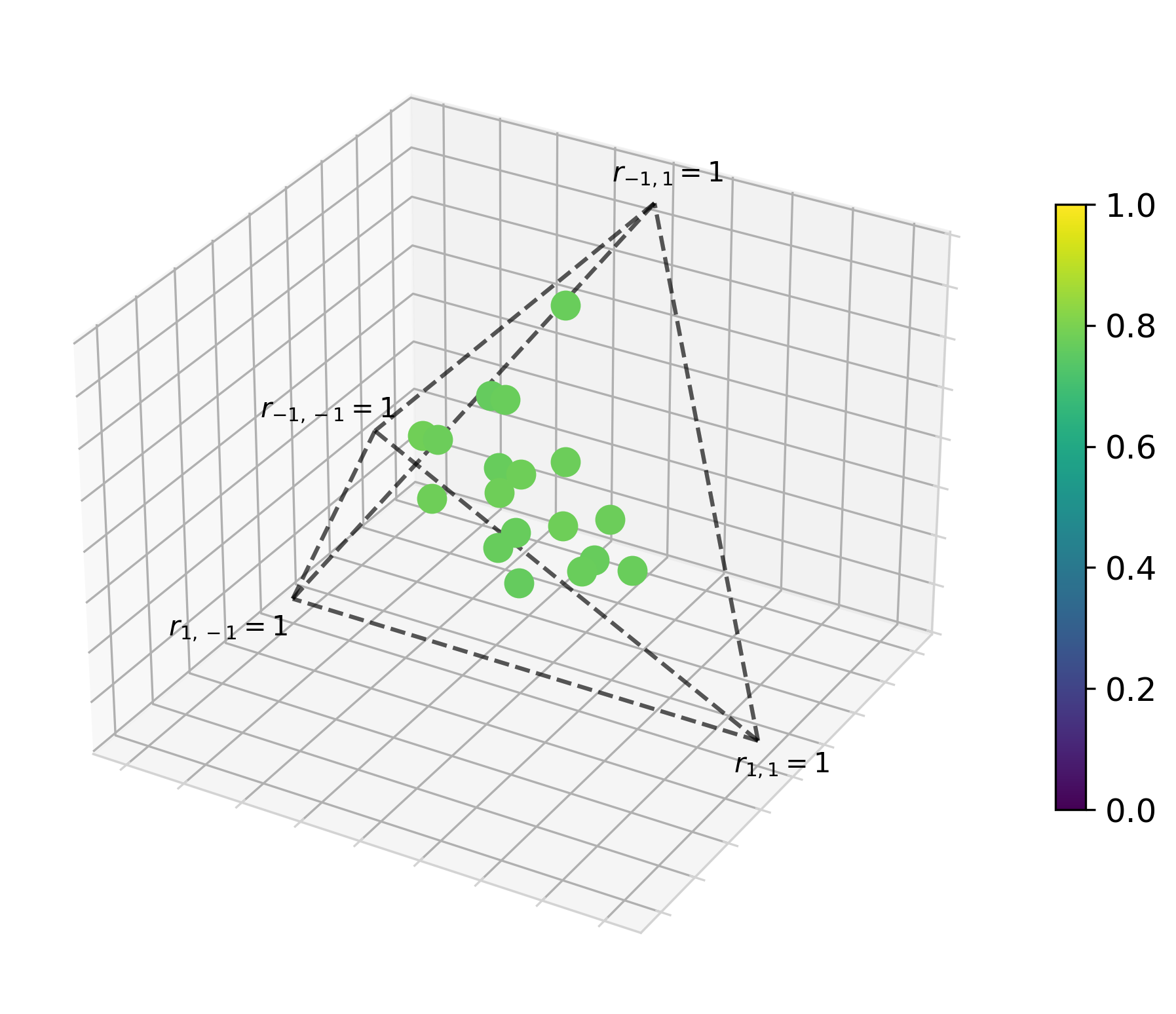}
        \caption{CivilComments-subset}
    \end{subfigure}
    \caption{Optimality of test WGA across multiple configurations.}
    \vspace{-2mm}
\end{figure*}

    Empirically, the directions derived from adjusted accuracy also improve WGA.
    Table~\ref{tab:wga-efficiency} gives the step-15 comparison underlying this conclusion: Grad exceeds Bal in every evaluated dataset--configuration pair. Table~\ref{tab:wga-auc} shows that the corresponding WGA-AUC comparison has the same ordering in all 12 pairs.
    \begin{table}[t]
    \centering
    \caption{WGA (\%) after 15 allocation steps. Each cell reports Grad / Bal (mean over three seeds).}
    \label{tab:wga-efficiency}
    \small
    \setlength{\tabcolsep}{4pt}
    \begin{tabular}{lcccc}
        \toprule
        & Waterbirds-syn & UrbanCars-syn & CelebA-subset & CivilComments-subset \\
        \midrule
        Example 1 & 84.15 / 79.68 & 57.87 / 50.93 & 80.19 / 69.26 & 59.70 / 54.46 \\
        Example 2 & 73.24 / 65.79 & 40.80 / 29.87 & 63.98 / 51.79 & 32.30 / 30.12 \\
        Example 3 & 90.47 / 88.41 & 49.33 / 45.87 & 76.94 / 75.54 & 61.61 / 51.42 \\
        \bottomrule
    \end{tabular}
\end{table}

\begin{table}[t]
    \centering
    \caption{WGA-AUC over the allocation trajectory. Each cell reports Grad / Bal (mean over three seeds).}
    \label{tab:wga-auc}
    \small
    \setlength{\tabcolsep}{4pt}
    \begin{tabular}{lcccc}
        \toprule
        & Waterbirds-syn & UrbanCars-syn & CelebA-subset & CivilComments-subset \\
        \midrule
        Example 1 & 19.95 / 19.41 & 12.30 / 10.95 & 18.35 / 16.76 & 13.63 / 12.32 \\
        Example 2 & 17.00 / 14.88 & 8.21 / 6.06 & 13.53 / 11.07 & 7.82 / 6.95 \\
        Example 3 & 21.84 / 21.44 & 11.12 / 10.21 & 18.02 / 17.74 & 14.86 / 12.96 \\
        \bottomrule
    \end{tabular}
\end{table}

\begin{figure}[t!]
    \centering
    \begin{minipage}[t]{\linewidth}
        \centering
        \begin{subfigure}{0.49\linewidth}
            \includegraphics[width=\linewidth]{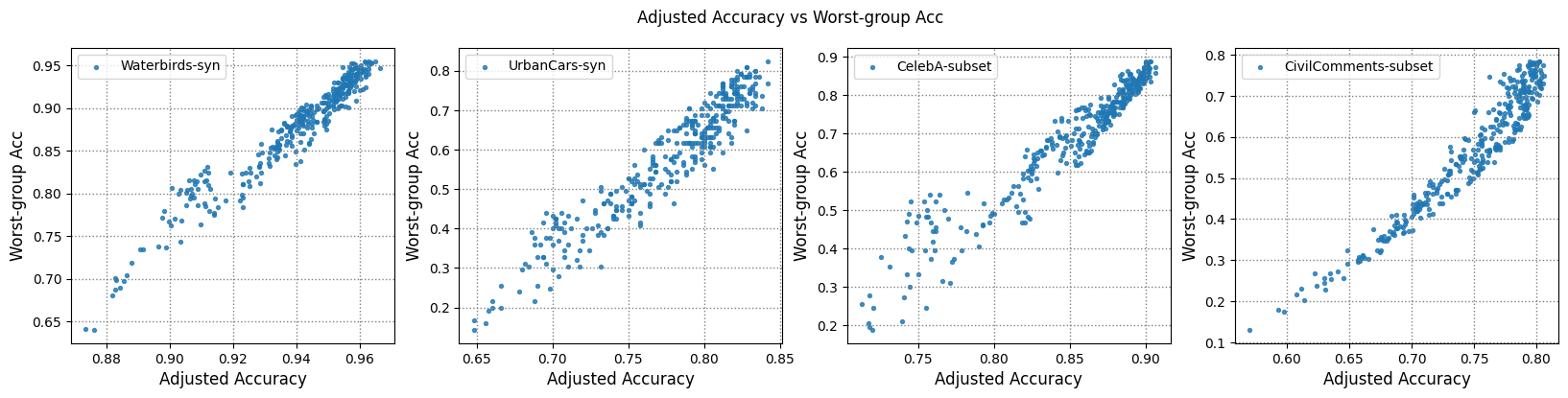}
            \caption{Adjusted Accuracy vs. WGA}
        \end{subfigure}
        \hfill
        \begin{subfigure}{0.49\linewidth}
            \includegraphics[width=\linewidth]{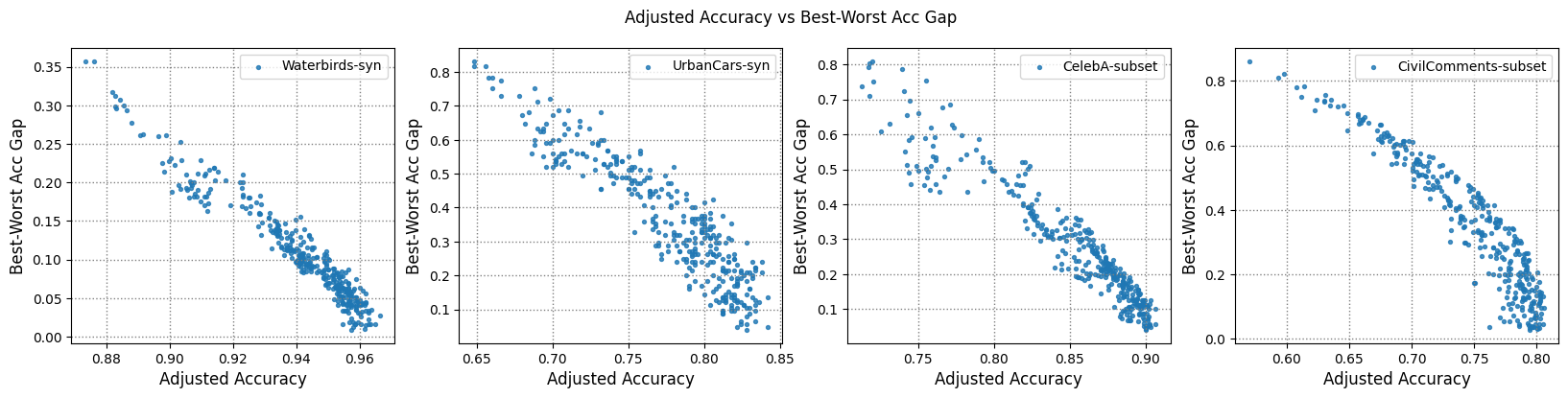}                        \caption{Adjusted Accuracy vs. Accuracy Gap}
        \end{subfigure}
        \caption{Relationship between Adjusted Accuracy, WGA, and Accuracy Gap.}
        \label{appd:fig:tradeoff}
    \end{minipage}
\end{figure}

    We further show the scatter between the adjusted accuracy and WGA or accuracy gap in Figure~\ref{appd:fig:tradeoff}, which shows a positive/negative correlation between adjusted accuracy and WGA or accuracy gap, respectively. This shows that maximizing adjusted accuracy can improve WGA or reduce the accuracy gap without an explicit tradeoff.

\subsection{Worst-group Accuracy Direction}\label{appd:wga-direction}
    Because Proposition~1 derives the accuracy gradient for each subgroup, the same framework can estimate a direction that directly improves WGA. Let
    \begin{equation}
        \mathrm{WGA}(\vp)=\min_{\va\in\mathcal{A}}Acc_{\va}(\vp),
        \qquad
        \mathcal{I}(\vp)=\arg\min_{\va\in\mathcal{A}}Acc_{\va}(\vp)
    \end{equation}
    denote WGA and the set of currently worst-performing groups. If the worst group is unique, $\mathcal{I}(\vp)=\{\va^*\}$, and remains unchanged in a neighborhood of $\vp$, then WGA is locally smooth and
    \begin{equation}
        \frac{\partial\mathrm{WGA}}{\partial p}
        =
        \frac{\partial Acc_{\va^*}}{\partial p},
        \qquad
        p\in\{\alpha_+,\alpha_-,\beta\}.
    \end{equation}
    Substituting the group-wise derivative from Proposition~1 gives $\nabla_{\vp}\mathrm{WGA}$, and the normalized local update is
    \begin{equation}
        d_{\mathrm{WGA}}
        =
        \frac{\nabla_{\vp}\mathrm{WGA}}
        {\|\nabla_{\vp}\mathrm{WGA}\|_2}.
    \end{equation}
    While WGA employs the minimum operator, it is non-smooth and becomes non-differentiable when the worst-performing group shifts, making a global gradient-based analysis inapplicable. 

    Table~\ref{tab:aa-wga-direction} compares the two objectives on Waterbirds-syn Example~1 where the local smoothness is ensured. The WGA-derived direction gives a small increase in WGA at step 15, whereas their WGA-AUC and AA results are similar. Thus the AA-derived direction provides comparable allocation behavior while retaining a smooth objective.
    \begin{table}[t]
    \centering
    \caption{AA-derived versus WGA-derived allocation on Waterbirds-syn Example~1. Results are reported as mean $\pm$ standard deviation over three seeds.}
    \label{tab:aa-wga-direction}
    \small
    \setlength{\tabcolsep}{5pt}
    \begin{tabular}{lcccc}
        \toprule
        Direction & AA@15 & WGA@15 & AA-AUC & WGA-AUC \\
        \midrule
        AA-derived
            & $92.45{\pm}0.29$ & $83.93{\pm}1.03$ & $22.81{\pm}0.09$ & $20.05{\pm}0.24$ \\
        WGA-derived
            & $92.37{\pm}0.60$ & $84.29{\pm}1.55$ & $22.77{\pm}0.12$ & $19.99{\pm}0.36$ \\
        \bottomrule
    \end{tabular}
\end{table}

\subsection{Comparison with a Data-allocation Baseline}\label{appd:allocation-baseline}
    We compare Grad with the scaling-law allocation method of~\cite{rolf2021representation}, adapted to our local-update setting. At each direction estimate, the adapted method adds 10\% of the current data to each subgroup in turn, trains the resulting models, and uses the measured performance changes to estimate a normalized update direction. We re-estimate this direction only at evaluation checkpoints because it requires one vanilla and four additional training runs, whereas Grad is updated analytically at negligible cost.

    Table~\ref{tab:allocation-baseline} reports results on Waterbirds-syn. Grad outperforms the adapted baseline in AA and WGA at step 15 as well as in both trajectory-level AUC metrics, while requiring no additional model fits to estimate each direction.
    \begin{table}[t]
    \centering
    \caption{Comparison with the adapted scaling-law data-allocation baseline~\cite{rolf2021representation} on Waterbirds-syn. AA and WGA are measured after 15 allocation steps.}
    \label{tab:allocation-baseline}
    \small
    \setlength{\tabcolsep}{7pt}
    \begin{tabular}{lcccc}
        \toprule
        Method & AA & WGA & AA-AUC & WGA-AUC \\
        \midrule
        Scaling-law allocation
            & 91.78 & 79.38 & 22.69 & 19.40 \\
        Grad (ours)
            & 92.50 & 84.15 & 22.78 & 19.95 \\
        \bottomrule
    \end{tabular}
\end{table}

\subsection{Group Sizes in Grad-guided Resampling.}\label{appd:app}
    Table~\ref{tab:resampling_group_sizes} reports the subgroup sizes used in the resampling experiment in Table~\ref{tb:use-case}. Vanilla over-sampling and under-sampling target the fully balanced configuration by matching all groups to the largest or smallest original group size, respectively. In contrast, Grad-guided resampling targets the configuration induced by the derived Grad direction, which does not necessarily move all groups toward equal size. The values in parentheses show the change relative to the original data, highlighting that Grad-guided resampling can allocate samples unevenly across groups while targeting a configuration predicted to improve robust performance.

    \begin{table}[t]
    \centering
    \caption{
    Group sizes for the Waterbirds-syn data variants used in the resampling experiment.
    Values in parentheses indicate the change relative to the original group size.
    }
    \label{tab:resampling_group_sizes}
    \begin{tabular}{lcccc}
        \toprule
        Method & Group 1 & Group 2 & Group 3 & Group 4 \\
        \midrule
        Original 
        & 1124 & 28 & 6404 & 1444 \\
        Over-sampling 
        & 6404 (+5280) & 6404 (+6376) & 6404 (+0) & 6404 (+4960) \\
        Over-sampling (Grad) 
        & 5799 (+4675) & 857 (+829) & 6404 (+0) & 811 (-633) \\
        Under-sampling 
        & 28 (-1096) & 28 (+0) & 28 (-6376) & 28 (-1416) \\
        Under-sampling (Grad) 
        & 190 (-934) & 28 (+0) & 209 (-6195) & 27 (-1417) \\
        \bottomrule
    \end{tabular}
\end{table}

\subsection{Statistical Significance Test}\label{appd:stats-test}
Table~\ref{tab:significance} reports two-sided tests for the AA comparisons in Figure~\ref{fig:allocation}. Toward the end, several configurations can converge to similar performance levels, causing the curves for different directions to overlap. Therefore, we report the p-values for the Grad/Bal results after 15 update steps. This choice represents a moderate update magnitude while avoiding the effects of convergence. The results indicate that most configurations achieve statistically significant improvements, with $p < 0.05$. Moreover, we compare Grad and Bal using the area under the performance curve (AUC) with adjusted accuracy. Unlike performance at a single update step, AUC summarizes the entire allocation trajectory, capturing the efficiency of data allocations. The AUC analysis provides more consistent evidence of Grad's advantage across datasets and experimental settings.

\begin{table}[t]
    \centering
    \caption{$p$-values for Grad versus Bal in AA at step 15 / AA-AUC over the allocation trajectory (three seeds).}
    \label{tab:significance}
    \small
    \setlength{\tabcolsep}{4pt}
    \begin{tabular}{lcccc}
        \toprule
        & Waterbirds-syn & UrbanCars-syn & CelebA-subset & CivilComments-subset \\
        \midrule
        Example 1 & .055 / .062 & .066 / .028 & .003 / .003 & .056 / .060 \\
        Example 2 & .012 / .0001 & .012 / .005 & .001 / .002 & .246 / .010 \\
        Example 3 & .037 / .017 & .081 / .024 & .007 / .030 & .002 / .011 \\
        \bottomrule
    \end{tabular}
\end{table}

\subsection{Integrating Mitigation Methods}\label{appd:gdro}
    As noted in the main paper, prior work mainly focuses on model-side mitigation, whereas our work studies the data side: how subgroup configurations affect downstream robustness and how to optimize them. Therefore, a direct comparison with these mitigation methods confounds model and data effects and obscures the contribution of our analysis.

\begin{figure}[h]
    \centering
    \includegraphics[width=\linewidth]{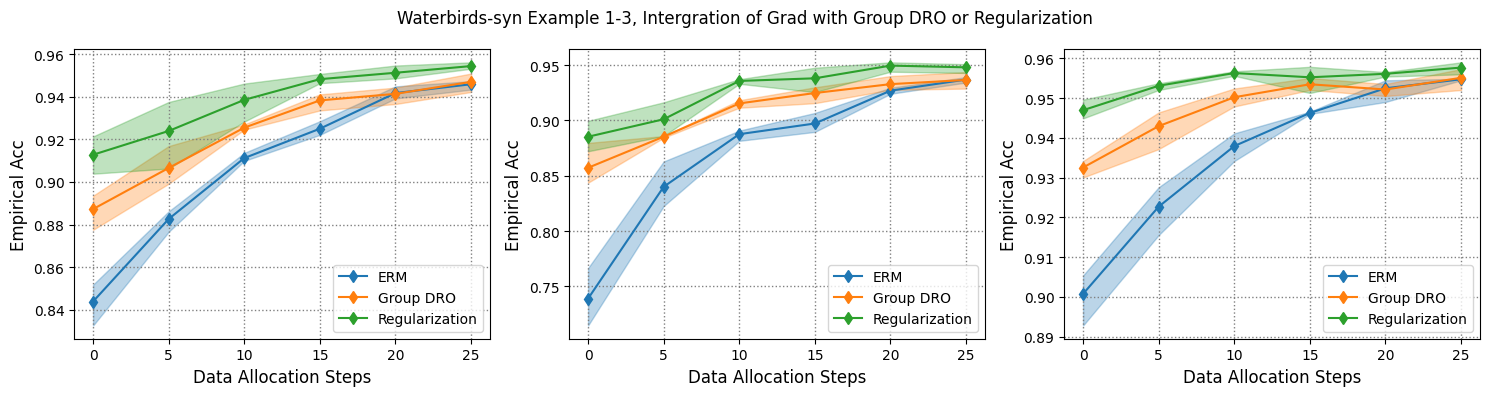}
    \caption{Integration of Group DRO and Regularization with Our Method of Allocating Data.}
    \label{appd:fig:integration}
\end{figure}

    In the meantime, these methods can be integrated into our method: One can first improve the data distribution along the Grad direction and then apply mitigation methods on top to further boost the performance. We added Group DRO and regularization results on Waterbirds-syn using the setup of Figure~\ref{fig:allocation}a, shown in Figure~\ref{appd:fig:integration}, and these methods with allocation yield further gains over ERM and the same methods without additional allocation.

\subsection{Robustness to Separability Estimation Errors}\label{appd:est-error}
    As we claimed in the paper, the Sign direction is more robust to the errors in separability estimation. We demonstrated this by showing the data allocation for Waterbirds-syn Example 2, where we perturb the estimated separability by multiplying each of them by 0.9 or 1.1, which produces 9 combinations. The results show that the Grad direction shows a certain randomness across different separability combinations, but overall still outperforms the results obtained from Bal. In contrast, Sign yields the same results across the different separability combinations, validating its robustness to separability estimation errors.
    
    \begin{figure}[h]
    \centering
    \includegraphics[width=0.4\linewidth]{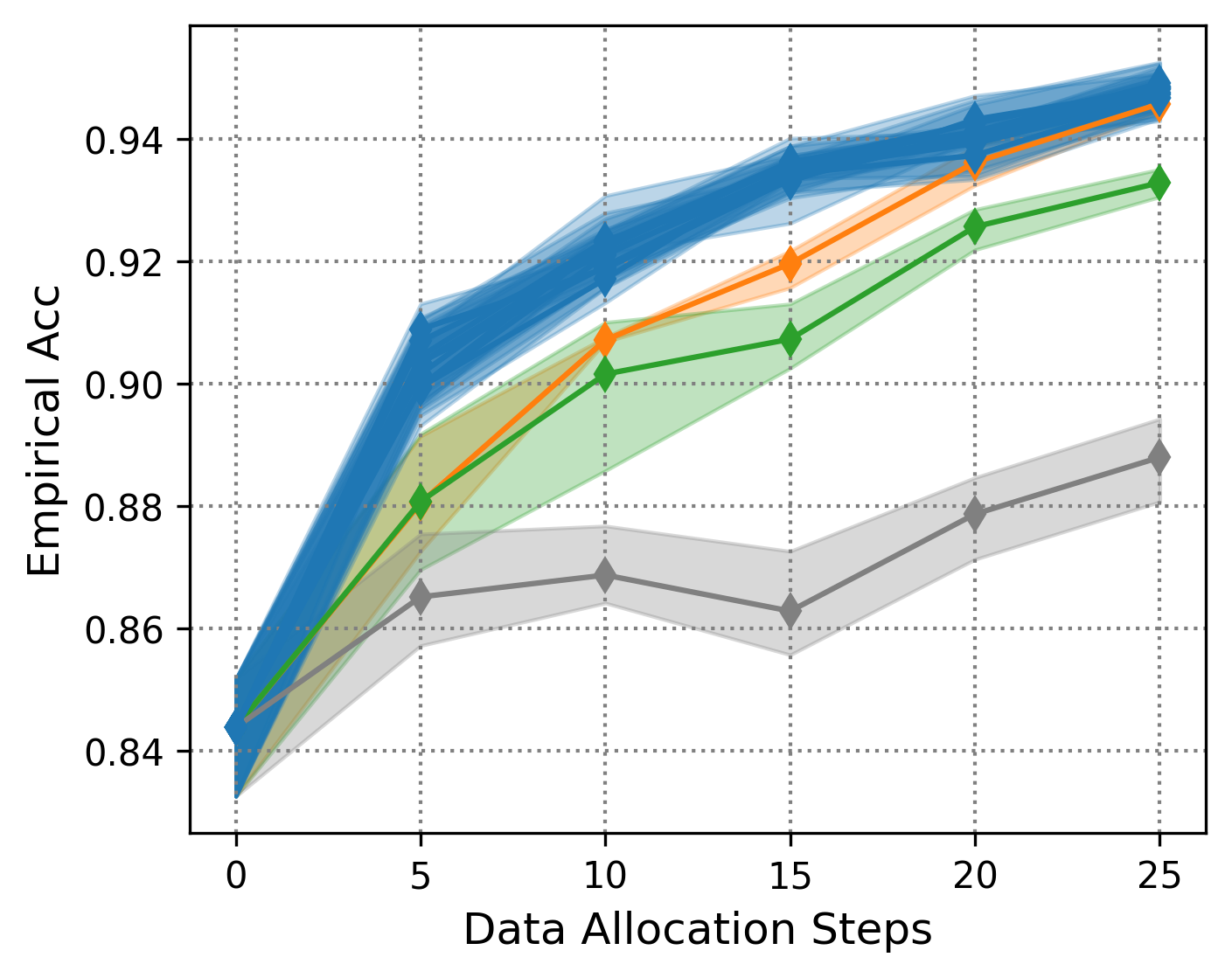}
    \caption{Grad Robustness to Separability Estimation Error (Overall).}
    \label{appd:fig:est-error1}
\end{figure}

\begin{figure*}[h]
    \centering

    \begin{subfigure}{0.30\linewidth}
        \includegraphics[width=\linewidth]{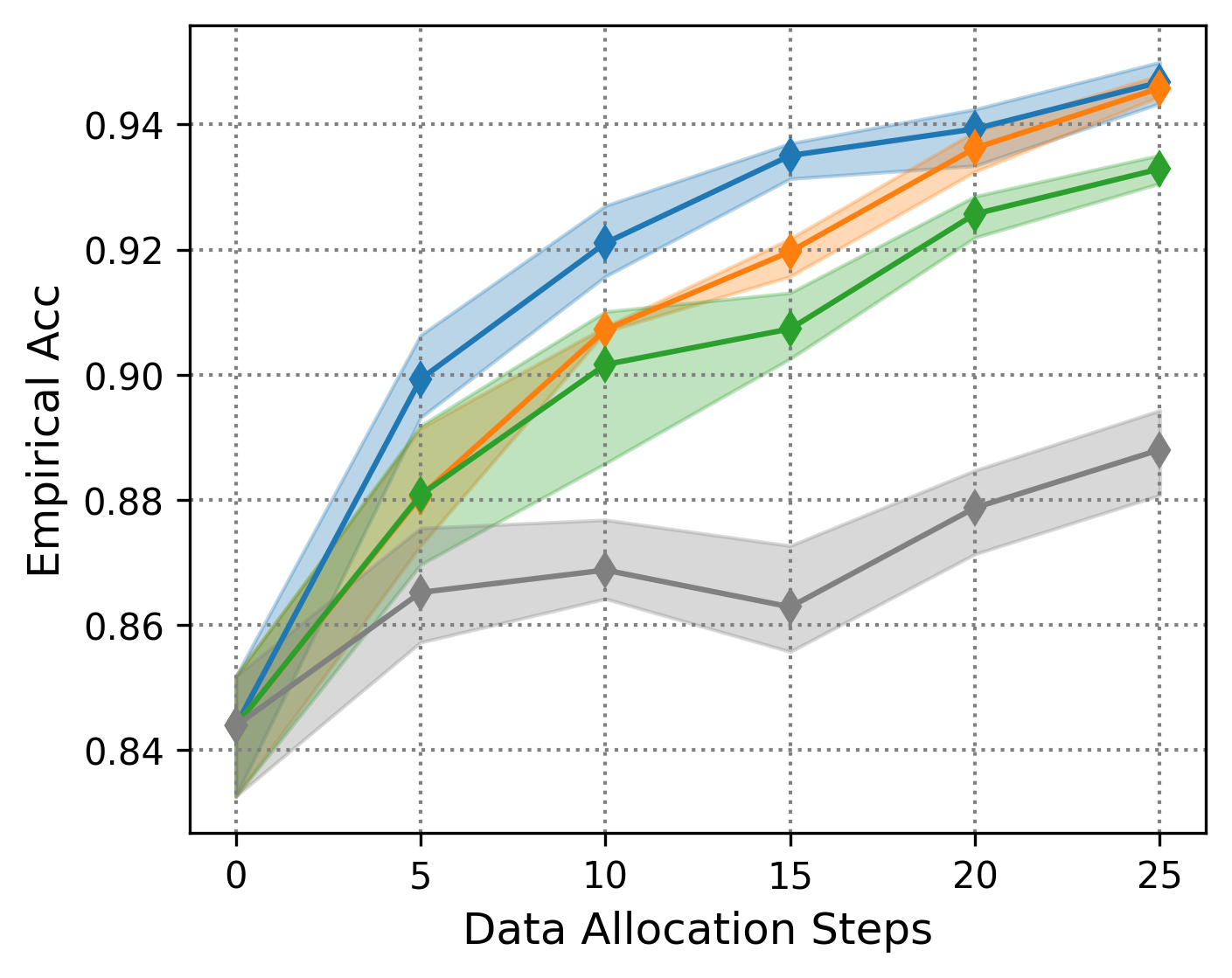}
        \caption{$m_1, m_2$}
    \end{subfigure}
    \hfill
    \begin{subfigure}{0.30\linewidth}
        \includegraphics[width=\linewidth]{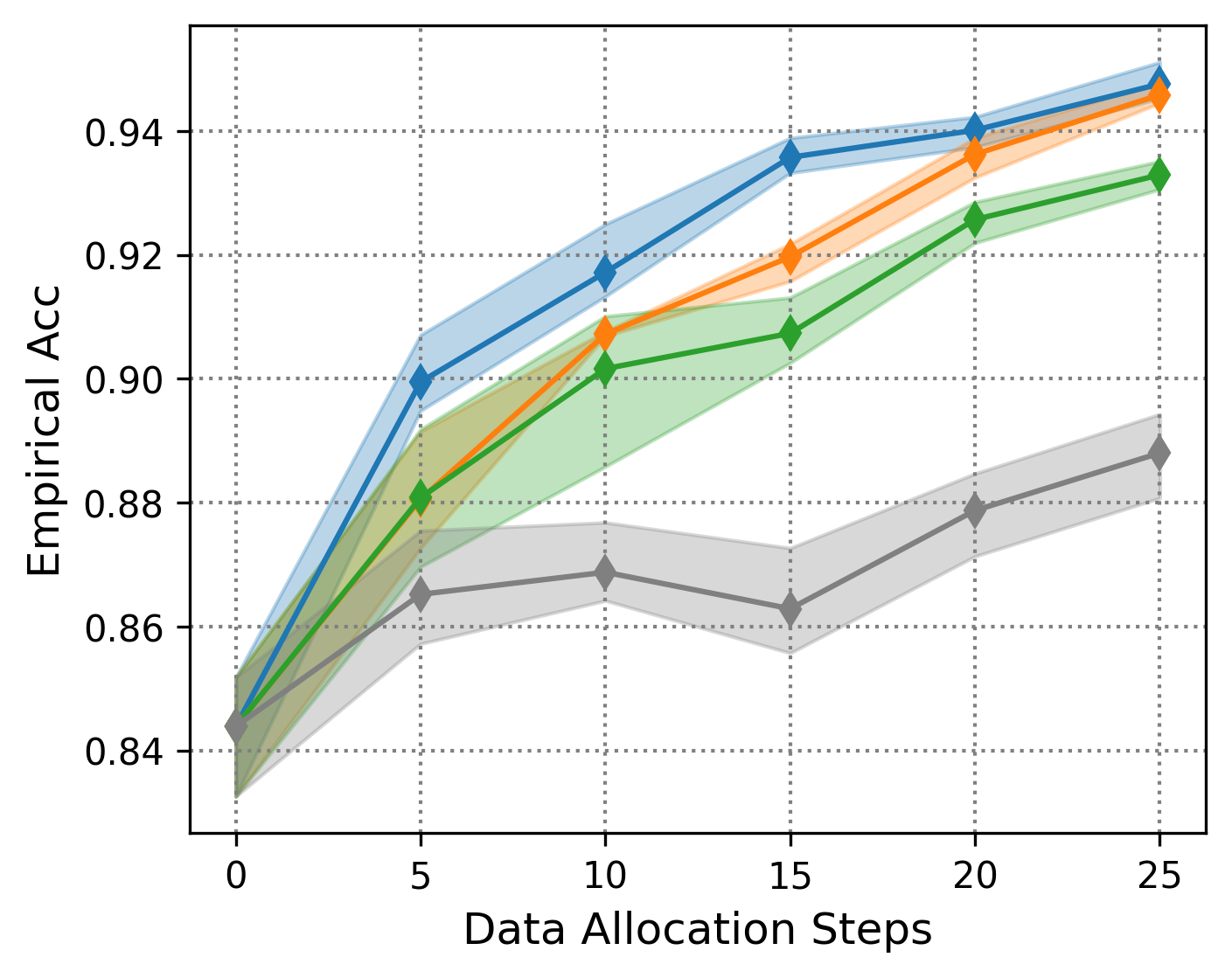}
        \caption{$m_1, 0.9 \cdot m_2$}
    \end{subfigure}
    \hfill
    \begin{subfigure}{0.30\linewidth}
        \includegraphics[width=\linewidth]{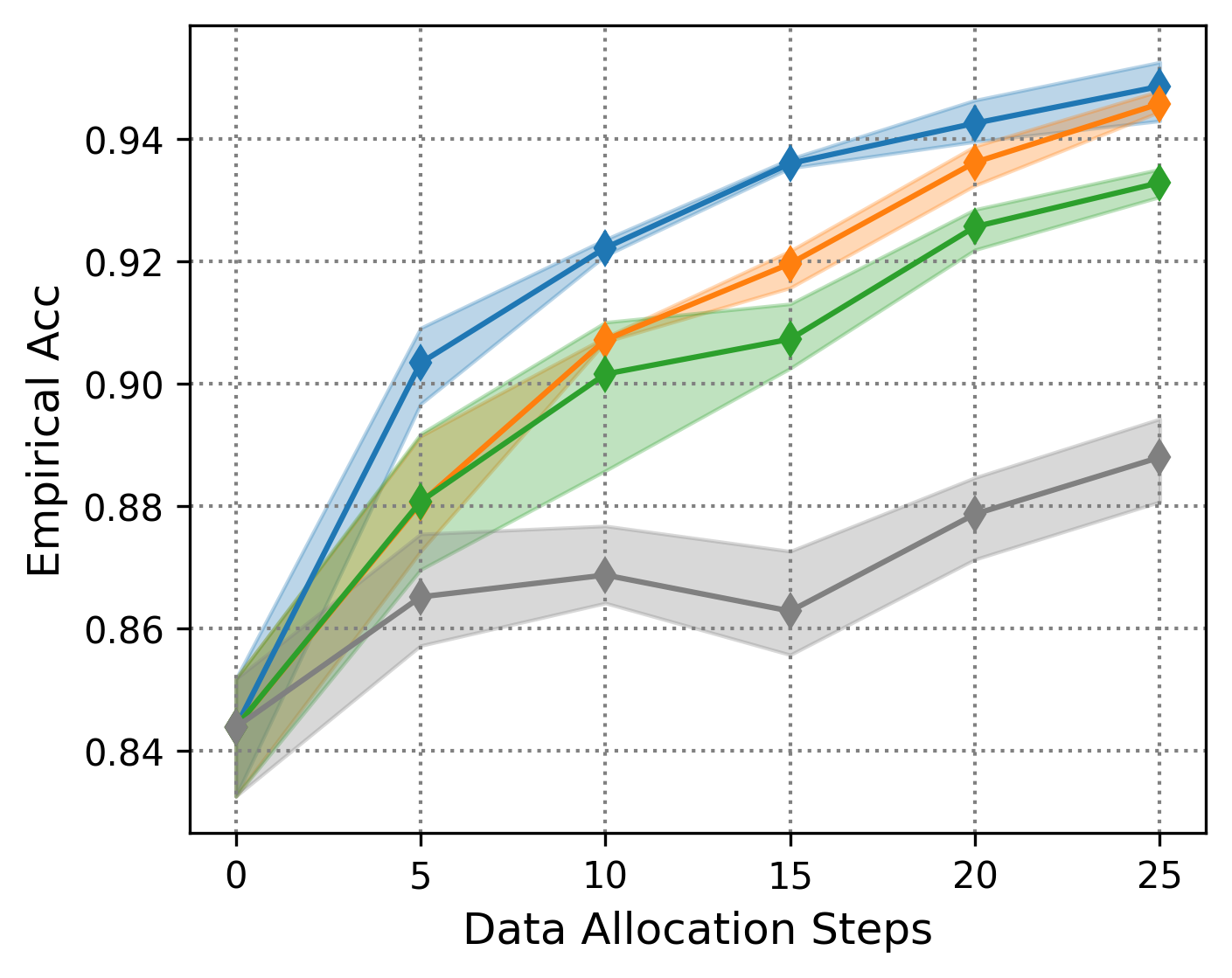}
        \caption{$m_1, 1.1 \cdot m_2$}
    \end{subfigure}

    \vspace{0.5em}

    \begin{subfigure}{0.30\linewidth}
        \includegraphics[width=\linewidth]{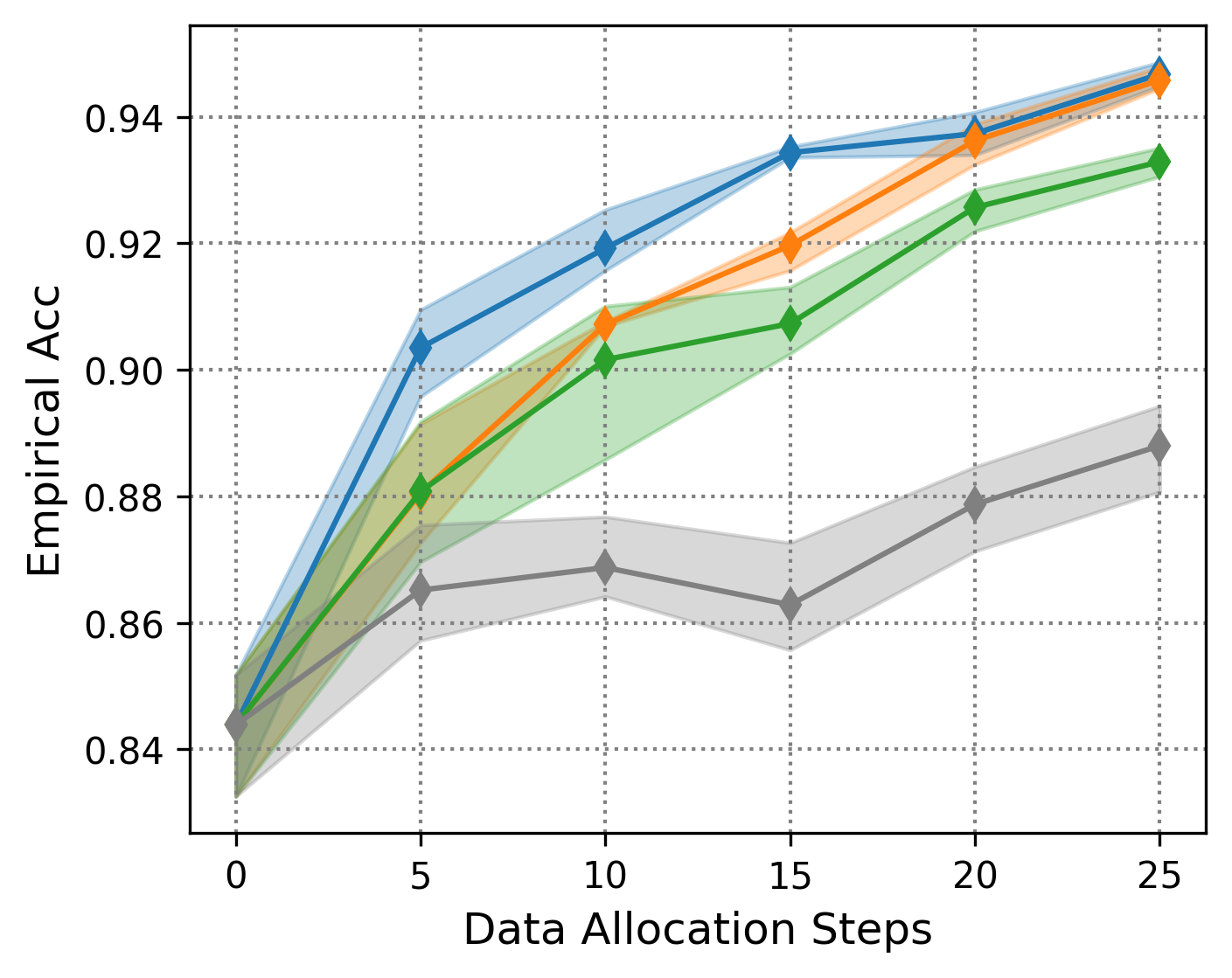}
        \caption{$0.9 \cdot m_1, m_2$}
    \end{subfigure}
    \hfill
    \begin{subfigure}{0.30\linewidth}
        \includegraphics[width=\linewidth]{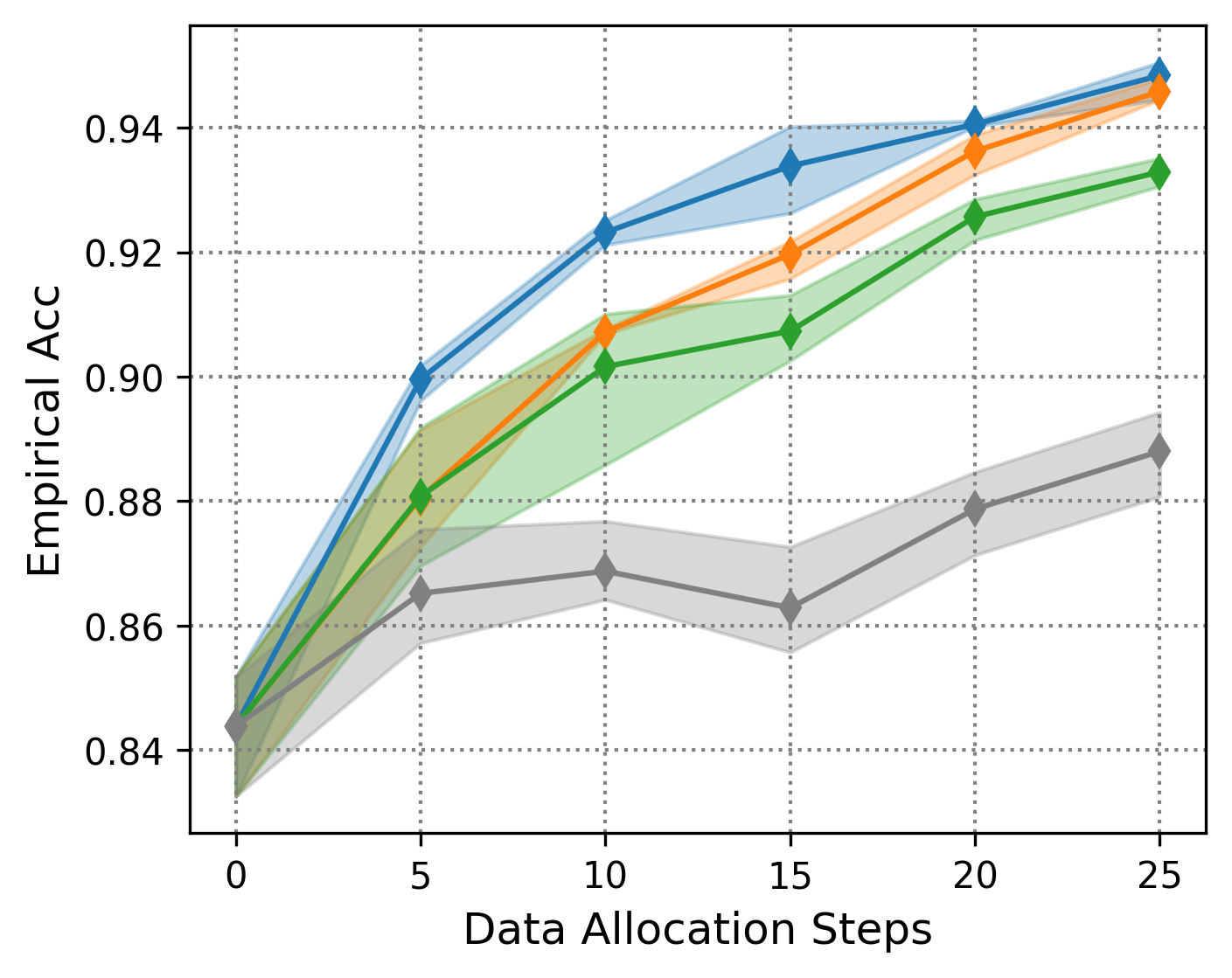}
        \caption{$0.9 \cdot m_1, 0.9 \cdot m_2$}
    \end{subfigure}
    \hfill
    \begin{subfigure}{0.30\linewidth}
        \includegraphics[width=\linewidth]{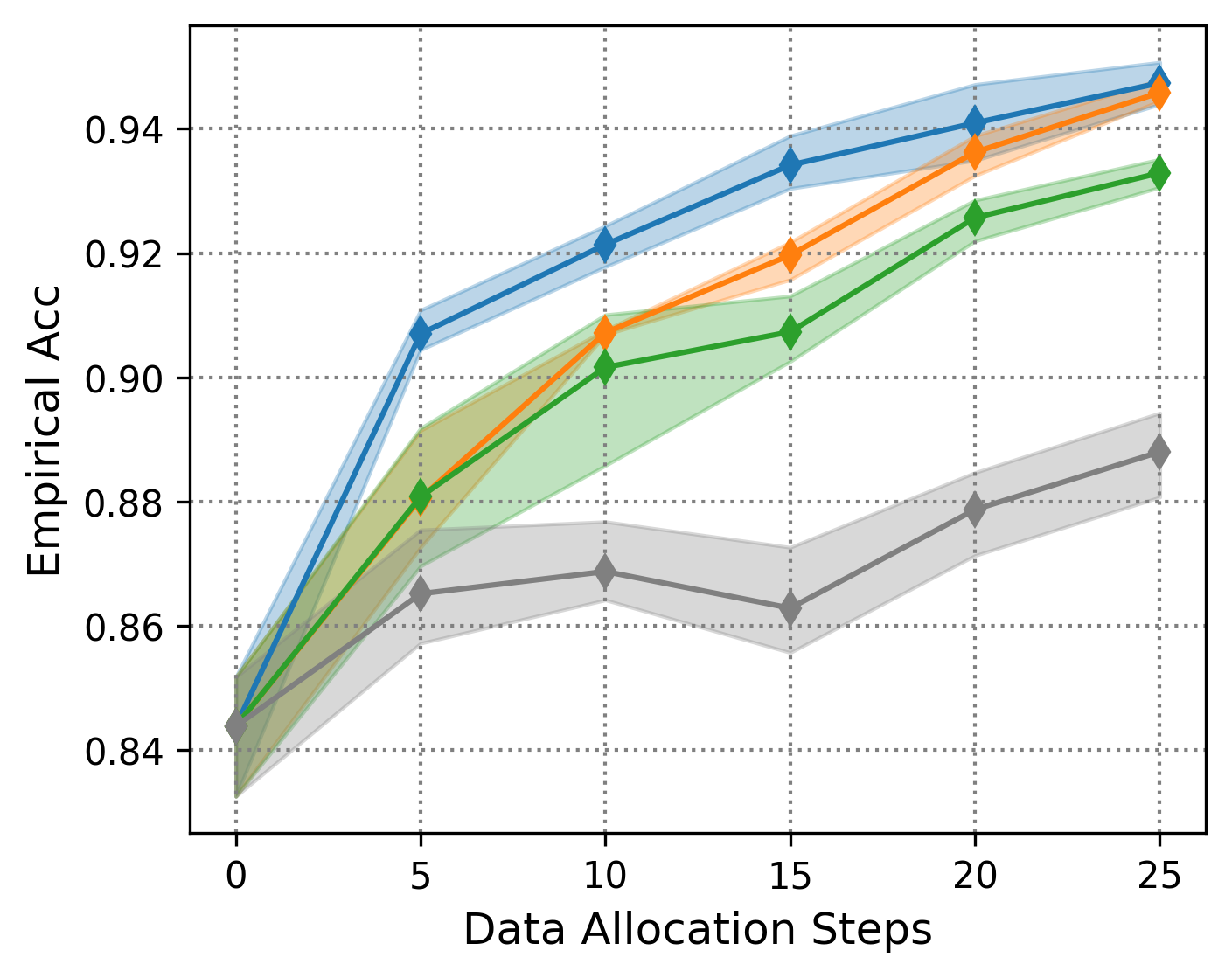}
        \caption{$0.9 \cdot m_1, 1.1 \cdot m_2$}
    \end{subfigure}

    \vspace{0.5em}

    \begin{subfigure}{0.30\linewidth}
        \includegraphics[width=\linewidth]{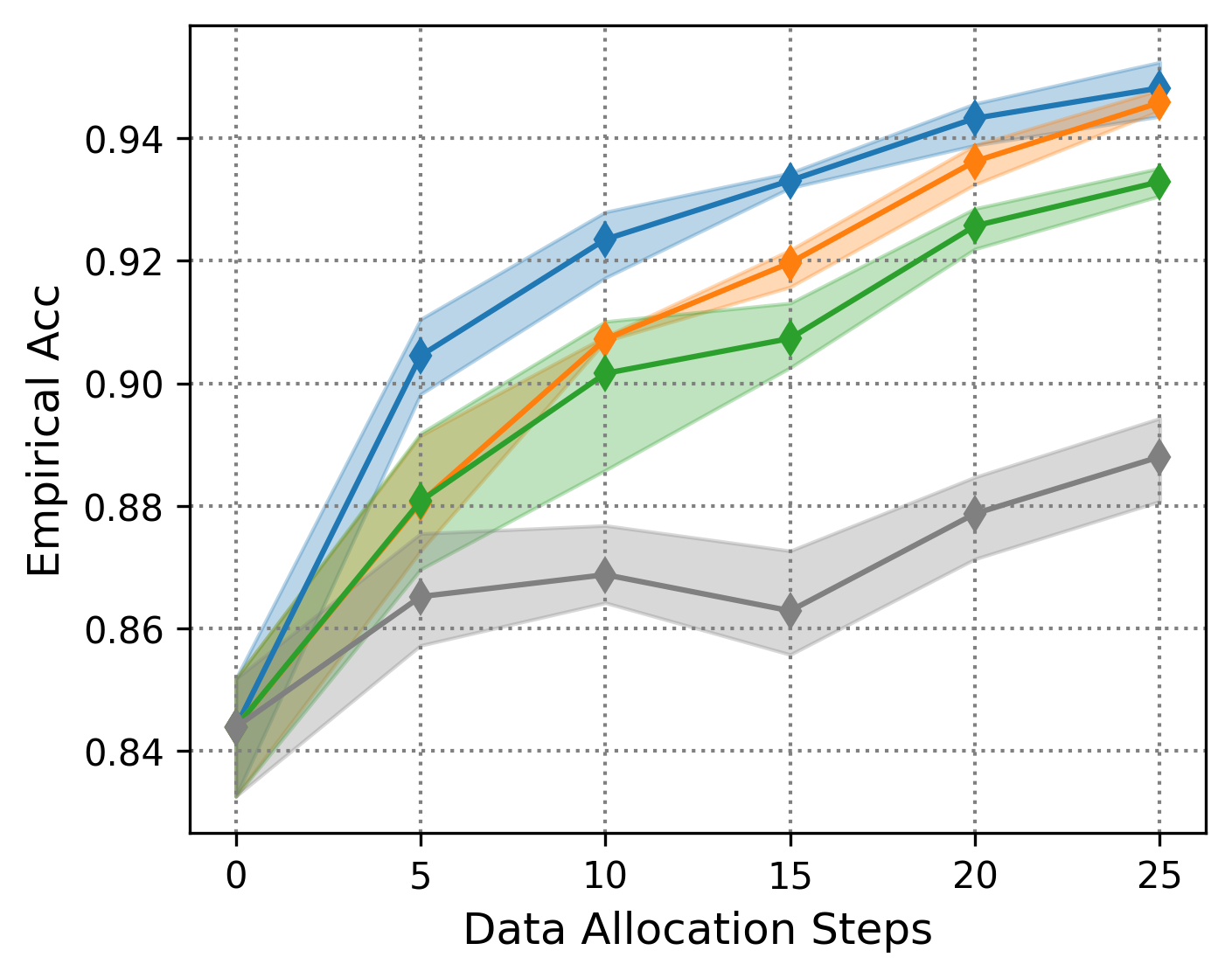}
        \caption{$1.1 \cdot m_1, m_2$}
    \end{subfigure}
    \hfill
    \begin{subfigure}{0.30\linewidth}
        \includegraphics[width=\linewidth]{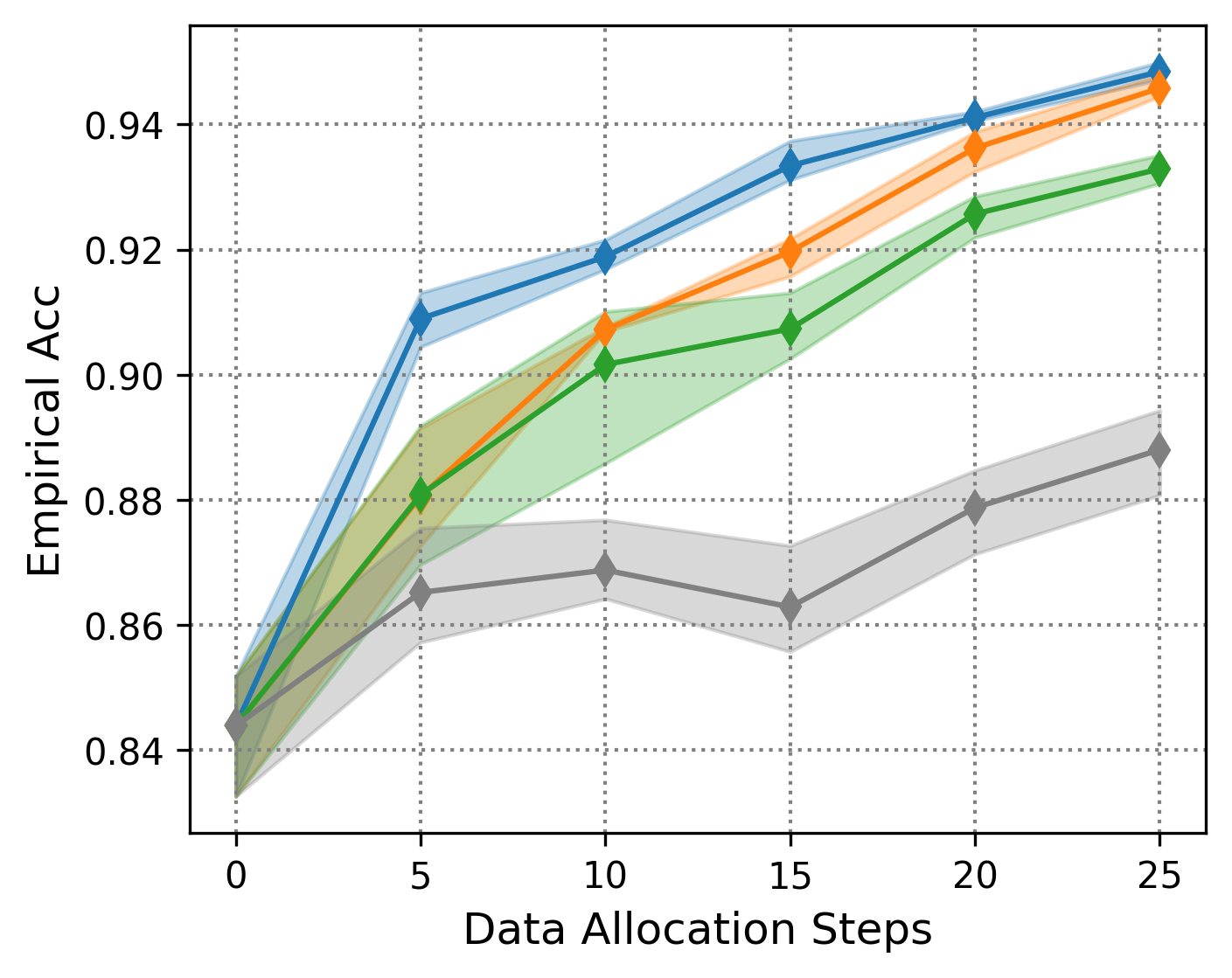}
        \caption{$1.1 \cdot m_1, 0.9 \cdot m_2$}
    \end{subfigure}
    \hfill
    \begin{subfigure}{0.30\linewidth}
        \includegraphics[width=\linewidth]{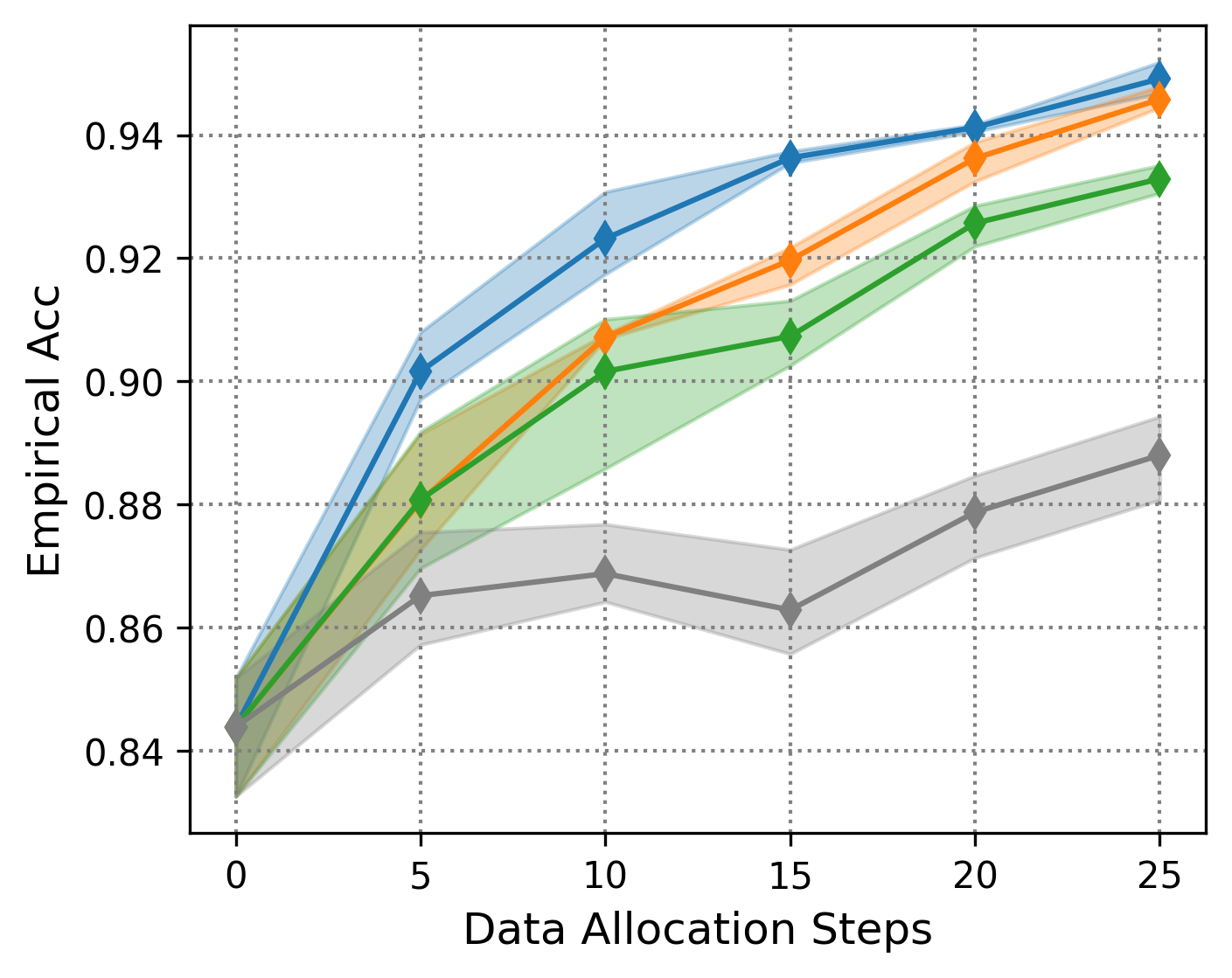}
        \caption{$1.1 \cdot m_1, 1.1 \cdot m_2$}
    \end{subfigure}

    \caption{Grad Robustness to Separability Estimation Error. We vary the estimated separability by 0.9, 1, and 1.1 to synthesize the estimation error}
    \label{appd:fig:est-error2}
\end{figure*}

\subsection{Three-attribute Setup Demonstration}\label{appd:n3}
\begin{figure}[t!]
    \centering
    \begin{minipage}[t]{0.8\linewidth}
        \centering
        \begin{subfigure}{0.49\linewidth}
            \includegraphics[width=\linewidth]{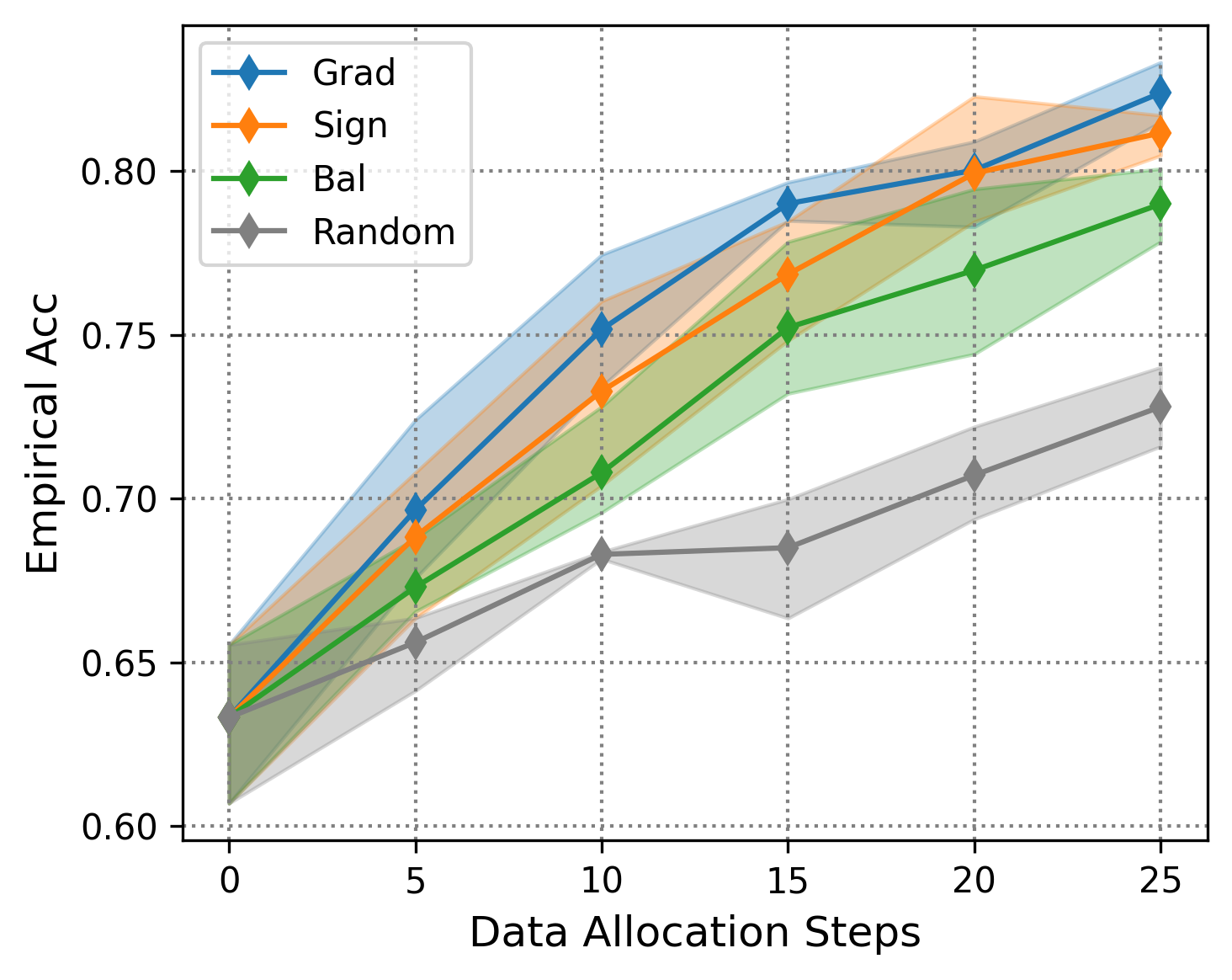}
            \caption{Adjusted Accuracy}
        \end{subfigure}
        \hfill
        \begin{subfigure}{0.49\linewidth}
            \includegraphics[width=\linewidth]{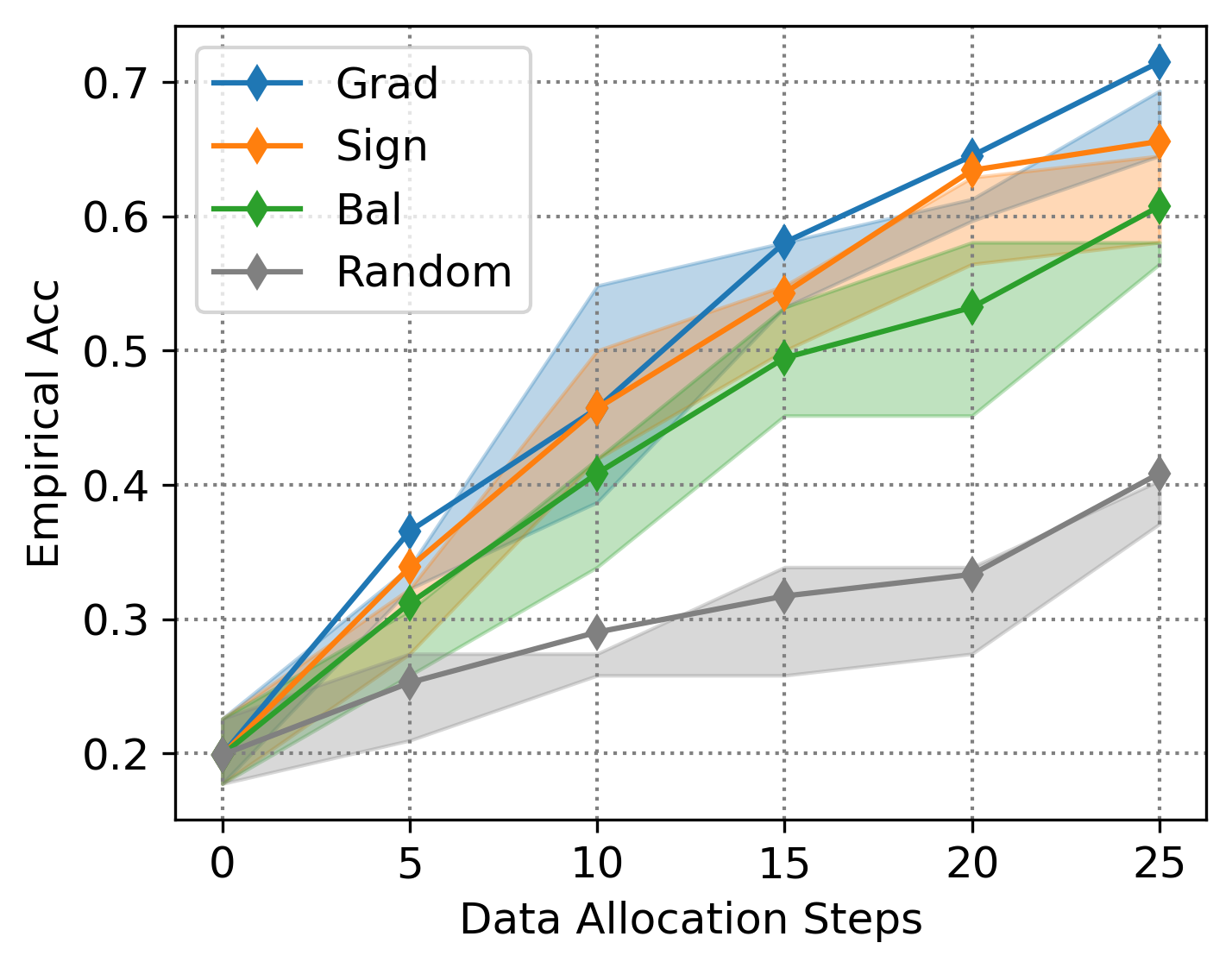}
            \caption{WGA}
        \end{subfigure}
    \caption{Data Allocation on UrbanCars-syn with Three Attributes.}
    \label{appd:fig:n3}
    \end{minipage}
\end{figure}

    We show the flexibility of our framework by demonstrating its effectiveness in a three-attribute setup. Our main text focuses on a two-attribute setup, which is commonly used in the study of subpopulation shifts and spurious correlation. 
    For the three-attribute setting, we extend UrbanCars-syn by incorporating both background and co-occurring objects as spurious attributes in addition to the object label, resulting in $2^3=8$ subgroups. The training configuration is parameterized by $\alpha^{(2)}_+,\alpha^{(2)}_-,\alpha^{(3)}_+,\alpha^{(3)}_-,\beta$, where $\alpha^{(2)}_{\pm}$ controls the alignment between the object label and background, $\alpha^{(3)}_{\pm}$ controls the alignment between the object label and co-occurring object, and $\beta$ controls the class prior. We first estimate the feature separabilities from empirical group accuracies obtained under a moderately imbalanced calibration configuration $(\alpha^{(2)}_+,\alpha^{(2)}_-,\alpha^{(3)}_+,\alpha^{(3)}_-,\beta) = (0.85, 0.85, 0.85, 0.85, 0.5)$. With the estimated separabilities fixed as $0.739, 0.274, 0.098$, we then evaluate data-allocation trajectories by updating the initial configuration parameters of $(0.976, 0.184, 0.976, 0.184, 0.128)$ along the derived gradient direction and comparing against other directions. The results are shown in Figure~\ref{appd:fig:n3}, where the estimated Grad and Sign also outperform the allocation along other directions.

\subsection{Additional Pretrained Encoders and Random Seeds}~\label{appd:encoder}

    In addition to our primary encoders (ResNet50 and BERT) in our main results, we additionally conducted experiments with Vision Transformer (ViT)~\cite{dosovitskiyimage} and DeBERTa~\cite{he2020deberta}. Instead of evaluating all $7^3=343$ configurations for optimality experiments, we tested the top 10\% configurations identified under ResNet50 (Figure 6a) on Waterbirds-syn with ViT and found multiple configurations comparable or better than the Balanced setup in Table~\ref{appd:tb:vit1}. 
    
    \begin{table}[t]
        \begin{center}
        \caption{Optimality experiments on Waterbirds-syn with top 10\% configurations on ResNet50.}
        \label{appd:tb:vit1}
        \setlength\tabcolsep{3pt}
        \resizebox{0.5\linewidth}{!}{
        \begin{tabular}{lcc}
        \toprule
         & Accuracy Range & Accuracy at Balanced Setup \\
        \midrule
        ResNet50 & 0.959 - 0.966 & 0.953 \\
        ViT & 0.958 - 0.973 & 0.961 \\
        \bottomrule
        \end{tabular}
       }
        \end{center}
    \end{table}
    
\begin{figure}
    \centering
    \includegraphics[width=\linewidth]{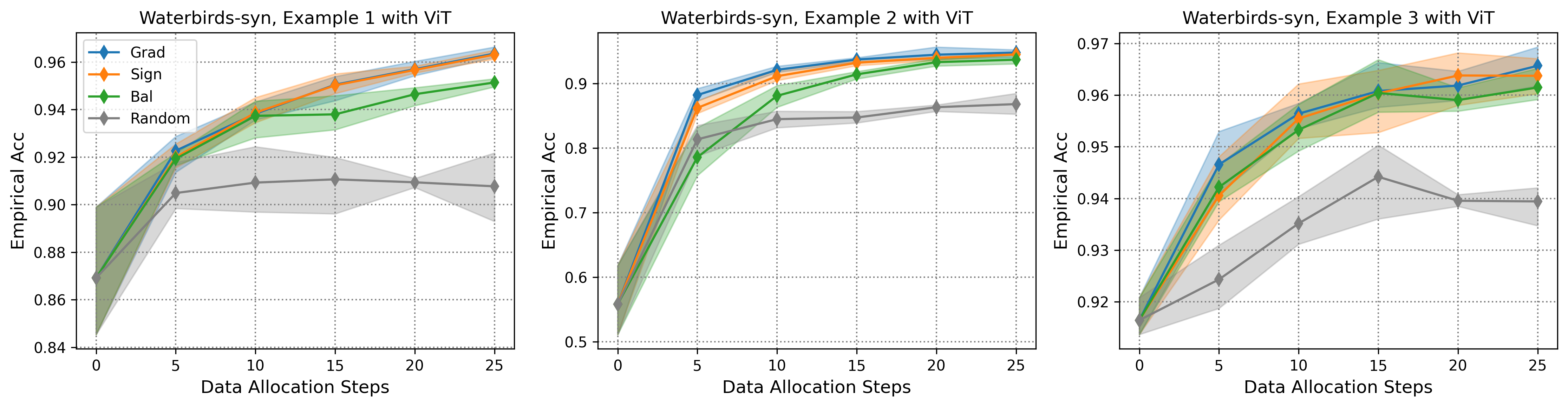}
    \caption{Efficiency of data allocation along directions across initial configurations on Waterbirds-syn with ViT}
    \label{appd:fig:vit-wb}
\end{figure}

\begin{figure}
    \centering
    \includegraphics[width=\linewidth]{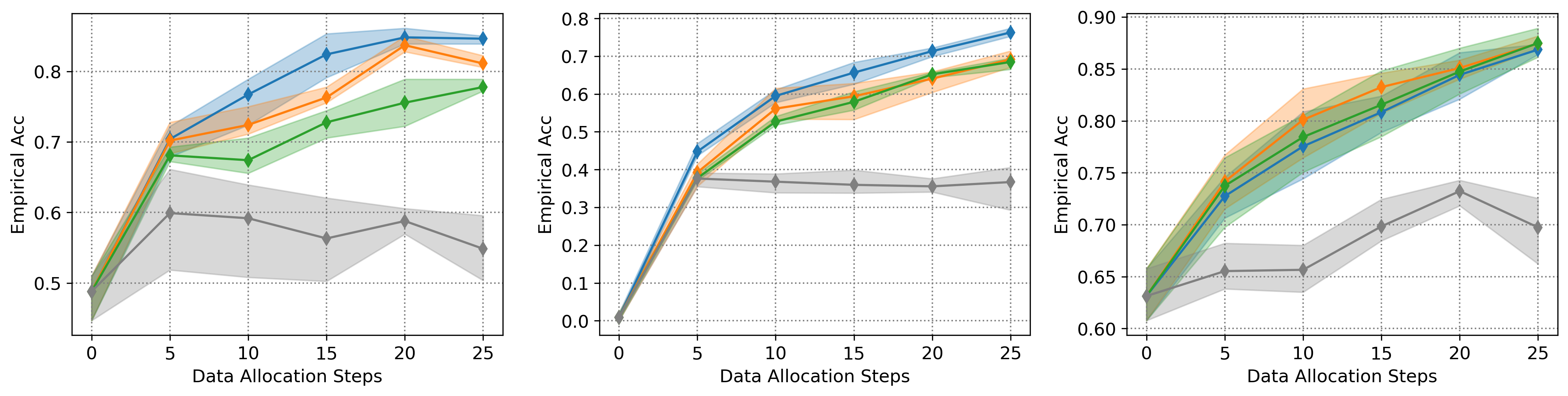}
    \caption{Efficiency of data allocation along directions across initial configurations on CelebA-subset with ViT}
    \label{appd:fig:vit-ca}
\end{figure}

\begin{figure}
    \centering
    \includegraphics[width=\linewidth]{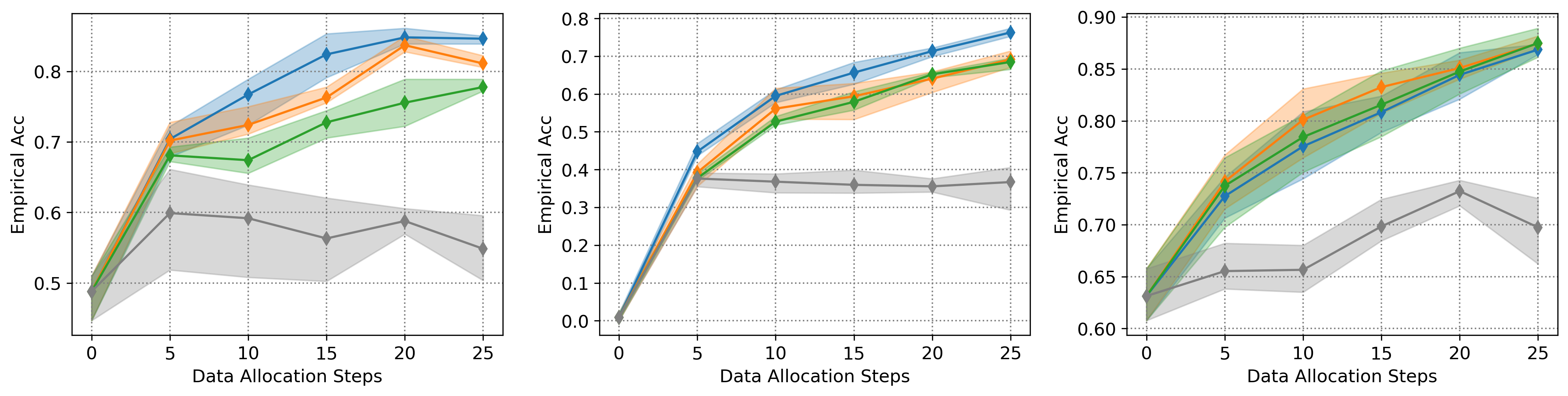}
    \caption{Efficiency of data allocation along directions across initial configurations on CivilComments-subset with DeBERTa}
    \label{appd:fig:cc}
\end{figure}

    We also provide the efficiency experiments with ViT/DeBERTa, whose trends are similar to those of ResNet50/BERT, in Figures~\ref{appd:fig:vit-wb}, ~\ref{appd:fig:vit-ca} and ~\ref{appd:fig:cc}.

    As Figure~\ref{fig:max_acc_emp} evaluates 343 configurations to exhaustively evaluate the optimality, it only uses one random seed to present the results. We report standard deviations over 3 random seeds for the configurations highlighted for near-optimal performance, showing that the visualized multi-optima is not an artifact of seed randomness.

\section{Broader Impacts}
    This work presents a data-centric perspective on achieving robust generalization, and our analysis guides the construction of datasets for better subpopulation robustness of downstream model training and deployment, particularly helpful in high-stakes domains where balance is infeasible to achieve. Although we focus on data-side effects without direct comparison with model-based methods to keep the theoretical insights clear, integrating our data-optimization principles with existing model-side approaches is a promising direction for real-world deployment. In Appendix~\ref{appd:gdro}, we provide an example showing that integrating our data allocation with Group DRO~\cite{sagawa2019distributionally} and Regularization~\cite{wang2024effect} can further improve performance. However, it could be misused to justify neglecting minority or vulnerable subgroups, so the method should be applied with caution and additional auditing.
    
    \begin{table}[h]
        \begin{center}
        \caption{Optimality experiments with 3 random seeds on configurations with top performance.}
        \label{appd:tb:seeds}
        \setlength\tabcolsep{3pt}
        \resizebox{0.5\linewidth}{!}{
        \begin{tabular}{lc}
        \toprule
        Dataset & Accuracy Range \\
        \midrule
        Waterbirds-syn & 0.954$\pm$ 0.003 -- 0.965$\pm$ 0.002 \\
        Urbancars-syn & 0.826$\pm$ 0.007 -- 0.844$\pm$ 0.007 \\
        CelebA-subset & 0.893$\pm$ 0.003 -- 0.902$\pm$ 0.002 \\
        CivilComments-subset & 0.796$\pm$ 0.003 -- 0.807$\pm$ 0.003 \\
        \bottomrule
        \end{tabular}
        }
        \end{center}
    \end{table}

    \clearpage
    \begin{minipage}[H]{0.6\textwidth}
\begin{algorithm}[H]
\small
    \caption{\small Direction estimation and configuration updating.}
    \label{alg}
    \definecolor{codeblue}{rgb}{0.25,0.5,0.5}
    \lstset{
      backgroundcolor=\color{white},
      basicstyle=\fontsize{7.2pt}{7.2pt}\ttfamily\selectfont,
      columns=fullflexible,
      breaklines=true,
      captionpos=b,
      commentstyle=\fontsize{7.2pt}{7.2pt}\color{codeblue},
      keywordstyle=\fontsize{7.2pt}{7.2pt},
      escapechar=\&%
    }
\begin{lstlisting}[language=python]
# Parameters:
#   m1, m2: Feature separability parameters
#   num_steps: Number of optimization iterations
#   delta: Step size for configuration updates

# Inputs:
#   alpha1, alpha0, beta: Initial training data configuration

# Initialize estimator with separability parameters
est = Estimator(m1, m2)

for step in range(num_steps):

    # 1. Estimate classifier parameters for the current configuration
    tau, b = est.estimate_classifier(alpha1, alpha0, beta)

    # 2. Estimate partial derivatives of accuracy
    #    w.r.t. (alpha1, alpha0, beta)
    d_alpha1, d_alpha0, d_beta = est.estimate_partials(
        alpha1, alpha0, beta, tau, b
    )

    # 3. Construct Update Directions
    # 3.1 Grad: Standard gradient ascent for smoothest path
    grad = normalize([d_alpha1, d_alpha0, d_beta])      
    
    # 3.2 Sign: Robust update using only derivative signs
    sign = normalize([d_alpha1.sign(), d_alpha0.sign(), d_beta.sign()])
    
    # 3.3 Balancing: Heuristic direction toward the balanced configuration
    ratios = param2ratio([alpha1, alpha0, beta])
    bal_r = ratios - [0.25, 0.25, 0.25, 0.25] 
    bal = normalize(transform(bal_r))
    
    # 4. Update step (example using Grad)
    [alpha1, alpha0, beta] += delta * grad

# Output the optimized configuration
return alpha1, alpha0, beta
\end{lstlisting}
\end{algorithm}
\end{minipage}

\newpage
\section*{NeurIPS Paper Checklist}

\begin{enumerate}

\item {\bf Claims}
    \item[] Question: Do the main claims made in the abstract and introduction accurately reflect the paper's contributions and scope?
    \item[] Answer: \answerYes{} %
    \item[] Justification: The abstract and introduction state the main claims: balanced subgroup data is not uniquely optimal, and the paper derives gradient-based directions for adjusting subgroup proportions, and the claims are validated with controlled image/text experiments.
    \item[] Guidelines:
    \begin{itemize}
        \item The answer \answerNA{} means that the abstract and introduction do not include the claims made in the paper.
        \item The abstract and/or introduction should clearly state the claims made, including the contributions made in the paper and important assumptions and limitations. A \answerNo{} or \answerNA{} answer to this question will not be perceived well by the reviewers. 
        \item The claims made should match theoretical and experimental results, and reflect how much the results can be expected to generalize to other settings. 
        \item It is fine to include aspirational goals as motivation as long as it is clear that these goals are not attained by the paper. 
    \end{itemize}

\item {\bf Limitations}
    \item[] Question: Does the paper discuss the limitations of the work performed by the authors?
    \item[] Answer: \answerYes{} %
    \item[] Justification: The paper includes a discussion on the limitations in the end of the main paper: the broader evaluation with more attributes and the separability estimation error. The appendix also presents some results addressing the limitation.
    \item[] Guidelines:
    \begin{itemize}
        \item The answer \answerNA{} means that the paper has no limitation while the answer \answerNo{} means that the paper has limitations, but those are not discussed in the paper. 
        \item The authors are encouraged to create a separate ``Limitations'' section in their paper.
        \item The paper should point out any strong assumptions and how robust the results are to violations of these assumptions (e.g., independence assumptions, noiseless settings, model well-specification, asymptotic approximations only holding locally). The authors should reflect on how these assumptions might be violated in practice and what the implications would be.
        \item The authors should reflect on the scope of the claims made, e.g., if the approach was only tested on a few datasets or with a few runs. In general, empirical results often depend on implicit assumptions, which should be articulated.
        \item The authors should reflect on the factors that influence the performance of the approach. For example, a facial recognition algorithm may perform poorly when image resolution is low or images are taken in low lighting. Or a speech-to-text system might not be used reliably to provide closed captions for online lectures because it fails to handle technical jargon.
        \item The authors should discuss the computational efficiency of the proposed algorithms and how they scale with dataset size.
        \item If applicable, the authors should discuss possible limitations of their approach to address problems of privacy and fairness.
        \item While the authors might fear that complete honesty about limitations might be used by reviewers as grounds for rejection, a worse outcome might be that reviewers discover limitations that aren't acknowledged in the paper. The authors should use their best judgment and recognize that individual actions in favor of transparency play an important role in developing norms that preserve the integrity of the community. Reviewers will be specifically instructed to not penalize honesty concerning limitations.
    \end{itemize}

\item {\bf Theory assumptions and proofs}
    \item[] Question: For each theoretical result, does the paper provide the full set of assumptions and a complete (and correct) proof?
    \item[] Answer: \answerYes{} %
    \item[] Justification: The paper uses the Gaussian feature-modeling assumptions and Bayesian classifier setup; proofs and derivative details are provided in the appendix.
    \item[] Guidelines:
    \begin{itemize}
        \item The answer \answerNA{} means that the paper does not include theoretical results. 
        \item All the theorems, formulas, and proofs in the paper should be numbered and cross-referenced.
        \item All assumptions should be clearly stated or referenced in the statement of any theorems.
        \item The proofs can either appear in the main paper or the supplemental material, but if they appear in the supplemental material, the authors are encouraged to provide a short proof sketch to provide intuition. 
        \item Inversely, any informal proof provided in the core of the paper should be complemented by formal proofs provided in appendix or supplemental material.
        \item Theorems and Lemmas that the proof relies upon should be properly referenced. 
    \end{itemize}

    \item {\bf Experimental result reproducibility}
    \item[] Question: Does the paper fully disclose all the information needed to reproduce the main experimental results of the paper to the extent that it affects the main claims and/or conclusions of the paper (regardless of whether the code and data are provided or not)?
    \item[] Answer: \answerYes{} %
    \item[] Justification: The paper describes the datasets, constructed subsets/synthetic variants, evaluation metrics, model backbones, direction baselines, and allocation protocol in sufficient detail to reproduce the main experiments.
    \item[] Guidelines:
    \begin{itemize}
        \item The answer \answerNA{} means that the paper does not include experiments.
        \item If the paper includes experiments, a \answerNo{} answer to this question will not be perceived well by the reviewers: Making the paper reproducible is important, regardless of whether the code and data are provided or not.
        \item If the contribution is a dataset and\slash or model, the authors should describe the steps taken to make their results reproducible or verifiable. 
        \item Depending on the contribution, reproducibility can be accomplished in various ways. For example, if the contribution is a novel architecture, describing the architecture fully might suffice, or if the contribution is a specific model and empirical evaluation, it may be necessary to either make it possible for others to replicate the model with the same dataset, or provide access to the model. In general. releasing code and data is often one good way to accomplish this, but reproducibility can also be provided via detailed instructions for how to replicate the results, access to a hosted model (e.g., in the case of a large language model), releasing of a model checkpoint, or other means that are appropriate to the research performed.
        \item While NeurIPS does not require releasing code, the conference does require all submissions to provide some reasonable avenue for reproducibility, which may depend on the nature of the contribution. For example
        \begin{enumerate}
            \item If the contribution is primarily a new algorithm, the paper should make it clear how to reproduce that algorithm.
            \item If the contribution is primarily a new model architecture, the paper should describe the architecture clearly and fully.
            \item If the contribution is a new model (e.g., a large language model), then there should either be a way to access this model for reproducing the results or a way to reproduce the model (e.g., with an open-source dataset or instructions for how to construct the dataset).
            \item We recognize that reproducibility may be tricky in some cases, in which case authors are welcome to describe the particular way they provide for reproducibility. In the case of closed-source models, it may be that access to the model is limited in some way (e.g., to registered users), but it should be possible for other researchers to have some path to reproducing or verifying the results.
        \end{enumerate}
    \end{itemize}

\item {\bf Open access to data and code}
    \item[] Question: Does the paper provide open access to the data and code, with sufficient instructions to faithfully reproduce the main experimental results, as described in supplemental material?
    \item[] Answer: \answerYes{} %
    \item[] Justification: The paper uses existing public datasets and provides implementation details in the appendix. We release the code for the theoretical estimation and empirical validation on GitHub.
    \item[] Guidelines:
    \begin{itemize}
        \item The answer \answerNA{} means that paper does not include experiments requiring code.
        \item Please see the NeurIPS code and data submission guidelines (\url{https://neurips.cc/public/guides/CodeSubmissionPolicy}) for more details.
        \item While we encourage the release of code and data, we understand that this might not be possible, so \answerNo{} is an acceptable answer. Papers cannot be rejected simply for not including code, unless this is central to the contribution (e.g., for a new open-source benchmark).
        \item The instructions should contain the exact command and environment needed to run to reproduce the results. See the NeurIPS code and data submission guidelines (\url{https://neurips.cc/public/guides/CodeSubmissionPolicy}) for more details.
        \item The authors should provide instructions on data access and preparation, including how to access the raw data, preprocessed data, intermediate data, and generated data, etc.
        \item The authors should provide scripts to reproduce all experimental results for the new proposed method and baselines. If only a subset of experiments are reproducible, they should state which ones are omitted from the script and why.
        \item At submission time, to preserve anonymity, the authors should release anonymized versions (if applicable).
        \item Providing as much information as possible in supplemental material (appended to the paper) is recommended, but including URLs to data and code is permitted.
    \end{itemize}

\item {\bf Experimental setting/details}
    \item[] Question: Does the paper specify all the training and test details (e.g., data splits, hyperparameters, how they were chosen, type of optimizer) necessary to understand the results?
    \item[] Answer: \answerYes{} %
    \item[] Justification: The appendix specifies model architectures/backbones, optimizers, learning rates, weight decay, momentum, batch sizes, training steps, and evaluation metrics.
    \item[] Guidelines:
    \begin{itemize}
        \item The answer \answerNA{} means that the paper does not include experiments.
        \item The experimental setting should be presented in the core of the paper to a level of detail that is necessary to appreciate the results and make sense of them.
        \item The full details can be provided either with the code, in appendix, or as supplemental material.
    \end{itemize}

\item {\bf Experiment statistical significance}
    \item[] Question: Does the paper report error bars suitably and correctly defined or other appropriate information about the statistical significance of the experiments?
    \item[] Answer: \answerYes{} %
    \item[] Justification: The data-allocation experiments in Section 4.3 are repeated over three random seeds, with mean performance and min–max error bars reported. Statistical significance tests and trajectory-level AUC comparisons are provided in Appendix I.5. Due to computational cost, the large-scale configuration sweeps in Sections 4.2 and 4.4 use a single random seed, with additional multi-seed experiments provided in Appendix I.9 to assess the stability of the near-optimal configurations.
    \item[] Guidelines:
    \begin{itemize}
        \item The answer \answerNA{} means that the paper does not include experiments.
        \item The authors should answer \answerYes{} if the results are accompanied by error bars, confidence intervals, or statistical significance tests, at least for the experiments that support the main claims of the paper.
        \item The factors of variability that the error bars are capturing should be clearly stated (for example, train/test split, initialization, random drawing of some parameter, or overall run with given experimental conditions).
        \item The method for calculating the error bars should be explained (closed form formula, call to a library function, bootstrap, etc.)
        \item The assumptions made should be given (e.g., Normally distributed errors).
        \item It should be clear whether the error bar is the standard deviation or the standard error of the mean.
        \item It is OK to report 1-sigma error bars, but one should state it. The authors should preferably report a 2-sigma error bar than state that they have a 96\% CI, if the hypothesis of Normality of errors is not verified.
        \item For asymmetric distributions, the authors should be careful not to show in tables or figures symmetric error bars that would yield results that are out of range (e.g., negative error rates).
        \item If error bars are reported in tables or plots, the authors should explain in the text how they were calculated and reference the corresponding figures or tables in the text.
    \end{itemize}

\item {\bf Experiments compute resources}
    \item[] Question: For each experiment, does the paper provide sufficient information on the computer resources (type of compute workers, memory, time of execution) needed to reproduce the experiments?
    \item[] Answer: \answerYes{} %
    \item[] Justification: The paper provides the GPU type and the memory for the experiments.
    \item[] Guidelines:
    \begin{itemize}
        \item The answer \answerNA{} means that the paper does not include experiments.
        \item The paper should indicate the type of compute workers CPU or GPU, internal cluster, or cloud provider, including relevant memory and storage.
        \item The paper should provide the amount of compute required for each of the individual experimental runs as well as estimate the total compute. 
        \item The paper should disclose whether the full research project required more compute than the experiments reported in the paper (e.g., preliminary or failed experiments that didn't make it into the paper). 
    \end{itemize}
    
\item {\bf Code of ethics}
    \item[] Question: Does the research conducted in the paper conform, in every respect, with the NeurIPS Code of Ethics \url{https://neurips.cc/public/EthicsGuidelines}?
    \item[] Answer: \answerYes{} %
    \item[] Justification: The authors have reviewed the NeurIPS Code of Ethics, and the research conforms to the NeurIPS Code of Ethics.
    \item[] Guidelines:
    \begin{itemize}
        \item The answer \answerNA{} means that the authors have not reviewed the NeurIPS Code of Ethics.
        \item If the authors answer \answerNo, they should explain the special circumstances that require a deviation from the Code of Ethics.
        \item The authors should make sure to preserve anonymity (e.g., if there is a special consideration due to laws or regulations in their jurisdiction).
    \end{itemize}

\item {\bf Broader impacts}
    \item[] Question: Does the paper discuss both potential positive societal impacts and negative societal impacts of the work performed?
    \item[] Answer: \answerYes{} %
    \item[] Justification: This work included a discussion on the broader impacts in the main paper and the appendix.
    \item[] Guidelines:
    \begin{itemize}
        \item The answer \answerNA{} means that there is no societal impact of the work performed.
        \item If the authors answer \answerNA{} or \answerNo, they should explain why their work has no societal impact or why the paper does not address societal impact.
        \item Examples of negative societal impacts include potential malicious or unintended uses (e.g., disinformation, generating fake profiles, surveillance), fairness considerations (e.g., deployment of technologies that could make decisions that unfairly impact specific groups), privacy considerations, and security considerations.
        \item The conference expects that many papers will be foundational research and not tied to particular applications, let alone deployments. However, if there is a direct path to any negative applications, the authors should point it out. For example, it is legitimate to point out that an improvement in the quality of generative models could be used to generate Deepfakes for disinformation. On the other hand, it is not needed to point out that a generic algorithm for optimizing neural networks could enable people to train models that generate Deepfakes faster.
        \item The authors should consider possible harms that could arise when the technology is being used as intended and functioning correctly, harms that could arise when the technology is being used as intended but gives incorrect results, and harms following from (intentional or unintentional) misuse of the technology.
        \item If there are negative societal impacts, the authors could also discuss possible mitigation strategies (e.g., gated release of models, providing defenses in addition to attacks, mechanisms for monitoring misuse, mechanisms to monitor how a system learns from feedback over time, improving the efficiency and accessibility of ML).
    \end{itemize}
    
\item {\bf Safeguards}
    \item[] Question: Does the paper describe safeguards that have been put in place for responsible release of data or models that have a high risk for misuse (e.g., pre-trained language models, image generators, or scraped datasets)?
    \item[] Answer: \answerNA{} %
    \item[] Justification: This paper evaluates methods using existing datasets and standard pretrained encoders without releasing high-risk data or models.
    \item[] Guidelines:
    \begin{itemize}
        \item The answer \answerNA{} means that the paper poses no such risks.
        \item Released models that have a high risk for misuse or dual-use should be released with necessary safeguards to allow for controlled use of the model, for example by requiring that users adhere to usage guidelines or restrictions to access the model or implementing safety filters. 
        \item Datasets that have been scraped from the Internet could pose safety risks. The authors should describe how they avoided releasing unsafe images.
        \item We recognize that providing effective safeguards is challenging, and many papers do not require this, but we encourage authors to take this into account and make a best faith effort.
    \end{itemize}

\item {\bf Licenses for existing assets}
    \item[] Question: Are the creators or original owners of assets (e.g., code, data, models), used in the paper, properly credited and are the license and terms of use explicitly mentioned and properly respected?
    \item[] Answer: \answerYes{} %
    \item[] Justification: The paper properly cites the existing datasets and pretrained models and follows the terms of using them.
    \item[] Guidelines:
    \begin{itemize}
        \item The answer \answerNA{} means that the paper does not use existing assets.
        \item The authors should cite the original paper that produced the code package or dataset.
        \item The authors should state which version of the asset is used and, if possible, include a URL.
        \item The name of the license (e.g., CC-BY 4.0) should be included for each asset.
        \item For scraped data from a particular source (e.g., website), the copyright and terms of service of that source should be provided.
        \item If assets are released, the license, copyright information, and terms of use in the package should be provided. For popular datasets, \url{paperswithcode.com/datasets} has curated licenses for some datasets. Their licensing guide can help determine the license of a dataset.
        \item For existing datasets that are re-packaged, both the original license and the license of the derived asset (if it has changed) should be provided.
        \item If this information is not available online, the authors are encouraged to reach out to the asset's creators.
    \end{itemize}

\item {\bf New assets}
    \item[] Question: Are new assets introduced in the paper well documented and is the documentation provided alongside the assets?
    \item[] Answer: \answerYes{} %
    \item[] Justification: The paper constructs experimental subsets and synthetic variants from existing datasets, but it does not introduce or release a new standalone dataset as a contribution. The code used in this paper is released and well-documented on GitHub.
    \item[] Guidelines:
    \begin{itemize}
        \item The answer \answerNA{} means that the paper does not release new assets.
        \item Researchers should communicate the details of the dataset\slash code\slash model as part of their submissions via structured templates. This includes details about training, license, limitations, etc. 
        \item The paper should discuss whether and how consent was obtained from people whose asset is used.
        \item At submission time, remember to anonymize your assets (if applicable). You can either create an anonymized URL or include an anonymized zip file.
    \end{itemize}

\item {\bf Crowdsourcing and research with human subjects}
    \item[] Question: For crowdsourcing experiments and research with human subjects, does the paper include the full text of instructions given to participants and screenshots, if applicable, as well as details about compensation (if any)? 
    \item[] Answer: \answerNA{} %
    \item[] Justification: The paper does not report any new crowdsourcing experiments or new human-subjects studies.
    \item[] Guidelines:
    \begin{itemize}
        \item The answer \answerNA{} means that the paper does not involve crowdsourcing nor research with human subjects.
        \item Including this information in the supplemental material is fine, but if the main contribution of the paper involves human subjects, then as much detail as possible should be included in the main paper. 
        \item According to the NeurIPS Code of Ethics, workers involved in data collection, curation, or other labor should be paid at least the minimum wage in the country of the data collector. 
    \end{itemize}

\item {\bf Institutional review board (IRB) approvals or equivalent for research with human subjects}
    \item[] Question: Does the paper describe potential risks incurred by study participants, whether such risks were disclosed to the subjects, and whether Institutional Review Board (IRB) approvals (or an equivalent approval/review based on the requirements of your country or institution) were obtained?
    \item[] Answer: \answerNA{} %
    \item[] Justification: The work does not involve new crowdsourcing or new human-subjects research.
    \item[] Guidelines:
    \begin{itemize}
        \item The answer \answerNA{} means that the paper does not involve crowdsourcing nor research with human subjects.
        \item Depending on the country in which research is conducted, IRB approval (or equivalent) may be required for any human subjects research. If you obtained IRB approval, you should clearly state this in the paper. 
        \item We recognize that the procedures for this may vary significantly between institutions and locations, and we expect authors to adhere to the NeurIPS Code of Ethics and the guidelines for their institution. 
        \item For initial submissions, do not include any information that would break anonymity (if applicable), such as the institution conducting the review.
    \end{itemize}

\item {\bf Declaration of LLM usage}
    \item[] Question: Does the paper describe the usage of LLMs if it is an important, original, or non-standard component of the core methods in this research? Note that if the LLM is used only for writing, editing, or formatting purposes and does \emph{not} impact the core methodology, scientific rigor, or originality of the research, declaration is not required.
    \item[] Answer: \answerNA{} %
    \item[] Justification: The core method does not involve LLMs as an important, original, or non-standard component; BERT is used only as a standard pretrained encoder for the CivilComments text experiment.
    \item[] Guidelines:
    \begin{itemize}
        \item The answer \answerNA{} means that the core method development in this research does not involve LLMs as any important, original, or non-standard components.
        \item Please refer to our LLM policy in the NeurIPS handbook for what should or should not be described.
    \end{itemize}

\end{enumerate}

\end{document}